\documentclass[5p]{elsarticle}

\usepackage{lineno,hyperref}

\usepackage{soul}
\soulregister\cite7 
\soulregister\citep7 
\soulregister\citet7 
\soulregister\ref7 
\soulregister\pageref7 
\usepackage{mathrsfs}
\usepackage[noend]{algpseudocode}

\usepackage{algorithmicx,algorithm}
\usepackage{bbm}
\usepackage{graphicx}
\usepackage{amsmath}
\usepackage{amssymb}
\usepackage{multirow}
\usepackage{helvet}
\usepackage{courier}
\usepackage{float}

\usepackage{amsfonts}
\usepackage{textcomp}
\usepackage{hyperref}

\usepackage{subfigure}
\usepackage{hyperref}
\usepackage{times}
\usepackage{epsfig}
\usepackage[usenames,dvipsnames]{color}

\usepackage{amsmath,amsfonts}

\usepackage{array}
\usepackage{textcomp}
\usepackage{stfloats}
\usepackage{url}
\usepackage{verbatim}
\usepackage{arydshln}
\usepackage{bbding}
\usepackage{wrapfig}
\usepackage{caption}
\usepackage{booktabs}
\usepackage{colortbl}
\usepackage{newfloat}
\usepackage{listings}
\usepackage{bm}

\definecolor{ceiling}{RGB}{210,42,33}
\definecolor{floor}{RGB}{40,159,7}
\definecolor{wall}{RGB}{162,215,224}
\definecolor{windows}{RGB}{113,158,201}
\definecolor{chair}{RGB}{202,205,76}
\definecolor{bed}{RGB}{224,190,159}
\definecolor{sofa}{RGB}{153,98,192}
\definecolor{table}{RGB}{23,124,181}
\definecolor{tvs}{RGB}{160,187,34}
\definecolor{furniture}{RGB}{225,126,18}
\definecolor{objects}{RGB}{199,174,221}

\definecolor{road}{RGB}{255,1,252}
\definecolor{sidewalk}{RGB}{76,0,74}
\definecolor{parking}{RGB}{255,149,255}
\definecolor{other_grnd}{RGB}{178,4,75}
\definecolor{building}{RGB}{255,198,0}
\definecolor{car}{RGB}{98,152,240}
\definecolor{truck}{RGB}{82,28,183}
\definecolor{bicycle}{RGB}{101,229,248}
\definecolor{motorcycle}{RGB}{35,57,148}
\definecolor{other_veh}{RGB}{100,82,245}
\definecolor{vegetation}{RGB}{2,176,0}
\definecolor{trunk}{RGB}{131,62,6}
\definecolor{terrain}{RGB}{153,238,85}
\definecolor{person}{RGB}{255,26,29}
\definecolor{bicyclist}{RGB}{255,40,202}
\definecolor{motorcycl}{RGB}{150,29,99}
\definecolor{fence}{RGB}{148,125,42}
\definecolor{pole}{RGB}{255,239,143}
\definecolor{traf_sign}{RGB}{248,2,3}

\journal{Information Fusion}

\begin{document}

\begin{sloppypar}
\begin{frontmatter}

\title{Local Feature Matching Using Deep Learning:  A Survey}

\author[bupt_address]{Shibiao Xu}
\author[bupt_address]{Shunpeng Chen}
\author[ucas_address]{Rongtao Xu}
\author[ucas_address]{Changwei Wang}
\author[bupt_address]{Peng Lu}
\author[bupt_address]{Li Guo}

\address[bupt_address]{School of Artificial Intelligence, Beijing University of Posts and Telecommunications, China}
\address[ucas_address]{The State Key Laboratory of Multimodal Artificial Intelligence Systems, Institute of Automation, Chinese Academy of Sciences, China}
\fntext[myfootnote]{Rongtao Xu is the corresponding author.}

\date{} 

\begin{abstract}
Local feature matching enjoys wide-ranging applications in the realm of computer vision, encompassing domains such as image retrieval, 3D reconstruction, and object recognition. However, challenges persist in improving the accuracy and robustness of matching due to factors like viewpoint and lighting variations. In recent years, the introduction of deep learning models has sparked widespread exploration into local feature matching techniques. The objective of this endeavor is to furnish a comprehensive overview of local feature matching methods. These methods are categorized into two key segments based on the presence of detectors. The Detector-based category encompasses models inclusive of Detect-then-Describe, Joint Detection and Description, Describe-then-Detect, as well as Graph Based techniques. In contrast, the Detector-free category comprises CNN Based, Transformer Based, and Patch Based methods. 
Our study extends beyond methodological analysis, incorporating evaluations of prevalent datasets and metrics to facilitate a quantitative comparison of state-of-the-art techniques. The paper also explores the practical application of local feature matching in diverse domains such as Structure from Motion, Remote Sensing Image Registration, and Medical Image Registration, underscoring its versatility and significance across various fields. 
Ultimately, we endeavor to outline the current challenges faced in this domain and furnish future research directions, thereby serving as a reference for researchers involved in local feature matching and its interconnected domains.
A comprehensive list of studies in this survey is available at \href{https://github.com/vignywang/Awesome-Local-Feature-Matching}{https://github.com/vignywang/Awesome-Local-Feature-Matching}.
\end{abstract}

\end{frontmatter}


\section{Introduction}
\label{sec:intro}

In the field of image processing, the core objective of local feature matching tasks is to establish precise feature correspondences between different images. This encompasses various types of image features, such as keypoints, feature regions, straight lines, and curves, among others. Establishing correspondences between similar features in different images serves as the foundation for many computer vision tasks, including image fusion~\cite{tang2022image,cao2023pcnet,hu2023multiscale,sun2023unified,hou2024pos}, visual localization~\cite{sattler2012improving,sattler2017large,cai2019ground,zhang2021reference}, Structure from Motion (SfM)~\cite{agarwal2011building,heinly2015reconstructing,schonberger2016structure,wang2021deep}, Simultaneous Localization and Mapping (SLAM)~\cite{cadena2016past,mur2017orb,zhao2019gslam}, optical flow estimation~\cite{liu2010sift,weinzaepfel2013deepflow,dosovitskiy2015flownet}, image retrieval~\cite{radenovic2018fine,cao2020unifying,chhabra2020content}, and more.

\begin{figure*}[!ht]
    \centering
    \includegraphics[width=\linewidth]{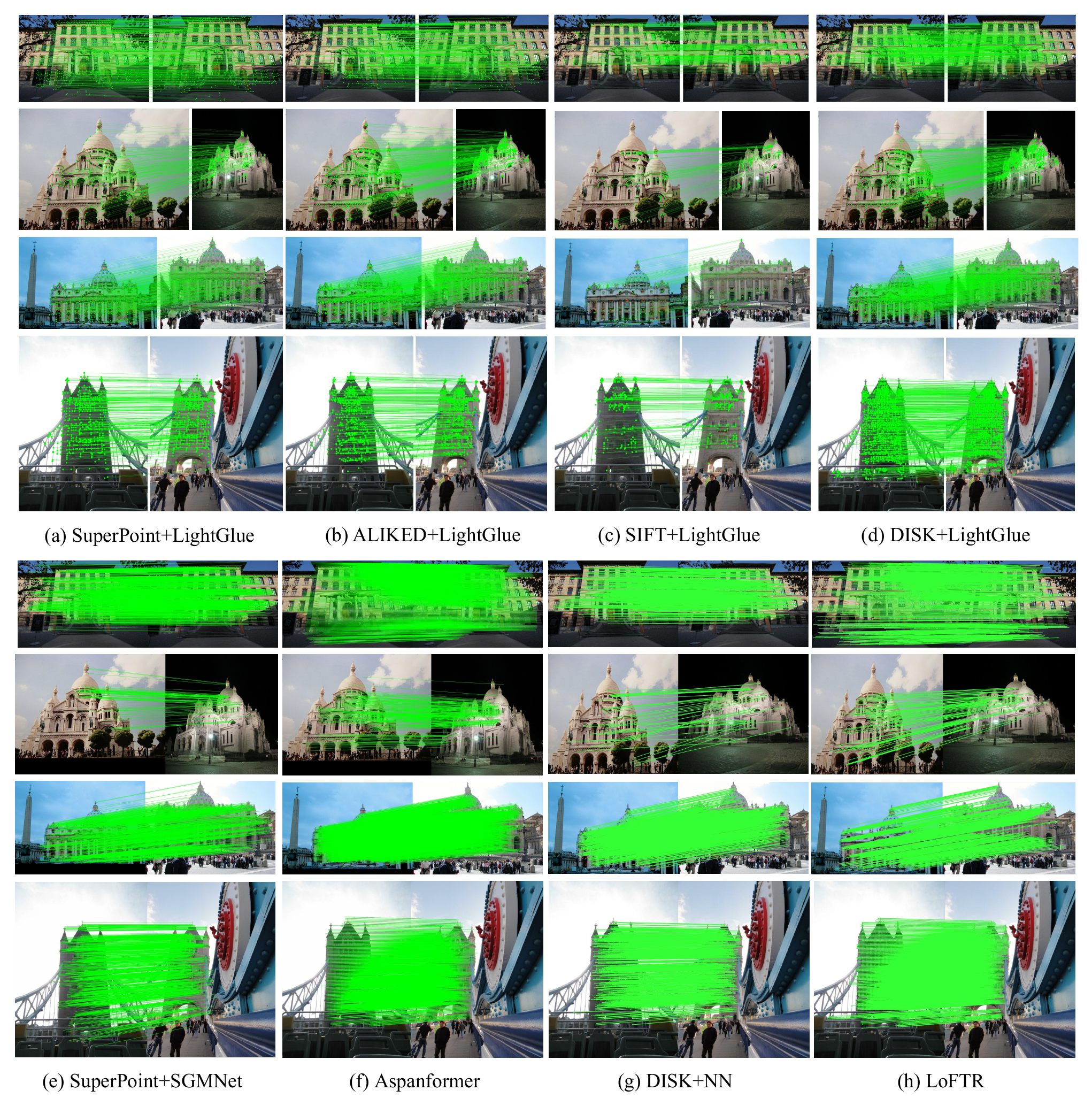}
    \caption{Matching results for outdoor images. It can be observed that for images with significant variations in viewpoint and lighting conditions, the matching task encounters considerable challenges.
}
    \label{fig:Matching_new}
\end{figure*}

Owing to influences such as scale transformations, viewpoint diversity, shifts in illumination, pattern recurrences, and texture variations, the depiction of an identical physical space within distinct images may exhibit substantial divergence. For instance, Figure~\ref{fig:Matching_new} provides a visual representation of the performance of several popular deep learning models engaged in local image matching tasks. Nevertheless, guaranteeing the establishment of precise correspondences between distinct images mandates surmounting manifold perplexities and challenges, engendered by the aforementioned factors. Consequently, the quest for accuracy and dependability in local feature matching continues to be a formidable problem beset with intricacies.

In conventional image matching pipelines, the process can be decomposed into four fundamental steps: feature detection, feature description, feature matching, and geometric transformation estimation. Prior to the advent of deep learning~\cite{zhang2023task,wang2023generalized,cao2023event}, many celebrated algorithms were tailored to focus primarily on one or several stages within this pipeline. Various techniques were committed to the process of feature detection~\cite{harris1988combined,smith1997susan,rosten2005fusing,matas2004robust}, while others were honed on locally executing the task of feature description~\cite{dalal2005histograms,mikolajczyk2005performance,calonder2010brief}. Additionally, certain algorithms have been devised to cater to both feature detection and description~\cite{lowe2004distinctive,bay2006surf,rublee2011orb,leutenegger2011brisk,alcantarilla2012kaze,alcantarilla2011fast}. For the purpose of feature matching, conventional approaches typically relied on either minimizing or maximizing specific well-established metrics, such as the Sum of Squared Differences or correlation. During the stage of geometric transformation estimation, algorithms were generally employed on the basis of techniques akin to RANSAC~\cite{fischler1981random} to estimate the underlying epipolar geometry or homographies. Both traditional handcrafted methods and learning-centric approaches that were constructed upon low-level image features like gradients and sequences of grayscale. Despite being theoretically resilient to certain forms of transformations, these techniques were inherently restricted by the inherent prior knowledge imposed by researchers on their tasks.

In recent years, substantial advancements have been realized in addressing the challenges associated with local feature matching~\cite{xu2022domaindesc,xu2023domainfeat,wu2023segment}, especially those posed by scale variations, shifts in viewpoint, and other forms of diversities. The extant methods of image matching can be sorted into two major categories: Detector-based and Detector-free methods.
Detector-based methods hinge on the detection and description of sparsely distributed keypoints in order to establish matches between images.   The efficacy of these methods is largely dependent on the performance of keypoint detectors and feature descriptors, given their significant role in the process.
Contrastingly, Detector-free methods sidestep the necessity for separate keypoint detection and feature description stages by tapping into the rich contextual information prevalent within the images. These methods enable end-to-end image matching, thereby offering a distinct mechanism to tackle the task.

Image matching plays a pivotal role in the domain of image registration, where it contributes significantly by enabling the precise fitting of transformation functions through a reliable set of feature matches. This functionality positions image matching as a crucial area of study within the broader context of image fusion~\cite{jiang2021review}. To coherently encapsulate the evolution of the local feature matching domain and stimulate innovative research avenues, this paper presents an exhaustive review and thorough analysis of the latest progress in local feature matching, particularly emphasizing the use of deep learning algorithms. In addition, we re-examine pertinent datasets and evaluation criteria, and conduct detailed comparative analyses of key methodologies. Our investigation addresses both the gap and potential bridging between traditional manual methods and modern deep learning technologies. We emphasize the ongoing relevance and collaboration between these two approaches by analyzing the latest developments in traditional manual methods alongside deep learning techniques. Further, we address the emerging focus on multi-modal images. This includes a detailed overview of methods specifically tailored for multi-modal image analysis. Our survey also identifies and discusses the gaps and future needs in existing datasets for evaluating local feature matching methods, highlighting the importance of adapting to diverse and dynamic scenarios. In keeping with current trends, we examine the role of large foundation models in feature matching. These models represent a significant shift from traditional semantic segmentation models~\cite{xu2022instance,xu2021dc,xu2023wave,xu2023spectral,wang2023treating}, offering superior generalization capabilities for a wide array of scenes and objects.

In summary, some of the key contributions of this survey can be summarized as follows:
\begin{itemize}
    \item This survey extensively covers the literature on contemporary local feature matching problems and provides a detailed overview of various local feature matching algorithms proposed since 2018. Following the prevalent image matching pipeline, we primarily categorize these methods into two major classes: Detector-based and Detector-free, and provide a comprehensive review of matching algorithms employing deep learning.
    \item We scrutinize the deployment of these methodologies in a myriad of real-world scenarios, encompassing SfM, Remote Sensing Image Registration, and Medical Image Registration. This investigation highlights the versatility and extensive applicability inherent in local feature matching techniques.
    \item We start from relevant computer vision tasks, review the major datasets involved in local feature matching, and classify them according to different tasks to delve into specific research requirements within each domain.
    \item We analyze various metrics used for performance evaluation and conduct a quantitative comparison of key local feature matching methods.
    \item We present a series of challenges and future research directions, offering valuable guidance for further advancements in this field.
\end{itemize}

\begin{figure*}[ht]
    \centering
    \includegraphics[width=\linewidth]{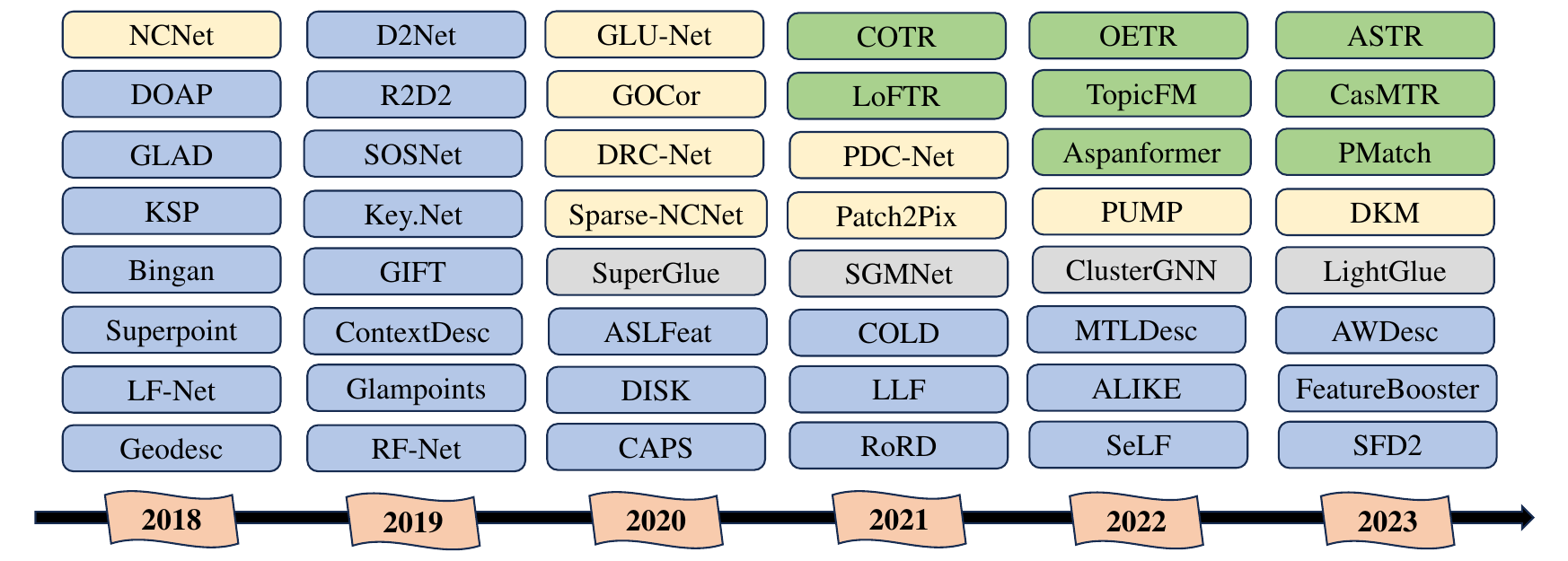}
    \caption{Representative local feature matching methods. Blue and gray represent Detector-based Models, where gray represents the Graph Based method. The yellow and green blocks represent the CNN Based and Transformer Based methods in Detector-free Models, respectively.In 2018, Superpoint~\cite{detone2018superpoint} pioneered the computation of keypoints and descriptors within a single network. Subsequently, numerous works such as D2Net~\cite{dusmanu2019d2}, R2D2~\cite{revaud2019r2d2}, and others attempted to integrate keypoint detection and description for matching purposes. Concurrently, the NCNet~\cite{rocco2018neighbourhood} method introduced four-dimensional cost volumes into local feature matching, initiating a trend in utilizing correlation-based or cost volume-based convolutional neural networks for Detector-free matching research. Building upon this trend, methods like Sparse-NCNet~\cite{rocco2020efficient}, DRC-Net~\cite{li2020dual}, GLU-Net~\cite{truong2020glu}, and PDC-Net~\cite{truong2021learning} emerged. In 2020, SuperGlue~\cite{sarlin2020superglue} framed the task as a graph matching problem involving two sets of features. Following this, SGMNet~\cite{chen2021learning} and ClusterGNN~\cite{shi2022clustergnn} focused on improving the graph matching process by addressing the complexity of matching. In 2021, approaches such as LoFTR~\cite{sun2021loftr} and Aspanformer~\cite{chen2022aspanformer} successfully incorporated Transformer or Attention mechanisms into the Detector-free matching process. They achieved this by employing interleaved self and cross-attention modules, significantly expanding the receptive field and further advancing deep learning-based matching techniques.
}
    \label{fig:methods}
\end{figure*}

It is important to note that the initial surveys~\cite{awrangjeb2012performance,li2015survey,krig2016interest} primarily focused on manual methods, hence they do not provide sufficient reference points for research centered around deep learning. Although recent surveys~\cite{joshi2020recent,ma2021image,jing2022image} have incorporated trainable methods, they have failed to timely summarize the plethora of literature that has emerged in the past five years. Furthermore, many are limited to specific aspects of image matching within the field, such as some articles introducing only the feature detection and description methods of local features, but not including matching~\cite{joshi2020recent}, some particularly focusing on matching of cultural heritage images~\cite{bellavia2022challenges}, and others solely concentrating on medical image registration~\cite{haskins2020deep, bharati2022deep,chen2023survey}, remote sensing image registration~\cite{paul2021comprehensive,zhu2023advances}, and so on.
In this survey, our goal is to provide the most recent and comprehensive overview by assessing the existing methods of image matching, particularly the state-of-the-art learning-based approaches. Importantly, we not only discuss the existing methods that serve natural image applications, but also the wide application of feature matching in SfM, remote sensing images, and medical images. We illustrate the close connection of this research with the field of information fusion through a detailed discussion on the matching of multimodal images. Additionally, we have conducted a thorough examination and analysis of the recent mainstream methods, discussions that are evidently missing in the existing literature.
Figure~\ref{fig:methods} showcases a representative timeline of local feature matching methodologies, which provides insights on the evolution of these methods and their pivotal contributions towards spearheading advancements in the field.

\section{Detector-based Models}

Detector-based methodologies have been the prevailing approach for local feature matching for a considerable duration. Numerous well-established handcrafted works, including SIFT~\cite{lowe2004distinctive} and ORB~\cite{rublee2011orb}, have been broadly adopted for varied tasks within the field of 3D computer vision~\cite{zhang20233d,zhang2022asro}. These traditional Detector-based methodologies typically comprise three primary stages: feature detection, feature description, and feature matching. 
Initially, a set of sparse key points is extracted from the images. Subsequently, in the feature description stage, these key points are characterized using high-dimensional vectors, often designed to encapsulate the specific structure and information of the region surrounding these points. Lastly, during the feature matching stage, correspondences at a pixel level are established through mechanisms like nearest-neighbor searches or more complex matching algorithms. Notable among these are GMS (Grid-based Motion Statistics) by Bian et al.\cite{bian2017gms} and OANET (Order-Aware Network) by Zhang et al.\cite{zhang2019learning}. GMS enhances feature correspondence quality using grid-based motion statistics, simplifying and accelerating matching, while OANET innovatively optimizes two-view matching by integrating spatial contexts for precise correspondence and geometry estimation.
This is typically done by comparing the high-dimensional vectors of keypoints between different images and identifying matches based on the level of similarity – often defined by a distance function in the vector space.

\begin{figure}[ht]
    \centering
    \includegraphics[width=\linewidth]{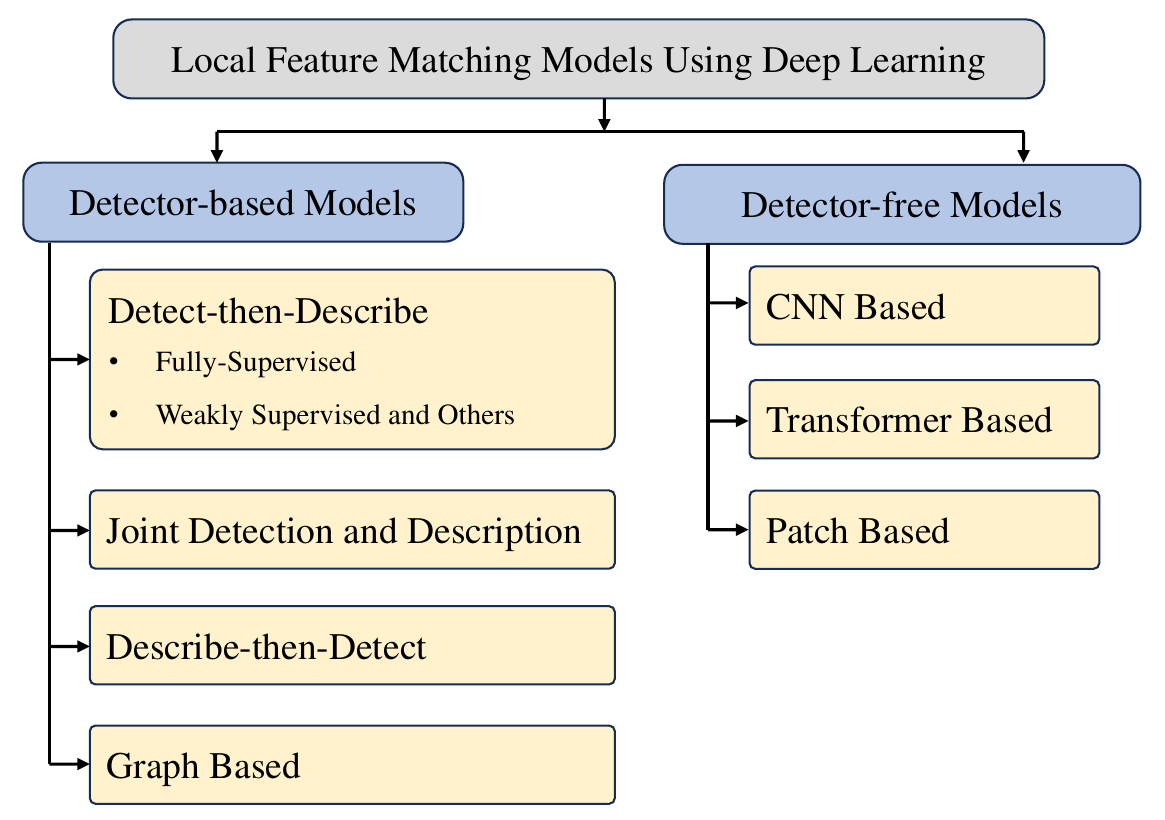}
    \caption{Overview of the Local Feature Matching Models and taxonomy of the most relevant approaches.
}
    \label{fig:Overview}
\end{figure}

However, in the era of deep learning, the rise of data-driven methods has made approaches like LIFT~\cite{yi2016lift} popular. These methods leverage CNNs to extract more robust and discriminative keypoint descriptors, resulting in significant progress in handling large viewpoint changes and local feature illumination variations.
Currently, Detector-based methods can be categorized into four main classes: 1. Detect-then-Describe methods; 2. Joint Detection and Description methods; 3. Describe-then-Detect methods; 4. Graph Based methods. Additionally, we further subdivide Detect-then-Describe methods based on the type of supervised learning into Fully-Supervised methods, Weakly Supervised methods, and Other forms of Supervision methods. This classification is visually depicted in Figure~\ref{fig:Overview}.

\subsection{Detect-then-Describe}

\begin{figure}[ht]
    \centering
    \includegraphics[width=\linewidth]{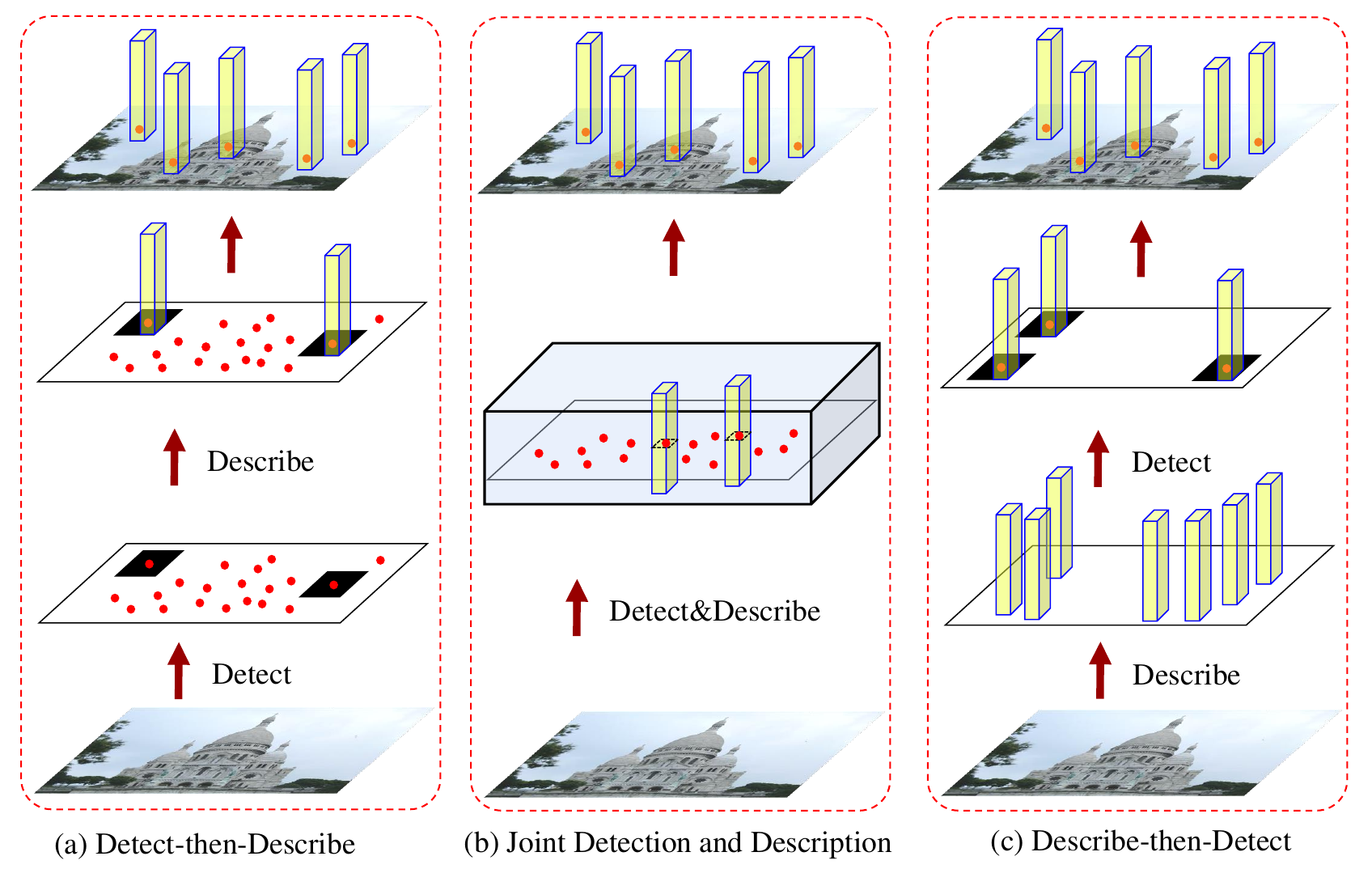}
    \caption{The comparison of various prominent Detector-based pipelines for trainable local feature matching is presented. Here, the categorization is based on the relationship between the detection and description steps: (a) Detect-then-Describe framework, (b) Joint Detection and Description framework, and (c) Describe-then-Detect framework.
}
    \label{fig:Detector-based}
\end{figure}

In feature matching methodologies, the adoption of sparse-to-sparse feature matching is rather commonplace. These methods adhere to a 'detect-then-describe' paradigm, where the primary step involves the detection of keypoint locations. The detector subsequently extracts feature descriptors from patches centered on each detected keypoint. These descriptors are then relayed to the feature description stage. This procedure is typically trained utilizing metric learning methods, which aim to learn a distance function where similar points are close and dissimilar points are distant in the feature space.
To enhance efficiency, feature detectors often focus on small image regions~\cite{yi2016lift}, generally emphasizing low-level structures such as corners~\cite{harris1988combined} or blobs~\cite{lowe2004distinctive}. The descriptors, on the other hand, aim to capture more nuanced, higher-level information within larger patches encompassing the keypoints. Providing verbosity and distinctive details, these descriptors serve as the defining features for matching purposes.
Figure~\ref{fig:Detector-based}(a) illustrates the common structure of the Detect-then-Describe pipeline.

\subsubsection{Fully-Supervised}
The field of local feature matching has undergone a remarkable transformation, primarily driven by the advent of annotated patch datasets~\cite{brown2010discriminative} and the integration of deep learning technologies. This transformation marks a departure from traditional handcrafted methods to more data-driven methodologies, reshaping the landscape of feature matching. This section aims to trace the historical development of these changes, emphasizing the sequential progression and the interconnected nature of various fully supervised methods.
At the forefront of this evolution are CNNs, which have been pivotal in revolutionizing the process of descriptor learning. By enabling end-to-end learning directly from raw local patches, CNNs have facilitated the construction of a hierarchy of local features. This capability has allowed CNNs to capture complex patterns in data, leading to the creation of more specialized and distinct descriptors that significantly enhance the matching process. This revolutionary shift was largely influenced by innovative models like L2Net~\cite{tian2017l2}, which pioneered a progressive sampling strategy. L2Net's approach accentuated the relative distances between descriptors while applying additional supervision to intermediate feature maps. This strategy significantly contributed to the development of robust descriptors, setting a new standard in descriptor learning.

The shift towards these data-driven methodologies, underpinned by CNNs, has not only improved the accuracy and efficiency of local feature matching but has also opened new avenues for research and innovation in this area. As we explore the chronological advancements in this field, we observe a clear trajectory of growth and refinement, moving from the conventional to the contemporary, each method building upon the successes of its predecessors while introducing novel concepts and techniques.
OriNet~\cite{yi2016learning} present a method using CNNs to assign a canonical orientation to feature points in an image, enhancing feature point matching. They introduce a Siamese network~\cite{bromley1993signature} training approach that eliminates the need for predefined orientations and propose a novel GHH activation function, showing significant performance improvements in feature descriptors across multiple datasets.
Building on the architectural principles of L2Net, HardNet~\cite{mishchuk2017working} streamlined the learning process by focusing on metric learning and eliminating the need for auxiliary loss terms, setting a precedent for subsequent models to simplify learning objectives.
DOAP~\cite{he2018local} shifted the focus to a learning-to-rank formulation, optimizing local feature descriptors for nearest-neighbor matching, a methodology that found success in specific matching scenarios and influenced later models to consider ranking-based approaches.
The KSP~\cite{wei2018kernelized} method is notable for its introduction of a subspace pooling methodology, leveraging CNNs to learn invariant and discriminative descriptors.
DeepBit~\cite{lin2018unsupervised} offers an unsupervised deep learning framework to learn compact binary descriptors. It encodes crucial properties like rotation, translation, and scale invariance of local descriptors into binary representations.
Bingan~\cite{zieba2018bingan} proposes a method to learn compact binary image descriptors using regularized Generative Adversarial Networks (GANs).
GLAD~\cite{wei2018glad} addresses the Person Re-Identification task by considering both local and global cues from human bodies. A four-stream CNN framework is implemented to generate discriminative and robust descriptors.
Geodesc~\cite{luo2018geodesc} advances descriptor computation by integrating geometric constraints from SfM algorithms. This approach emphasizes two aspects: first, the construction of training data using geometric information to measure sample hardness, where hardness is defined by the variability between pixel blocks of the same 3D point and uniformity for different points. Second, a geometric similarity loss function is devised, promoting closeness among pixel blocks corresponding to the same 3D point. These innovations enable Geodesc to significantly enhance descriptor effectiveness in 3D reconstruction tasks.
For GIFT~\cite{liu2019gift} and COLD~\cite{lee2021learning}, the former underscores the importance of incorporating underlying structural information from group features to construct potent descriptors. Through the utilization of group convolutions, GIFT generates dense descriptors that exhibit both distinctiveness and invariance to transformation groups. In contrast, COLD introduces a novel approach through a multi-level feature distillation network architecture. This architecture leverages intermediate layers of ImageNet pre-trained convolutional neural networks to encapsulate hierarchical features, ultimately extracting highly compact and robust local descriptors.

Advancing the narrative, our exploration extends to recent strides in fully-supervised methodologies, constituting a noteworthy augmentation of the repertoire of local feature matching capabilities. These pioneering approaches, building upon the foundational frameworks expounded earlier, synergistically elevate and finesse the methodologies that underpin the field.
Continuing the trend of enhancing descriptor robustness, SOSNet~\cite{tian2019sosnet} extends HardNet by introducing a second-order similarity regularization term for descriptor learning. This enhancement involves integrating second-order similarity constraints into the training process, thereby augmenting the performance of learning robust descriptors. The term "second-order similarity" denotes a metric that evaluates the consistency of relative distances among descriptor pairs in a training batch. It measures the similarity between a descriptor pair not only directly but also by considering their relative distances to other descriptor pairs within the same batch.
Ebel et al.~\cite{ebel2019beyond} proposes a local feature descriptor based on a log-polar sampling scheme to achieve scale invariance. This unique approach allows for keypoint matching across different scales and exhibits less sensitivity to occlusion and background motion. Thus, it effectively utilizes a larger image region to improve performance.
To design a better loss function, HyNet~\cite{tian2020hynet} introduces a mixed similarity measure for triplet margin loss and implements a regularization term to constrain descriptor norms, thus establishing a balanced and effective learning framework.
CNDesc~\cite{wang2022cndesc} also investigates L2 normalization, presenting an innovative dense local descriptor learning approach. It uses a special cross-normalization technique instead of L2 normalization, introducing a new way of normalizing the feature vectors.
Key.Net~\cite{barroso2019key} proposes a keypoint detector that combines handcrafted and learned CNN features, and uses scale space representation in the network to extract keypoints at different levels.
To address the non-differentiability issue in keypoint detection methods, ALIKE~\cite{zhao2022alike} offers a differentiable keypoint detection (DKD) module based on the score map. In contrast to methods relying on non-maximum suppression (NMS), DKD can backpropagate gradients and produce keypoints at subpixel levels. This enables the direct optimization of keypoint locations.
S-TREK~\cite{santellani2023s} introduces an advanced local feature extractor that combines a translation and rotation equivariant keypoint detector with a lightweight descriptor extractor. Trained through a reinforcement learning-inspired framework to optimize keypoint repeatability, S-TREK achieves remarkable performance in repeatability and pose recovery across multiple benchmarks, especially excelling in scenarios with in-plane rotations.
ZippyPoint~\cite{kanakis2023zippypoint} is designed based on KP2D~\cite{tang2020neural}, introduces an entire set of accelerated extraction and matching techniques. This method suggests the use of a binary descriptor normalization layer, thereby enabling the generation of unique, length-invariant binary descriptors.

Implementing contextual information into feature descriptors has been a rising trend in the advancement of local feature matching methods.
ContextDesc~\cite{luo2019contextdesc} introduces context awareness to improve off-the-shelf local feature descriptors. It encodes both geometric and visual contexts by using keypoint locations, raw local features, and high-level regional features as inputs. A novel aspect of its training process is the use of an N-pair loss, which is self-adaptive and requires no parameter tuning. This dynamic loss function can allow for a more efficient learning process.
MTLDesc~\cite{wang2022mtldesc} offers a strategy to address the inherent locality issue confronted in the domains of convolutional neural networks. This is attained by introducing an adaptive global context enhancement module and multiple local context enhancement modules to inject non-local contextual information. By adding these non-local connections, it can efficiently learn high-level dependencies between distant features.
Building upon MTLDesc, AWDesc~\cite{wang2023attention} seeks to transfer knowledge from a larger, more complex model (teacher) to a smaller and simpler one (student). This approach leverages the knowledge learned by the teacher, while enabling significantly faster computations with the student, allowing the model to achieve an optimal balance between accuracy and speed.
The focus on context-awareness in these methods emphasizes the importance of considering more global information when describing local features. Each method leverages this information in a slightly different way, leading to diverse but potentially complementary approaches for tackling the challenge of feature matching.

In light of the limitations inherent in traditional image feature descriptors (like gradients, grayscale, etc.), which struggle to handle the geometric and radiometric disparities across different modal image types~\cite{chen2021igs}, there is an emerging focus on frequency-domain-based feature descriptors. These descriptors exhibit improved proficiency in matching cross-modal images. For instance, RIFT~\cite{li2019rift} utilizes FAST ~\cite{rosten2006machine} for extracting repeatable feature points on the phase congruency (PC) map, subsequently constructing robust descriptors using frequency domain information to tackle the challenges in multimodal image feature matching. Building on RIFT, SRIFT ~\cite{cui2020modality} further refines this approach by establishing a nonlinear diffusion scale (NDS) space, thus constructing a multiscale space that not only achieves scale and rotation invariance but also addresses the issue of slow inference speeds associated with RIFT.
With the evolution of deep learning technologies, depth-based methods have demonstrated significant prowess in feature extraction. SemLA~\cite{xie2023semantics} uses semantic guidance in its registration and fusion processes. The feature matching is limited to the semantic sensing area, so as to provide the most accurate registration effect for image fusion tasks.

\subsubsection{Weakly Supervised and Others}

Weakly supervised learning presents opportunities for models to learn robust features without requiring densely annotated labels, offering a solution to one of the largest challenges in training deep learning models. Several weakly supervised local feature learning methods have emerged, leveraging easily obtainable geometric information from camera poses.
AffNet~\cite{mishkin2018repeatability} represents a key advancement in weakly supervised local feature learning, focusing on the learning of affine shape of local features. This method challenges the traditional emphasis on geometric repeatability, showing that it is insufficient for reliable feature matching and stressing the importance of descriptor-based learning. AffNet introduces a hard negative-constant loss function to improve the matchability and geometric accuracy of affine regions. This has proven effective in enhancing the performance of affine-covariant detectors, especially in wide baseline matching and image retrieval. The approach underscores the need to consider both descriptor matchability and repeatability for developing more effective local feature detectors.
GLAMpoints~\cite{truong2019glampoints} presents a semi-supervised keypoint detection method, creatively drawing insights from reinforcement learning loss formulations. Here, rewards are used to calculate the significance of detecting keypoints based on the quality of the final alignment. This method has been noted to significantly impact the matching and registration quality of the final images.
CAPS~\cite{wang2020learning} introduces a weakly supervised learning framework that utilizes the relative camera poses between image pairs to learn feature descriptors. By employing epipolar geometric constraints as supervision signals, they designed differentiable matching layers and a coarse-to-fine architecture, resulting in the generation of dense descriptors.
DISK~\cite{tyszkiewicz2020disk} maximizes the potential of reinforcement learning to integrate weakly supervised learning into an end-to-end Detector-based pipeline using policy gradients. This integrative approach of weak supervision with reinforcement learning can provide more robust learning signals and achieve effective optimization.
~\cite{lee2023learning} proposes a group alignment approach that leverages the power of group-equivariant CNNs. These CNNs are efficient in extracting discriminative rotation-invariant local descriptors. The authors use a self-supervised loss for better orientation estimation and efficient local descriptor extraction.
Weakly and semi-supervised methods using camera pose supervision and other techniques provide useful strategies to tackle the challenges of training robust local feature methods and may pave the way for more efficient and scalable learning methods in this domain.

\subsection{Joint Detection and Description}

Sparse local feature matching has indeed proved very effective under a variety of imaging conditions. Yet, under extreme variations like day-night changes~\cite{zhou2016evaluating}, different seasons~\cite{sattler2018benchmarking}, or weak-textured scenes~\cite{taira2018inloc}, the performance of these features can deteriorate significantly.
The limitations may stem from the nature of keypoint detectors and local descriptors. Detecting keypoints often involves focusing on small regions of the image and might rely heavily on low-level information, such as pixel intensities. This procedure makes keypoint detectors more susceptible to variations in low-level image statistics, which are often affected by changes in lighting, weather, and other environmental factors.
Moreover, when trying to individually learn or train keypoint detectors or feature descriptors, even after carefully optimizing the individual components, integrating them into a feature matching pipeline could still lead to information loss or inconsistencies. This is due to the fact that the optimization of individual components might not fully consider the dependencies and information sharing between the components.
To tackle these issues, the approach of Joint Detection and Description has been proposed. In this approach, the tasks of keypoint detection and description are integrated and learned simultaneously within a single model. This can enable the model to fuse information from both tasks during optimization, better adapting to specific tasks and data, and allowing deeper feature mappings through CNNs.
Such a unified approach can benefit the task by allowing the detection and description process to be influenced by higher-level information, such as structural or shape-related features of the image. Additionally, dense descriptors involve richer image context, which generally leads to better performance.
Figure~\ref{fig:Detector-based} (b) illustrates the common structure of the Joint Detection and Description pipeline.

Image-based descriptor methods, which take the entire image as input and utilize fully convolutional neural networks~\cite{long2015fully} to generate dense descriptors, have seen substantial progress in recent years. These methods often amalgamate the processes of detection and description, leading to improved performance in both tasks.
SuperPoint~\cite{detone2018superpoint} employs a self-supervised approach to simultaneously determine key-point locations at the pixel level and their descriptors. Initially, the model undergoes training on synthetic shapes and images through the application of random homographies. A crucial aspect of the method lies in its self-annotation process with real images. This process involves adapting homographies to enhance the model's relevance to real-world images, and the MS-COCO dataset is employed for additional training. Ground truth key points for these images are generated through various homographic transformations, and key-point extraction is performed using the MagicPoint model. This strategy, which involves aggregating multiple key-point heatmaps, ensures precise determination of key-point locations on real images.
Inspired by Q-learning, LF-Net~\cite{ono2018lf} predicts geometric relationships, such as relative depth and camera poses, between matched image pairs using an existing SfM model. It employs asymmetric gradient backpropagation to train a network for detecting image pairs without needing manual annotation.
Building upon LF-Net, RF-Net~\cite{shen2019rf} introduces a receptive field-based keypoint detector and designs a general loss function term, referred to as 'neighbor mask', which facilitates training of patch selection.
Reinforced SP~\cite{bhowmik2020reinforced} employs principles of reinforcement learning to handle the discreteness in keypoint selection and descriptor matching. It integrates a feature detector into a complete visual pipeline and trains learnable parameters in an end-to-end manner.
R2D2~\cite{revaud2019r2d2} combines grid peak detection with reliability prediction for descriptors using a dense version of the L2-Net architecture, aiming to produce sparse, repeatable, and reliable keypoints.
D2Net~\cite{dusmanu2019d2} adopts a joint detect-and-describe approach for sparse feature extraction. Unlike Superpoint, it shares all parameters between detection and description process and uses a joint formulation that optimizes both tasks simultaneously. Keypoints in their method are defined as local maxima within and across channels of the depth feature maps.
These techniques elegantly illustrate how the integration of detection and description tasks in a unified model leads to more efficient learning and superior performance for local feature extraction under different imaging conditions.

A dual-headed D2Net model with a correspondence ensemble is presented by RoRD~\cite{parihar2021rord} to address extreme viewpoint changes combing vanilla and rotation-robust feature correspondences. HDD-Net~\cite{barroso2020hdd} designs an interactively learnable detector and descriptor fusion network, handling detector and descriptor components independently and focusing on their interactions during the learning process. MLIFeat~\cite{zhang2020mlifeat} devises two lightweight modules used for keypoint detection and descriptor generation with multi-level information fusion utilized to jointly detect keypoints and extract descriptors. LLF~\cite{suwanwimolkul2021learning} proposes utilizing low-level features to supervise keypoint detection. It extends a single CNN layer from the descriptor backbone as a detector and co-learns it with the descriptor to maximize descriptor matching. FeatureBooster~\cite{wang2023featurebooster} introduces a descriptor enhancement stage into traditional feature matching pipelines. It establishes a generic lightweight descriptor enhancement framework that takes original descriptors and geometric attributes of keypoints as inputs. The framework employs self-enhancement based on MLP and cross-enhancement based on transformers~\cite{vaswani2017attention} to enhance descriptors. ASLFeat~\cite{luo2020aslfeat} improves the D2Net using channel and spatial peaks on multi-level feature maps. It introduces a precise detector and invariant descriptor as well as multi-level connections and deformable convolution networks. The dense prediction framework employs deformable convolution networks (DCN) to alleviate limitations caused by keypoint extraction from low-resolution feature maps. SeLF~\cite{fan2022learning} builds on the Aslfeat architecture to leverage semantic information from pre-trained semantic segmentation networks used to learn semantically aware feature mappings. It combines learned correspondences-aware feature descriptors with semantic features, therefore, enhancing the robustness of local feature matching for long-term localization. Lastly, SFD2~\cite{xue2023sfd2} proposes the extraction of reliable features from global regions (e.g., buildings, traffic lanes) with the suppression of unreliable areas (e.g., sky, cars) by implicitly embedding high-level semantics into the detection and description processes. This enables the model to extract globally reliable features end-to-end from a single network.

\subsection{Describe-then-Detect}

One common approach to local feature extraction is the Describe-then-Detect pipeline, entailing the description of local image regions first using feature descriptors followed by the detection of keypoints based on these descriptors. Figure~\ref{fig:Detector-based} (c) serves as an illustration of the standard structure of the Describe-then-Detect pipeline.

D2D~\cite{tian2020d2d} presents a novel framework for keypoint detection called Describe-to-Detect (D2D), highlighting the wealth of information inherent in the feature description phase. This framework involves the generation of a voluminous collection of dense feature descriptors followed by the selection of keypoints from this dataset. D2D introduces relative and absolute saliency measurements of local depth feature maps to define keypoints.
Due to the challenges arising from weak supervision inability to differentiate losses between detection and description stages, PoSFeat~\cite{li2022decoupling} presents a decoupled training approach in the describe-then-detect pipeline specifically designed for weakly supervised local feature learning. This pipeline separates the description network from the detection network, leveraging camera pose information for descriptor learning that enhances performance. Through a novel search strategy, the descriptor learning process more adeptly utilizes camera pose information.
ReDFeat~\cite{deng2022redfeat} uses a mutual weighting strategy to combine multimodal feature learning's detection and description aspects. 
SCFeat~\cite{sun2022shared} proposes a shared coupling bridge strategy for weakly supervised local feature learning. Through shared coupling bridges and cross-normalization layers, the framework ensures the individual, optimal training of description networks and detection networks. This segregation enhances the robustness and overall performance of descriptors.

\subsection{Graph Based}

\begin{figure}[ht]
    \centering
    \includegraphics[width=\linewidth]{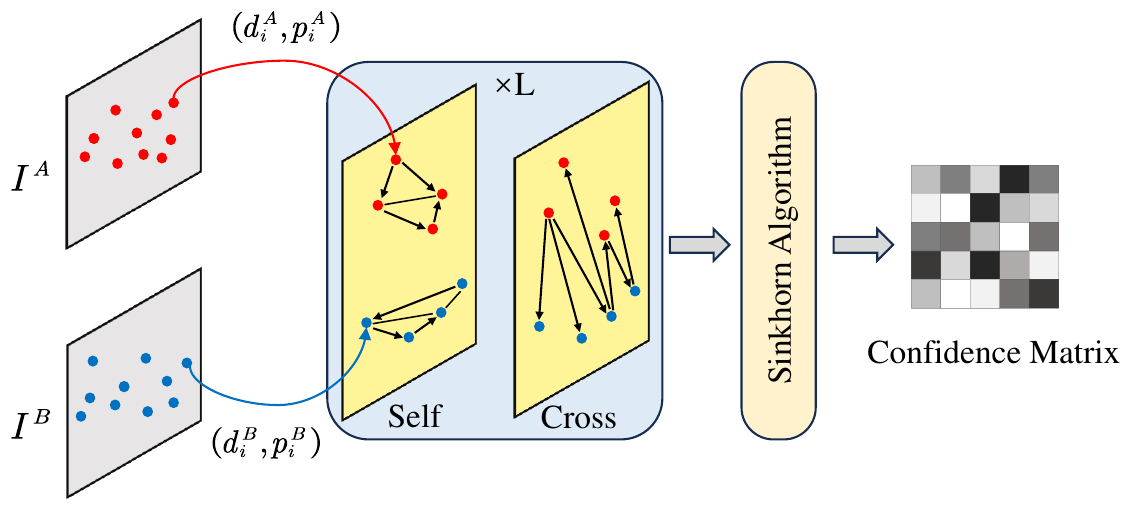}
    \caption{
    General GNN Matching Model Architecture. Firstly, keypoint positions $p_i$ along with their visual descriptors $d_i$ are mapped into individual vectors. Subsequently, self-attention layers and cross-attention layers are thereafter applied alternately, L times, within a graph neural network to create enhanced matching descriptors. Finally, the Sinkhorn Algorithm is utilized to determine the optimal partial assignment.
}
    \label{fig:GNN}
\end{figure}

In the conventional feature matching pipelines, correspondence relationships are established via nearest neighbor (NN) search of feature descriptors, and outliers are eliminated based on matching scores or mutual NN verification. In recent times, attention-based graph neural networks (GNNs)~\cite{zhang2023dualgats} have emerged as effective means to obtain local feature matching.
These approaches create GNNs with keypoints as nodes and utilize self-attention layers and cross-attention layers from Transformers to exchange global visual and geometric information among nodes. This exchange overcomes the challenges posed by localized feature descriptors solely. The ultimate outcome is the generation of matches based on the soft assignment matrix.
Figure~\ref{fig:GNN} provides a comprehensive depiction of the fundamental architecture of Graph-Based matching.

SuperGlue~\cite{sarlin2020superglue} adopts attention graph neural networks and optimal transport methods to address partial assignment problems. It processes two sets of interest points and their descriptors as inputs and leverages self and cross-attention to exchange messages between the two sets of descriptors. The complexity of this method grows quadratically with the number of keypoints, which prompted further exploration in subsequent works.
SGMNet~\cite{chen2021learning} builds on SuperGlue and adds a Seeding Module that processes only a subset of matching points as seeds. The fully connected graph is relinquished for a sparse connection graph. A seed graph neural network is then designed with an attention mechanism to aggregate information.
Keypoints usually exhibit strong correlations with just a few points, resulting in a sparsely connected adjacency matrix for most keypoints. Therefore, ClusterGNN~\cite{shi2022clustergnn} makes use of graph node clustering algorithms to partition nodes in a graph into multiple clusters. This strategy applies attention GNN layers with clustering to learn feature matching between two sets of keypoints and their related descriptors, thus training the subgraphs to reduce redundant information propagation.
MaKeGNN~\cite{li2023learning} introduces bilateral context-aware sampling and keypoint-assisted context aggregation in a sparse attention GNN architecture.

Inspired by SuperGlue, GlueStick~\cite{pautrat2023gluestick} incorporates point and line descriptors into a joint framework for joint matching and leveraging point-to-point relationships to link lines from matched images.
LightGlue~\cite{lindenberger2023lightglue}, in an effort to make SuperGlue adaptive in computational complexity, proposes the dynamic alteration of the network's depth and width based on the matching difficulty between each image pair. It devises a lightweight confidence classifier to forecast and hone state assignments.
DenseGAP~\cite{kuang2022densegap} devises a graph structure utilizing anchor points as sparse, yet reliable priors for inter-image and intra-image contexts. It propagates this information to all image points through directed edges.
HTMatch~\cite{cai2023htmatch} and Paraformer~\cite{lu2023paraformer} study the application of attention for interactive mixing and explore architectures that strike a balance between efficiency and effectiveness.
ResMatch~\cite{deng2023resmatch} presents the idea of residual attention learning for feature matching, re-articulating self-attention and cross-attention as learned residual functions of relative positional reference and descriptor similarity. It aims to bridge the divide between interpretable matching and filtering pipelines and attention-based feature matching networks that inherently possess uncertainty via empirical means.

\section{Detector-free Models}

While the feature detection stage enables a reduction in the search space for matching, handling extreme circumstances, such as image pairs involving substantial viewpoint changes and textureless regions, prove to be difficult when using detection-based approaches, notwithstanding perfect descriptors and matching methodologies~\cite{xie2023deepmatcher}. Detector-free methods, on the other hand, eliminate feature detectors and directly extract visual descriptors on a dense grid spread across the images to produce dense matches. Thus, compared to Detector-based methods, these techniques can capture keypoints that are repeatable across image pairs.

\subsection{CNN Based}

\begin{figure}[ht]
    \centering
    \includegraphics[width=\linewidth]{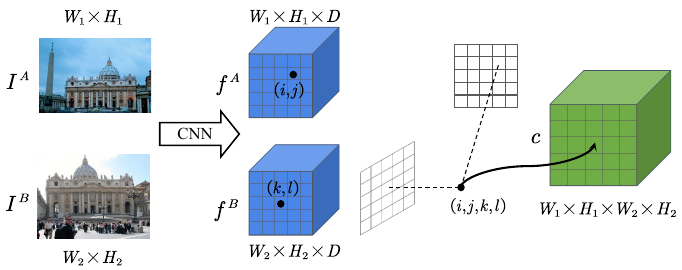}
    \caption{Overview of the 4D correspondence volume. Dense feature maps, denoted as $f^A$ and $f^B$, are extracted from images $I^A$ and $I^A$ using convolutional neural networks. Each individual feature match, $f_{ij}^{A}$ and $f_{kl}^{B}$, corresponds to the matching $\left( i,j,k,l\right)$ coordinates. The 4D correlation tensor $c$ is ultimately formed, which contains scores for all points between a pair of images that could potentially be corresponding points. Subsequently, matching pairs are obtained by analyzing the properties of corresponding points in the four-dimensional space.
}
    \label{fig:4D}
\end{figure}

In the early stages, detection-free matching methodologies often relied on CNNs that used correlation or cost volume to identify potential neighborhood consistencies~\cite{xie2023deepmatcher}. 
Figure~\ref{fig:4D} illustrates the fundamental architecture of the 4D correspondence volume.

NCNet~\cite{rocco2018neighbourhood} analyzes the neighborhood consistency in four-dimensional space of all possible corresponding points between a pair of images, obtaining matches without requiring a global geometric model.
Sparse-NCNet~\cite{rocco2020efficient} utilizes a 4D convolutional neural network on sparse correlation tensors and utilizes submanifold sparse convolutions to significantly reduce memory consumption and execution time.
DualRC-Net~\cite{li2020dual} introduces an innovative methodology for establishing dense pixel-wise correspondences between image pairs in a coarse-to-fine fashion. Utilizing a dual-resolution strategy with a Feature Pyramid Network (FPN)-like backbone, the approach generates a 4D correlation tensor from coarse-resolution feature maps and refines it through a learnable neighborhood consensus module, thereby augmenting matching reliability and localization accuracy.
GLU-Net~\cite{truong2020glu} introduces a global-local universal network applicable to estimating dense correspondences for geometric matching, semantic matching, and optical flow. It trains the network in a self-supervised manner.
GOCor~\cite{truong2020gocor} presents a fully differentiable dense matching module that predicts the global optimization matching confidence between two depth feature maps and can be integrated into state-of-the-art networks to replace feature correlation layers directly.
DCFlow~\cite{xu2017accurate} enhances optical flow estimation by utilizing the four-dimensional cost volume, drawing on methods used in stereo matching. By applying a learned feature embedding and adapting semi-global matching to four dimensions, DCFlow addresses the computational hurdles traditionally linked to this extensive approach. Its efficiency in constructing and processing the cost volume, combined with maintaining accuracy, marks an improvement in integrating optical flow and stereo estimation techniques.
Building on these conceptual advancements, RAFT~\cite{teed2020raft} further refines the approach to dense correspondence estimation. By extracting per-pixel features and constructing multi-scale 4D correlation volumes for all pixel pairs, RAFT introduces a recurrent processing unit that iteratively refines the flow field. This innovative strategy effectively addresses several limitations of previous methods, such as error propagation at coarse resolutions and the neglect of small, fast-moving objects, thereby enhancing the precision and reliability of flow estimation.
Following in the footsteps of these foundational methods, PDC-Net~\cite{truong2021learning} proposes a probabilistic depth network that estimates dense image-to-image correspondences and their associated confidence estimates. It introduces an architecture and an improved self-supervised training strategy to achieve robust uncertainty prediction that is generalizable.
PDC-Net+~\cite{truong2023pdc} introduces a probabilistic deep network designed to estimate dense image-to-image correspondences and their associated confidence estimates. They employ a constrained mixture model to parameterize the predictive distribution, enhancing the modeling capacity for handling outliers.
PUMP~\cite{revaud2022pump} combines unsupervised losses with standard self-supervised losses to augment synthetic images. By utilizing a 4D correlation volume, it leverages the non-parametric pyramid structure of DeepMatching~\cite{revaud2016deepmatching} to learn unsupervised descriptors.
DFM~\cite{efe2021dfm} utilizes a pre-trained VGG architecture as a feature extractor, capturing matches without requiring additional training strategies, thus demonstrating the robust power of features extracted from the deepest layers of the VGG network.

\subsection{Transformer Based}

The CNN's dense feature receptive field may have limitations in handling regions with low texture or discerning between keypoints with similar feature representations. In contrast, humans tend to consider both local and global information when matching in such regions. Given Transformers' success in computer vision tasks such as image classification~\cite{dosovitskiy2020image}, object detection~\cite{carion2020end}, and semantic segmentation~\cite{xu2023rssformer,xu2023scd,xu2023dual,10328690,10214591}, researchers have explored incorporating Transformers' global receptive field and long-range dependencies into local feature matching. Various approaches that integrate Transformers into feature extraction networks for local feature matching have emerged.

Given that the only difference between sparse matching and dense matching is the quantity of points to query, COTR~\cite{jiang2021cotr} combines the advantages of both approaches. It learns two matching images jointly with self-attention, using some keypoints as queries and recursively refining matches in the other image through a corresponding neural network. This integration combines both matches into one parameter-optimized problem.
ECO-TR~\cite{tan2022eco} strives to develop an end-to-end model to accelerate COTR by intelligently connecting multiple transformer blocks and progressively refining predicted coordinates in a coarse-to-fine manner on a shared multi-scale feature extraction network.
LoFTR~\cite{sun2021loftr} is groundbreaking because it creates a GNN with keypoints as nodes, utilizing self-attention layers and mutual attention layers to obtain feature descriptors for two images and generating dense matches in regions with low texture. To overcome the absence of local attention interaction in LoFTR, Aspanformer~\cite{chen2022aspanformer} proposes an uncertainty-driven scheme based on flow prediction probabilistic modeling that adaptively varies the local attention span to allocate different context sizes for different positions.
Contrary to the detect-then-describe strategy of S-TREK~\cite{santellani2023s}, which leverages a translation and rotation equivariant keypoint detector paired with a lightweight descriptor extractor, SE2-LoFTR~\cite{bokman2022case} adopts a detector-free paradigm, seamlessly extracting pixel-level correspondences between pairs of images without necessitating the preliminary step of keypoint detection. This model enhances the original LoFTR framework by incorporating a steerable CNN, thereby achieving inherent equivariance to translations and rotations. This modification significantly boosts the model's resilience to rotational variances, showcasing the model's unique contribution to the domain of feature matching through direct image correspondence. SE2-LoFTR's approach exemplifies the versatility and efficiency of detector-free models in handling complex image matching scenarios, particularly those involving significant rotational movements.

To address the challenges posed by the presence of numerous similar points in dense matching approaches and the limitations on the performance of linear transformers themselves, several recent works have proposed novel methodologies.
Quadtree~\cite{tang2022quadtree} introduces quadtree attention to quickly skip calculations in irrelevant regions at finer levels, reducing the computational complexity of visual transformations from quadratic to linear.
OETR~\cite{chen2022guide} introduces the Overlap Regression method, which uses a Transformer decoder to estimate the degree of overlap between bounding boxes in an image. It incorporates a symmetric center consistency loss to ensure spatial consistency in the overlapping regions. OETR can be inserted as a preprocessing module into any local feature matching pipeline.
MatchFormer~\cite{wang2022matchformer} devises a hierarchical transformer encoder and a lightweight decoder. In each stage of the hierarchical structure, cross-attention modules and self-attention modules are interleaved to provide an optimal combination path, enhancing multi-scale features.
CAT~\cite{ma2022correspondence} proposes a context-aware network based on the self-attention mechanism, where attention layers can be applied along the spatial dimension for higher efficiency or along the channel dimension for higher accuracy and a reduced storage burden.
TopicFM~\cite{giang2022topicfm} encodes high-level context in images, utilizing a topic modeling approach. This improves matching robustness by focusing on semantically similar regions in images.
ASTR~\cite{yu2023adaptive} introduces an Adaptive Spot-guided Transformer, which includes a point-guided aggregation module to allow most pixels to avoid the influence of irrelevant regions, while using computed depth information to adaptively adjust the size of the grid at the refinement stage.
DeepMatcher~\cite{xie2023deepmatcher} introduces the Feature Transformation Module to ensure a smooth transition of locally aggregated features extracted from CNNs to features with a global receptive field, extracted from Transformers. It also presents SlimFormer, which builds deep networks, employing a hierarchical strategy that enables the network to adaptively absorb information exchange within residual blocks, simulating human-like behavior.
OAMatcher~\cite{dai2023oamatcher} proposes the Overlapping Areas Prediction Module to capture keypoints in co-visible regions and conduct feature enhancement among them, simulating how humans shift focus from entire images to overlapping regions. They also propose a Matching Label Weight Strategy to generate coefficients for evaluating the reliability of true matching labels, using probabilities to determine whether the matching labels are correct.
CasMTR~\cite{cao2023improving} proposes to enhance the transformer-based matching pipeline by incorporating new stages of cascade matching and NMS detection.

PMatch~\cite{zhu2023pmatch} enhances geometric matching performance by pretraining with transformer modules using a paired masked image modeling pretext task, utilizing the LoFTR module.
To effectively leverage geometric priors, SEM~\cite{chang2023structured} introduces a structured feature extractor that models relative positional relationships between pixels and highly confident anchor points. It also incorporates epipolar attention and matching techniques to filter out irrelevant regions based on epipolar constraints.
DKM~\cite{edstedt2023dkm} addresses the two-view geometric estimation problem by devising a dense feature matching method. DKM presents a robust global matcher with a kernel regressor and embedded decoder, involving warp refinement through large depth-wise kernels applied to stacked feature maps.
Building on this, RoMa~\cite{edstedt2023roma} represents a significant advancement in dense feature matching by applying a Markov chain framework to analyze and improve the matching process. It introduces a two-stage approach: a coarse stage for globally consistent matching and a refinement stage for precise localization. This method, which separates the initial matching from the refinement process and employs robust regression losses for greater accuracy, has led to notable improvements in matching performance, outperforming current SotA.

\subsection{Patch Based}

\begin{figure}[ht]
    \centering
    \includegraphics[width=\linewidth]{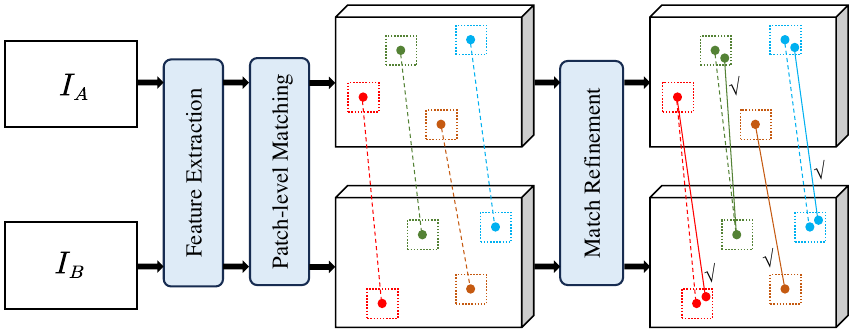}
    \caption{Illustration of the Patch-Based Pipeline. Following the extraction of image features, the Match Refinement process is performed on the matched areas obtained from patch-level matches, resulting in refined point matching.
}
    \label{fig:Patch_level}
\end{figure}

The Patch-Based matching approach enhances point correspondences by matching local image regions. It involves dividing images into patches, extracting descriptor vectors for each, and then matching these vectors to establish correspondences. This technique accommodates large displacements and is valuable in various computer vision applications. Figure~\ref{fig:Patch_level} illustrates the general architecture of the Patch-Based matching approach.

Patch2Pix~\cite{zhou2021patch2pix} proposes a weakly supervised method to learn correspondences consistent with extreme geometric transformations between image pairs. It adopts a two-stage detection-refinement strategy for correspondence prediction, where the first stage captures semantic information and the second stage handles local details. It introduces a novel refinement network that utilizes weak supervision from extreme geometric transformations and outputs confidence in match positions and outlier rejection, allowing for geometrically consistent correspondence prediction.
AdaMatcher~\cite{huang2023adaptive} addresses the geometric inconsistency issue caused by applying the one-to-one assignment criterion in patch-level matching. It adaptively assigns patch-level matches while estimating the scale between images to improve the performance of dense feature matching methods in extreme cases.
PATS~\cite{ni2023pats} proposes Patch Area Transportation with Subdivision (PATS) to learn scale differences in a self-supervised manner. It can handle multiple-to-multiple relationships unlike bipartite graph matching, which only handles one-to-one matches.
SGAM~\cite{zhang2023searching} presents a hierarchical feature matching framework that first performs region matching based on semantic clues, narrowing down the feature matching search space to region matches with significant semantic distributions between images. It then refines the region matches through geometric consistency to obtain accurate point matches.

\section{Local Feature Matching Applications}

\subsection{Structure from Motion}

SfM represents a foundational computational vision process, indispensable for deducing camera orientations, intrinsic parameters, and volumetric point clouds from an array of diverse scene images.  This process substantiates endeavors such as visual localization, multi-view stereo, and the synthesis of innovative perspectives.  The developmental trajectory of SfM, underscored by extensive scholarly investigations, has engendered firmly established methodologies, sophisticated open-source frameworks exemplified by Bundler~\cite{snavely2006photo} and COLMAP~\cite{schonberger2016structure}, and advanced proprietary software solutions.  These frameworks are meticulously tailored to ensure precision and scalability when handling expansive scenes.

Conventional SfM methodologies rely upon the identification and correlation of sparse characteristic points dispersed across manifold perspectives.  Nonetheless, this methodology encounters formidable challenges in regions with scant textural features, where the identification of keypoints becomes a vexing undertaking. Lindenberger et al.~\cite{lindenberger2021pixel} ameliorate this predicament by meticulously refining the initial keypoints and effecting subsequent adjustments to both points and camera orientations during post-processing. The proposed approach strategically balances an initial rudimentary estimation with sparse localized features and subsequent fine-tuning through locally-precise dense features, thereby elevating precision under challenging conditions.
Recent strides in SfM have transitioned towards holistic approaches that either directly regress poses~\cite{parameshwara2022diffposenet,zhang2022relpose} or employ differential bundle adjustment~\cite{tang2018ba,gu2021dro}.  These approaches, circumventing explicit feature correlation, sidestep the challenges associated with suboptimal feature matching.  He et al.~\cite{he2023detector} introduce an innovative SfM paradigm devoid of detectors, harnessing detector-free matchers to defer the determination of keypoints.  This strategy adeptly addresses the prevalent multi-view inconsistency in detector-free matchers, showcasing superior efficacy in texture-impoverished scenes relative to conventional detector-centric systems.

The evolutionary trajectory of SfM methodologies is discernible in the pivot from traditional sparse feature identification towards sophisticated, occasionally end-to-end, dense matching paradigms.  The assimilation of these pioneering methodologies into extant SfM workflows is enhancing precision and resilience, particularly in arduous scenes.  However, the seamless integration of these methodologies into contemporary SfM systems remains an intricate challenge.

\subsection{Remote Sensing Image registration}

In the domain of remote sensing, the emergence of deep learning has heralded a revolutionary epoch in Multimodal Remote Sensing Image Registration (MRSIR)~\cite{jiang2021review,hughes2020deep,ye2022multiscale,chen2022hierarchical}, augmenting conventional area- and feature-based techniques with a Learning-Based Pipeline (LBP)~\cite{liu2022progressive,ye2022robust}. This LBP diverges into several pioneering approaches: amalgamating deep learning with traditional registration methods, bridging the multimodal chasm through modality transformation, and directly regressing transformation parameters for a comprehensive MRSIR framework~\cite{zhu2023advances}. Techniques such as (pseudo-) Siamese networks and Generative Adversarial Networks (GANs) have played a pivotal role in this evolution, facilitating the management of geometric distortions and nonlinear radiometric disparities~\cite{hughes2018identifying,quan2018deep}. For instance, the utilization of conditional GANs has enabled the creation of pseudo-images~\cite{merkle2018exploring}, thereby enhancing the precision of established methods like NCC~\cite{shi2012visual} and SIFT~\cite{liu2010sift}.

In the LBP, a multitude of innovative methods and architectures have been formulated. MUNet~\cite{ye2022multiscale}, a multiscale strategy for learning transformation parameters, and fully convolutional networks for scale-specific feature extraction stand as quintessential examples of this innovation, addressing the challenges of nonrigid MRSIR~\cite{zampieri2018multimodal}. Enriching the LBP further, various research endeavors have concentrated on integrating middle- or high-level features extracted by CNNs with classical descriptors, surmounting limitations of traditional methodologies. For example, Ye et al.~\cite{ye2018remote,ye2022multiscale} devised a novel multispectral image registration technique employing a CNN and SIFT amalgamation, substantially enhancing registration efficacy. Similarly, Wang et al.~\cite{wang2017multi} developed an end-to-end deep learning architecture that discerns the mapping function between image patch-pairs and their matching labels, employing transfer learning for expedited training. Ma et al.~\cite{ma2019novel} introduced a coarse-to-fine registration method using CNN and local features, attaining a profound pyramid feature representation via VGG-16.
Zhou et al.~\cite{zhou2021robust} developed a deep learning-based method for matching synthetic aperture radar (SAR) and optical images, concentrating on extracting Multiscale Convolutional Gradient Features (MCGFs) using a shallow pseudo-Siamese network. This approach effectively captures commonalities between SAR and optical images, transcending the confines of handcrafted features and diminishing the necessity for extensive model parameters. Cui et al.~\cite{cui2021map} introduced MAP-Net, an image-centric convolutional network integrating Spatial Pyramid Aggregated Pooling (SPAP) and an attention mechanism, proficient in addressing geometric distortions and radiometric variations in cross-modal images by embedding original images to extract high-level semantic information and employing PCA for enhanced matching accuracy.

Notwithstanding these advancements, challenges in dataset construction and method generalization persist, principally owing to the diverse and intricate nature of remote sensing images~\cite{ye2022multiscale}. 
The development of comprehensive and representative training datasets, coupled with innovative methodologies meticulously tailored for remote sensing imagery, remains an imperative objective. Moreover, there is a dearth of valuable research in pixel-level fusion of radar and optical images, requiring more attention in future endeavors~\cite{wang2023review}.

\subsection{Medical Image Registration}

The field of medical image registration has undergone significant evolution with the integration of sophisticated deep learning techniques, particularly in motion estimation and 2D-3D registration. These advancements not only represent a technological leap but also open new vistas in various medical applications~\cite{chen2023survey}.

Motion estimation in medical imaging, a crucial aspect of registration, has been substantially refined with deep learning. Unsupervised optical flow and point tracking techniques, as expounded by researchers like Bian et al.~\cite{bian2022learning}, Ranjan et al.~\cite{ranjan2019competitive}, and Harley et al.~\cite{harley2022particle}, address the complexities inherent in medical image data, such as variability in patient anatomy and the need for maintaining anatomical integrity through diffeomorphism and incompressibility. Deep learning-based methods have shown efficacy in motion estimation across different organs, from the heart to the lungs. For example, the application of variational autoencoder-based models, as demonstrated by Qin et al.~\cite{qin2023generative}, by traversing the biomechanically reasonable deformation manifold to search for the best transformation of a given heart sequence, better motion tracking accuracy and more reasonable estimation of myocardial motion and strain are obtained, enhancing the realism and clinical reliability of motion estimation. DeepTag~\cite{ye2021deeptag} and DRIMET~\cite{bian2023drimet} showcase sophisticated methodologies for tracking internal tissue motion, particularly in the context of Tagged-MRI. These approaches exemplify the ability to estimate dense 3D motion fields through advanced unsupervised learning techniques in medical imaging.
The recent advancement in one-shot learning for deformable medical image registration is conspicuously exemplified in contemporary research, particularly in the application of one-shot learning to complex 3D and 4D medical datasets, thereby enhancing accuracy, reducing dependency on large training datasets, and broadening the scope of applicability.      
Fechter et al.~\cite{fechter2020one} introduced a one-shot learning approach for deep motion tracking in 3D and 4D datasets, addressing the challenge of requiring extensive training data. Their method concatenates images from different phases in the channel dimension, utilizing a U-Net architecture with a coarse-to-fine strategy. This approach allows the simultaneous calculation of forward and inverse transformations in 3D datasets.
Zhang et al.~\cite{zhang2021groupregnet} introduced GroupRegNet, a one-shot deep learning method designed for 4D image registration. It employs an implicit template, effectively reducing bias and accumulated error. The simplicity of GroupRegNet's network design and its direct registration process eliminates the need for image partitioning, resulting in a significant enhancement in both computational efficiency and accuracy.      Furthering the development in this domain, Ji et al.~\cite{ji2022one} proposed a temporal-spatial method for lung 4D-CT image registration.      This method integrates a CNN-ConvLSTM hybrid architecture, adeptly modeling the temporal motion of images while incorporating a dual-stream approach to address periodic motion constraints.
The Hybrid Paradigm-based Registration Network (HPRN)~\cite{iqbal2024hybrid} introduces an unsupervised learning framework for 4D-CT lung image registration, effectively handling large deformations without ground truth data. HPRN achieves superior registration accuracy by learning multi-scale features, incorporating advanced loss functions, and avoiding preprocessing steps like cropping and scaling.

2D-3D registration is a critical component, particularly in interventional procedures~\cite{pfandler2019technical}. This process is vital for accurately overlaying 2D images (like X-Ray, ultrasound, or endoscopic images) onto 3D pre-operative CT or MR images. The key challenge here lies in the accurate geometric alignment of these differing dimensionalities. Traditional approaches to 2D-3D registration have relied on iterative optimization methods with similarity metrics based on image intensity~\cite{maes1997multimodality}. However, these methods often struggle with the non-convex nature of the problem, potentially resulting in convergence to incorrect solutions if the initial estimate is not close to the actual solution. This is compounded by the inherent difficulty of representing 3D spatial information on 2D images, leading to registration ambiguity.
Recent advancements, however, have seen a shift towards deep learning-based methods. These approaches, unlike their traditional counterparts, do not require explicit functional mappings, allowing for more robust solutions to the registration challenge~\cite{unberath2021impact}. In the realm of recent developments concerning 2D-3D medical image registration, Jaganathan et al.~\cite{jaganathan2023self} have introduced a self-supervised paradigm designed for the fusion of X-ray and CT images. This method leverages simulated X-ray projections to facilitate the training of deep neural networks, culminating in a noteworthy enhancement of registration accuracy and success ratios. Simultaneously, Huang et al.~\cite{huang2022novel} have devised a two-stage framework tailored for neurological interventions. This innovative approach amalgamates CNN regression with centroid alignment, manifesting superior efficacy in real-time clinical applications. 
In addition to rigid 2D-3D registration, there is growing interest in non-rigid registration, which is crucial in applications like cephalometry, lung tumor tracking in radiation therapy, and Total Hip Arthroplasty (THA)~\cite{pei2017non,foote2019real,yu2017non}. Deep learning models, such as convolutional encoders, have been used to address the challenges of non-rigid registration~\cite{li2020non,dong20232d}.

The convergence of motion estimation and 2D-3D registration techniques in medical image registration addresses critical challenges in parameter optimization and ambiguity, enhancing both the speed and accuracy of medical imaging processes. The ongoing evolution in this field is poised to revolutionize diagnostic and interventional procedures, making them more efficient, patient-centric, and outcome-focused. Interested readers can refer to the comprehensive survey~\cite{chen2023survey, bharati2022deep} for a detailed overview of deep learning-based approaches to medical image registration.

\section{Local Feature Matching Datasets}

Local feature matching methods are often evaluated based on their effectiveness in downstream tasks.  In this section, we will provide summaries of some of the most widely used datasets for evaluating local feature matching.  We categorize these datasets into five groups: Image Matching Datasets, Relative Pose Estimation Datasets, Visual Localization Datasets, Optical Flow Estimation Datasets, and Structure from Motion Datasets.  For each dataset, we will provide detailed information about the features it encompasses.

\subsection{Image Matching Datasets}

\textbf{HPatches}~\cite{balntas2017hpatches} benchmark stands as a prominent yardstick for image matching endeavors. It comprises 116 scene sequences distinguished by fluctuations in viewpoint and luminosity. Within each scene, there are 5 pairs of images, with the inaugural image serving as the point of reference, and the subsequent images in the sequence progressively intensifying in complexity. This dataset is bifurcated into two distinct domains: viewpoint, encompassing 59 sequences marked by substantial viewpoint alterations, and illumination, encompassing 57 sequences marked by substantial variances in illumination, spanning both natural and artificial luminance conditions. In each test sequence, one reference image is paired with the remaining five images. It is worth noting that according to the evaluation method of D2Net, 56 sequences with significant viewpoint changes and 52 sequences with significant illumination changes are usually used to evaluate the performance of the network. Starting from SuperPoint~\cite{detone2018superpoint}, the HPatches dataset has also been used to evaluate the performance of local descriptors in homography estimation tasks.

\textbf{Roto-360}~\cite{lee2023learning} is an evaluation dataset consisting of 360 image pairs. These pairs feature rotations within a plane ranging from 0° to 350° in 10° intervals. The dataset was generated by randomly selecting and rotating ten HPatches images, making it valuable for assessing descriptor performance in terms of rotation invariance.

\textbf{ZEB}~\cite{shen2024gim} unveils an innovative zero-shot evaluation benchmark, meticulously designed to tackle the challenge of generalization in image matching across varied domains. It harmonizes data from eight real-world and four simulated datasets, encapsulating a broad spectrum of image resolutions, environmental conditions, and perspectives. This amalgamation culminates in a robust benchmark featuring 46,000 evaluation image pairs, curated across five distinct image overlap ratios, spanning from 10\% to 50\%, ascertained through verifiable ground truth poses and depth maps. ZEB's comprehensive and expansive dataset stands unparalleled in scope and diversity, markedly surpassing the conventional 1,500 in-domain image pairs utilized in prevailing methodologies, representing a monumental advancement in the evaluation of cross-domain generalization capabilities of image matching models. With its rigorous assembly, ZEB not only elucidates the generalization disparities of existing domain-specific models in uncontrolled scenarios but also establishes a novel benchmark for gauging the resilience and versatility of image matching algorithms against real-world adversities.

\textbf{The Image Matching Challenge (IMC) series}, initiated in 2019, has consistently pushed the boundaries of local feature matching methodologies through meticulously curated datasets and a comprehensive benchmark evaluation framework. Each year, the IMC introduces new challenges and datasets, reflecting the evolving complexities and diversities encountered in real-world image matching scenarios. From leveraging large collections of images for dense 3D reconstructions to exploring the nuances of phototourism, urban landscapes, and the intricate details captured by UAVs and DSLRs, the IMC series provides a robust platform for advancing image matching technologies. The IMC 2020~\cite{cvpr2020_image_matching_challenge} dataset was centered around generating dense and accurate 3D reconstructions from vast image collections using off-the-shelf SfM techniques, such as Colmap. This challenge presented the Phototourism dataset, derived from photo-tourism image collections of famous landmarks, to evaluate and benchmark different image matching methods against pseudo-ground truth poses and depth maps. Training data included images, poses, depth maps, and co-visibility estimates, with a validation set available for method fine-tuning before submission. Building upon the foundational work of IMC 2020, the IMC 2021~\cite{cvpr2021_image_matching_challenge} expanded its horizons by introducing three datasets: "Phototourism", "PragueParks", and "GoogleUrban", each showcasing unique challenges for image matching techniques. While "Phototourism" continued from the 2020 challenge, "PragueParks" introduced imagery from video sequences captured with modern smartphones, and "GoogleUrban" focused on evaluating localization algorithms with images collected from different cell phones across the globe. For IMC 2022~\cite{cvpr2022_image_matching_challenge} and IMC 2023~\cite{cvpr2023_image_matching_challenge}, participants are encouraged to explore the Kaggle official website for detailed information on the datasets and benchmarks. The IMC 2023 notably introduces three new datasets - "Heritage", "Haiper", and "Urban" - each presenting unique challenges such as UAV-to-ground imaging, day-night variations, repeated patterns, wiry objects, and scale changes. These datasets aim to test the limits of current image matching algorithms and foster the development of novel solutions capable of addressing the intricate challenges of real-world image matching scenarios.

\subsection{Relative Pose Estimation Datasets}

\textbf{ScanNet}~\cite{dai2017scannet} is a large-scale indoor dataset with well-defined training, validation, and test splits comprising approximately 230 million well-defined image pairs from 1613 scenes. This dataset includes ground truth and depth images and contains more regions with repetitive and weak textures compared to the Hpatches dataset, thus posing greater challenges.

\textbf{YFCC100M}~\cite{thomee2016yfcc100m} is a vast dataset with a diverse collection of internet images of various tourist landmarks. It comprises 100 million media objects, of which approximately 99.2 million are photos and 0.8 million are videos, with each media object represented by several metadata pieces, such as the Flickr identifier, owner name, camera information, title, tags, geo-location, and media source. Typically, a subset of YFCC100M is used for evaluation, consisting of four popular landmark image sets of 1000 image pairs each, resulting in a total of 4000 pairs for the test set and following the conventions used in~\cite{zhang2019learning,sarlin2020superglue,chen2021learning,chen2022aspanformer}.

\textbf{MegaDepth}~\cite{li2018megadepth} is a dataset designed to address the challenging task of matching under extreme viewpoint changes and repetitive patterns. It comprises 1 million image pairs from 196 different outdoor scenes, each with known poses and depth information, and can be used to validate pose estimation effectiveness in outdoor scenarios. The authors also provide depth maps generated from sparse reconstruction and multi-view stereo computation using COLMAP~\cite{schonberger2016structure}.

\textbf{EVD (Extreme Viewpoint Dataset)}~\cite{mishkin2015mods} is a meticulously curated dataset, specifically developed for the assessment of two-view matching algorithms under scenarios of extreme viewpoint alterations. It amalgamates image pairs from a variety of publicly accessible datasets, distinguished by their intricate geometric configurations. EVD's creation was motivated by the necessity to evaluate the resilience of matching methodologies in contexts characterized by pronounced variations in viewpoints.

\textbf{WxBS (Wide multiple Baseline Stereo)}~\cite{mishkin2015wxbs} addresses a more expansive challenge within the realm of wide baseline stereo matching, encompassing disparities in multiple facets of image acquisition such as viewpoint, lighting, sensor type, and visual alterations. This dataset comprises 37 image pairs, featuring a blend of urban and natural environments, systematically categorized based on the presence of various complicating factors. The ground truth for WxBS is established through a collection of manually selected correspondences, capturing the segments of the scene that are visible in both images. WxBS serves as a pivotal tool for the appraisal of algorithms tailored for image matching under a spectrum of demanding conditions.

\subsection{Visual Localization Datasets}

\textbf{Aachen Day-Night}~\cite{sattler2012image} is a dataset consisting of 4328 daytime images and 98 nighttime images, which are used for localization tasks. This benchmark challenges matching between daytime and nighttime images, making it a challenging dataset to work with.
\textbf{Aachen Day-Night v1.1}~\cite{zhang2021reference} is an updated version of the Aachen Day-Night dataset with 6697 daytime images and 1015 query images (824 for the day and 191 for the night). The presence of large illumination and viewpoint changes makes it a challenging dataset to work with.

\textbf{InLoc}~\cite{taira2018inloc} is an indoor dataset that includes 9972 RGBD images; 329 RGB images from it are used as queries to test the performance of long-term indoor visual localization algorithms. This dataset provides a variety of challenges due to its large size (around 10k images covering two buildings), significant differences in viewpoint and/or illumination between the database and query images, and temporal changes in the scene. In addition to this, the InLoc dataset provides a large collection of depth maps from 3D scanners. 

\textbf{RobotCar-Seasons (RoCaS)}~\cite{maddern20171} is a challenging dataset that contains 26121 reference images and 11934 query images. The dataset presents a variety of environmental conditions, including rain, snow, dusk, winter, and inadequate lighting in suburban areas. These factors make the task of feature matching and visual localization difficult.

\textbf{LaMAR}~\cite{sarlin2022lamar} addresses the foundational technology of localization and mapping in augmented reality (AR), introducing a new benchmark for realistic AR scenarios. The dataset is captured using AR devices in diverse environments, including indoor and outdoor scenes with dynamic objects and varied lighting. It features multi-sensor data streams (images, depth, IMU, etc.) from devices like HoloLens 2 and iPhones/iPads, covering over 45,000 square meters. LaMAR's ground-truth pipeline aligns AR trajectories against laser scans automatically, robustly handling data from heterogeneous devices. This benchmark is pivotal in evaluating AR-specific localization and mapping methods, highlighting the importance of considering additional data streams like radio signals in AR devices. LaMAR offers a realistic and comprehensive dataset for AR, guiding future research directions in visual localization and mapping.

\subsection{Optical Flow Estimation Datasets}

\textbf{KITTI}~\cite{geiger2013vision} is an image matching dataset collected in urban traffic scenarios with both a 2012 and 2015 version. KITTI-2012 consists of 194 training image pairs and 195 test image pairs of resolution 1226×370, while KITTI-2015 contains 200 training image pairs and 200 test image pairs of resolution 1242×375. The dataset includes sparse ground truth disparities obtained using a laser scanner. The scenes in KITTI-2012 are relatively simple, whereas the KITTI-2015 dataset presents challenges due to its dynamic scenes and complex scenarios.

\subsection{Structure from Motion Datasets}

\textbf{SUN3D}~\cite{xiao2013sun3d} offers a comprehensive RGB-D video database, meticulously annotated to capture the full 3D extent of a variety of indoor environments. Encompassing 415 sequences across 254 spaces within 41 buildings, it provides a rich resource for SfM evaluations. Utilizing an ASUS Xtion PRO LIVE sensor, SUN3D simulates human visual perspectives, facilitating an innovative combination of intuitive labeling and generalized bundle adjustment to enhance semantic segmentation and camera pose accuracy. This methodology not only addresses reconstruction errors but also enriches the dataset with detailed 3D object models, point clouds, and semantic maps, making it an invaluable asset for advancing research in local feature matching, scene understanding, and 3D reconstruction.

\textbf{DTU}~\cite{aanaes2016large} serves as an exhaustive resource dedicated to advancing the field of multi-view stereo (MVS). Comprising 80 meticulously documented scenes from either 49 or 64 precise camera locations, under a spectrum of seven distinct lighting scenarios, the DTU dataset introduces a diverse array of challenges, encompassing specular reflections, textural disparities, and geometric intricacies. It offers a rigorously controlled environment, capturing images at a resolution of 1200 x 1600 pixels, and is supplemented by meticulously accurate structured light scans for benchmarking and evaluation purposes. This dataset is crucial for delving into the nuances of balancing precision and completeness in 3D reconstructions, navigating the complexities of mesh generation, and evaluating the influence of lighting conditions on MVS efficacy.

\textbf{Tanks and Temples}~\cite{knapitsch2017tanks} is crafted as a benchmarking tool for assessing image-based 3D reconstruction pipelines. It encompasses a diverse array of outdoor vistas and indoor settings captured under realistic conditions. The ground-truth data is acquired via an industrial laser scanner, furnishing high-resolution video sequences as input. The dataset is stratified into intermediate and advanced tiers, incorporating substantial vehicles, architectural structures, and intricate geometric configurations to challenge prevailing reconstruction pipelines. Consisting of 14 scenes spanning from singular objects to expansive indoor and outdoor landscapes, Tanks and Temples pushes the boundaries of 3D reconstruction with its meticulously curated ground-truth point clouds and camera poses extrapolated from 4K video sequences. This dataset serves as a linchpin for the refinement and evaluation of algorithms adept at traversing the intricacies inherent in large-scale scene reconstruction, providing an indispensable asset for enhancing the resilience and accuracy of image-based 3D reconstruction methodologies.

\textbf{ETH}~\cite{schonberger2017comparative} is a dataset designed to evaluate descriptors for SfM tasks by building a 3D model from a set of available 2D images. Following D2Net, three medium-sized datasets are evaluated: Madrid Metropolis, Gendarmenmarkt, and Tower of London~\cite{wilson2014robust}. ETH dataset includes various cameras and conditions, providing a challenging benchmark to compare different methods' performance.

\textbf{ETH3D}~\cite{schops2017multi} is a comprehensive benchmark for multi-view stereo algorithms. The dataset encompasses a wide array of scenes, both indoors and outdoors, captured through high-resolution DSLR cameras and synchronized low-resolution stereo videos. What makes this dataset distinctive is its combination of high spatial and temporal resolution. With scenarios ranging from natural to man-made environments, it introduces novel challenges for detailed 3D reconstruction, with a specific focus on the application of hand-held mobile devices in stereo vision scenarios. ETH3D provides diverse evaluation protocols catering to high-resolution multi-view stereo, low-resolution multi-view on video data, and two-view stereo. Consequently, it stands as a valuable asset for advancing research in the field of dense 3D reconstruction.

\textbf{ENRICH}~\cite{marelli2023enrich} presents a synthetic, versatile dataset tailored for the assessment and comparison of photogrammetric and computer vision techniques, with a focus on SfM and MVS methodologies. It encompasses three distinct subsets—ENRICH-Aerial, ENRICH-Square, and ENRICH-Statue—each offering a collection of high-resolution images rendered under various lighting conditions, camera orientations, scales, and viewpoints to closely mimic real-world complexity. Created from 3D models of real objects to ensure genuine textures and geometries, ENRICH employs a virtual camera modeled after the Nikon D750 DSLR to capture distortion-free images. Beyond serving as a rigorous platform for SfM and MVS algorithm evaluation, ENRICH provides an extensive array of ground truth data, including GCP coordinates, depth maps, and 3D models. This rich dataset, with its quality and scale diversity, from aerial views to ground-level scenes, stands as a critical asset for pushing the boundaries of research in remote sensing, photogrammetry, and computer vision, facilitating a broad spectrum of applications including image matching and monocular depth estimation.

\subsection{Dataset Gaps and Future Needs}

Although the above datasets provide valuable resources for evaluating local feature matching methods, there are significant gaps that need to be addressed. 

One major gap is the lack of datasets that simulate extreme environmental conditions. While the presence of datasets like RoCaS~\cite{maddern20171} offers some variability in environmental conditions, including diverse weather scenarios and lighting conditions, there is a need for datasets that focus specifically on challenging weather scenarios like heavy rain, fog, or snow. These conditions pose unique challenges in feature matching and are crucial for applications in climate-sensitive areas.
Another gap is the limited representation of highly dynamic environments. Current datasets, including the widely used HPatches~\cite{balntas2017hpatches}, while comprehensive in examining variations in viewpoint and illumination, do not adequately capture the complexity of crowded urban areas or fast-moving scenes. This limitation is significant for applications requiring real-time monitoring and surveillance in densely populated areas. Datasets that can mimic the dynamics of such environments are essential for advancing feature matching techniques in these contexts.
Additionally, there is a noticeable lack of datasets tailored for specific application domains, such as underwater or aerial imagery. These domains have unique characteristics and challenges that are not addressed by datasets like ETH~\cite{schonberger2017comparative} or Aachen Day-Night~\cite{sattler2012image}. Specialized datasets in these areas would be invaluable for research and development in fields like marine biology or drone-based monitoring.

In conclusion, while existing datasets have significantly contributed to the field of local feature matching, there is a clear need for more specialized datasets. These datasets should aim to fill the existing gaps and cater to the evolving needs of various application domains, thereby enabling further advancements in local feature matching techniques.

\section{Performance Review}

\subsection{Metrics For Matching Models}
\label{sec:Metrics For Matching Models}

\subsubsection{Image Matching}

\textbf{Repeatability}~\cite{schmid2000evaluation,joshi2020recent}. The repeatability metric for comparing two images is computed by taking the count of matching feature areas found between the images and dividing it by the lesser count of feature areas found within either of the images, subsequently multiplying by 100 to express the result as a percentage. This quantitative assessment is essential for gauging the consistency of feature detectors when subjected to different geometric alterations:

\[ \text{Repeatability} = \frac{M}{\min(F_1, F_2)} \times 100 \]

\noindent where, \( M \) denotes the number of matching feature areas between the two images, \( F_1 \) represents the total number of feature areas detected in the first image, and \( F_2 \) is the total number of feature areas detected in the second image.

\textbf{Matching score (M-score)}~\cite{detone2018superpoint,yi2016lift}. The M-Score quantifies the effectiveness of a feature detection and description pipeline by calculating the average ratio of correctly matched features to the total features detected in the overlapping areas of two images.
\textbf{Mean Matching Accuracy (MMA)}~\cite{dusmanu2019d2} is used to measure how well feature matchings are performed between image pairs considering multiple pixel error thresholds. It represents the average percentage of correct matches in image pairs considering several pixel error thresholds. The metric considers only mutually nearest neighbor matches, and matches are considered correct if the reprojection error, estimated using provided homography, is below the given matching threshold.
\textbf{Features} and \textbf{Matches}~\cite{dusmanu2019d2} evaluate the performance of feature descriptors. Features refers to the average number of detected features per image, and Matches indicates the average number of successful feature matches.
\textbf{Percentage of Correct Keypoints (PCK)}~\cite{wang2020learning} metric is commonly used to measure the performance of dense matching. It involves extracting key points from the first image on the image grid and finding their nearest neighbors in the complete second image. Predicted matchings of query points are considered correct if they fall within a certain pixel threshold of ground truth matching.

\subsubsection{Homography Estimation}

The angular correctness metric is typically used in order to evaluate the performance of feature matching algorithms. The metric involves estimating the homography transformation $\hat{\mathcal H}$ between two images and comparing the transformed corners with those computed using the ground-truth homography $\mathcal H$~\cite{detone2018superpoint}. To ensure a fair comparison among methods that produce different numbers of matches, a correctness identifier is calculated based on the corner error between the images warped with $\hat{\mathcal H}$ and $\mathcal H$. If the average error of the four corners is less than a specified pixel threshold $\varepsilon$, typically ranging from 1 to 10 pixels, the estimated homography is considered correct.
Once the correctness of the estimated homography is established, the angle errors between images are evaluated using the \textbf{Area Under Curve (AUC)} metric. This metric calculates the area under the error accumulation curve at various thresholds, quantifying the accuracy and stability of matching. The AUC value represents the overall matching performance, with higher values indicating better performance.

\subsubsection{Relative Pose Estimation}

When evaluating the estimated camera pose, the typical method involves measuring the angular deviations in rotations and translations~\cite{choy2016universal}. In this method, if the angle deviation is less than a certain threshold, rotation or translation is considered to be correctly estimated, and the average accuracy at that threshold is reported. 
The interval between frames is represented by $d_{frame}$, where larger values indicate more challenging image pairs for matching. 
For the pose error at different thresholds. The most common metrics include \textbf{AUC}, \textbf{matching accuracy}, \textbf{matching score}. Among them, the maximum values of the translation error and the angular error are usually noted as the pose error.

\subsubsection{Visual Localization}

The evaluation process typically follows the general evaluation protocol outlined in visual localization benchmarks\footnote{https://www.visuallocalization.net}. Custom features are used as input to the system, and then the image registration process is performed using a framework such as COLMAP~\cite{schonberger2016structure}. Finally, the percentage of images successfully positioned within a predefined tolerance range is calculated. In order to report the performance of the evaluated methods, the cumulative \textbf{AUC} of pose error at different threshold values is often used.

\subsubsection{Optical Flow Estimation}


Evaluation metrics employed for optical flow estimation encompass the \textbf{Average End-point Error (AEPE)}, \textbf{flow outlier ratio (Fl)}, and \textbf{Percentage of Correct Keypoints (PCK)}~\cite{melekhov2019dgc,truong2020glu}. 
AEPE is characterized as the mean Euclidean separation between the estimated and ground truth correspondence map. Specifically, it quantifies the Euclidean disparity between predicted and actual flow fields, computed as the average over valid pixels within the target image.
Fl assesses the average percentage of outliers across all pixels, where outliers are defined as flow errors exceeding either 3 pixels or \( 5\% \) of the ground truth flows.
PCK elucidates the percentage of appropriately matched estimated points $\hat{x}_i$ situated within a specified threshold (in pixels) from the corresponding ground truth points $x_i$.

\subsubsection{Structure from Motion}

As delineated in the evaluation framework prescribed by \textbf{ETH}~\cite{schonberger2017comparative}, a suite of pivotal metrics is employed to rigorously evaluate the fidelity of the reconstruction process. These encompass the \textbf{number of registered images}, which serves as an indicator of the reconstruction's comprehensiveness, along with the \textbf{sparse points} metric, which provides insights into the depth and intricacy of the scene's depiction. Furthermore, the \textbf{total observations in image} metric is pivotal for the calibration of cameras and the triangulation process, denoting the confirmed image projections of sparse points. The \textbf{mean feature track length}, indicative of the average count of verified image observations per sparse point, plays a vital role in ensuring precise calibration and robust triangulation. Lastly, the \textbf{mean reprojection error} is a critical measure for gauging the accuracy of the reconstruction, encapsulating the cumulative reprojection error observed in bundle adjustment and influenced by the thoroughness of input data as well as the precision of keypoint detection.

The key metrics in \textbf{ETH3D}~\cite{schops2017multi} are crucial for evaluating the effectiveness of various SfM methods. The AUC of pose error at different thresholds is used to assess the accuracy of multi-view camera pose estimation. This metric reflects the precision of estimated camera poses in relation to ground truth. Accuracy and Completeness percentages at different distance thresholds evaluate the 3D Triangulation task. Accuracy indicates the proportion of reconstructed points within a certain distance from the ground truth, and Completeness measures the percentage of ground truth points that are adequately represented within the reconstructed point cloud.

\subsection{Quantitative Performance}

In this section, we analyze the performance of several key methods in terms of the evaluation scores provided in Section \ref{sec:Metrics For Matching Models}, which includes various algorithms discussed earlier and additional methods. We compile their performances on popular benchmarks into tables, where the data is sourced either from the original authors or from the best reported results of other authors under the same evaluation conditions.
Furthermore, some publications may report performances on non-standard benchmarks/databases or only involve certain subsets of popular benchmark test sets. We do not present the performance of these methods.

The following tables provide summaries of several major DL-based matching models' performances on different datasets. Table \ref{tab:HPatches} highlights the HPatches~\cite{balntas2017hpatches} test set, adopting the evaluation protocol utilized by the LoFTR approach~\cite{sun2021loftr}. Performance metrics are based on the AUC of corner error distances up to 3, 5, and 10 pixels.
Table \ref{tab:ScanNet} focuses on the ScanNet~\cite{dai2017scannet} test set, following the SuperGlue~\cite{sarlin2020superglue} testing protocol. The reported metric is the pose AUC error.
Table \ref{tab:YFCC100M} centers on the YFCC100M~\cite{thomee2016yfcc100m} test set, with a protocol based on RANSAC-Flow~\cite{shen2020ransac}. Additionally, the pose mAP (mean Average Precision) value is reported. A pose estimation is considered an outlier if its maximum degree error in translation or rotation exceeds the threshold.
Table \ref{tab:MegaDepth} highlights the MegaDepth~\cite{li2018megadepth} test set. The pose estimation AUC error is reported, following the SuperGlue~\cite{sarlin2020superglue} evaluation methodology.
Tables \ref{tab:Aachen Day-Night_HLoc} and \ref{tab:Aachen Day-Night_Night} emphasize the Aachen Day-Night v1.0~\cite{sattler2012image} and v1.1~\cite{zhang2021reference} test sets, respectively, in the local feature evaluation track and the full visual localization track.
Table \ref{tab:InLoc} focuses on the InLoc~\cite{taira2018inloc} test set. The reported metrics include the percentage of correctly localized queries under specific error thresholds, following the HLoc~\cite{sarlin2019coarse} pipeline.
Table \ref{tab:KITTI} emphasizes the KITTI~\cite{geiger2013vision} test set. The AEPE and the flow outlier ratio Fl are reported for both the 2012 and 2015 versions of the KITTI dataset.
Table \ref{tab:SFM} focuses on the ETH3D~\cite{schops2017multi}, presenting a detailed evaluation of various SfM methods as reported in the DetectorFreeSfM~\cite{he2023detector}. This evaluation thoroughly examines the effectiveness of these methods across three crucial metrics: AUC, Accuracy, and Completeness.

\begin{table}[h!]
\centering
\footnotesize
\caption{Evaluation on HPatches ~\cite{balntas2017hpatches} for homography estimation. We compare with two groups of the methods, Detector-based and Detector-free methods.}
\begin{tabular}{cccccc@{\hspace{0pt}}c} 
\specialrule{1pt}{0pt}{0pt}

\multirow{2}{*}{Category}       & \multirow{2}{*}{Methods} & \multicolumn{3}{c}{Pose Estimation AUC↑} & \multirow{2}{*}{matches} \\ \cline{3-5}
                                &                          & @3px        & @5px        & @10px        &                          \\ \hline
\multirow{5}{*}{Detector-based} & D2Net~\cite{dusmanu2019d2}+NN                 & 23.2        & 35.9        & 53.6         & 0.2K                     \\
                                & R2D2~\cite{revaud2019r2d2}+NN                  & 50.6        & 63.9        & 76.8         & 0.5K                     \\
                                & DISK~\cite{tyszkiewicz2020disk}+NN                  & 52.3        & 64.9        & 78.9         & 1.1K                     \\
                                & SP+GFM~\cite{nan2022learning}                   & 51.9        & 65.8        & 79.1         & 2.0k                     \\
                                & SP+SuperGlue~\cite{sarlin2020superglue}             & \textbf{53.9}        & \textbf{68.3}        & \textbf{81.7}         & 0.6K                     \\ \hline
\multirow{15}{*}{Detector-free} & COTR~\cite{jiang2021cotr}                     & 41.9        & 57.7        & 74.0         & 1.0K                     \\
                                & Sparse-NCNet~\cite{rocco2020efficient}             & 48.9        & 54.2        & 67.1         & 1.0K                     \\
                                & DualRC-Net~\cite{li2020dual}               & 50.6        & 56.2        & 68.3         & 1.0K                     \\
                                & Patch2Pix~\cite{zhou2021patch2pix}                 & 59.3        & 70.6        & 81.2         & 0.7K                     \\
                                & 3DG-STFM~\cite{mao20223dg}                 & 64.7        & 73.1        & 81.0         & 1.0k                     \\
                                & LoFTR~\cite{sun2021loftr}                    & 65.9        & 75.6        & 84.6         & 1.0K                     \\
                                & SE2-LoFTR~\cite{bokman2022case}                    & 66.2        & 76.6        & 86.0         & —                     \\
                                & QuadTree~\cite{tang2022quadtree}                 & 66.3        & 76.2        & 84.9         & 2.7k                     \\
                                & PDC-Net+~\cite{truong2023pdc}                 & 66.7        & 76.8        & 85.8         & 1.0k                     \\
                                & TopicFM~\cite{giang2022topicfm}                  & 67.3        & 77.0        & 85.7         & 1.0K                     \\
                                & ASpanFormer~\cite{chen2022aspanformer}              & 67.4        & 76.9        & 85.6         & —                        \\
                                & SEM~\cite{chang2023structured}                      & 69.6        & 79.0        & 87.1         & 1.0K                     \\
                                & CasMTR-2c~\cite{cao2023improving}                & 71.4        & 80.2        & 87.9         & 0.5k                     \\
                                & DKM~\cite{edstedt2023dkm}                      & 71.3        & 80.6        & 88.5         & 5.0K                     \\
                                & ASTR~\cite{yu2023adaptive}                     & 71.7        & 80.3        & 88.0         & 1.0K                     \\
                                & PMatch~\cite{zhu2023pmatch}                   & 71.9        & 80.7        & 88.5         & —                        \\
                                & RoMa~\cite{edstedt2023roma}                   & \textbf{72.2}        & \textbf{81.2}        & \textbf{89.1}         & —                        \\

\specialrule{1pt}{0pt}{0pt}
\end{tabular}
\label{tab:HPatches}
\end{table}

\begin{table}[h!]
\centering
\footnotesize
\caption{ScanNet~\cite{dai2017scannet} Two-View Camera Pose Estimation. We compare with two groups of the methods, Detector-based and Detector-free methods.}
\begin{tabular}{ccccc}
\specialrule{1pt}{0pt}{0pt}

\multirow{2}{*}{Category}        & \multirow{2}{*}{Methods} & \multicolumn{3}{c}{Pose Estimation AUC↑} \\ \cline{3-5} 
                                 &                          & @5°          & @10°        & @20°        \\ \hline
\multirow{11}{*}{Detector-based} & ORB+GMS~\cite{bian2017gms}                  & 5.2          & 13.7        & 25.4        \\
                                 & D2Net~\cite{dusmanu2019d2}+NN                 & 5.3          & 14.5        & 28.0        \\
                                 & ContextDesc+RT~\cite{luo2019contextdesc}           & 6.6          & 15.0        & 25.8        \\
                                 & ContextDesc+NN~\cite{luo2019contextdesc}           & 9.4          & 21.5        & 36.4        \\
                                 & SP+NN~\cite{detone2018superpoint}                    & 9.4          & 21.5        & 36.4        \\
                                 & SP+PointCN~\cite{yi2018learning}               & 11.4         & 25.5        & 41.4        \\
                                 & SP+HTMatch~\cite{cai2023htmatch}               & 15.1         & 31.4        & 48.2        \\
                                 & SP+SGMNet~\cite{chen2021learning}                & 15.4         & 32.1        & 48.3        \\
                                 & ContextDesc+SGMNet~\cite{chen2021learning}       & 15.4         & 32.3        & 48.8        \\
                                 & SP+SuperGlue~\cite{sarlin2020superglue}             & 16.2         & 33.8        & 51.8        \\
                                 & SP+DenseGAP~\cite{kuang2022densegap}              & \textbf{17.0}         & \textbf{36.1}        & \textbf{55.7}        \\ \hline
\multirow{16}{*}{Detector-free}  & DualRC-Net~\cite{li2020dual}               & 7.7          & 17.9        & 30.5        \\
                                 & SEM~\cite{chang2023structured}                      & 18.7         & 36.6        & 52.9        \\
                                 & PDC-Net(H)~\cite{truong2021learning}               & 18.7         & 37.0        & 54.0        \\
                                 & PDC-Net+(H)~\cite{truong2023pdc}              & 20.3         & 39.4        & 57.1        \\
                                 & LoFTR-DT~\cite{sun2021loftr}                 & 22.1         & 40.8        & 57.6        \\
                                 & 3DG-STFM~\cite{mao20223dg}                 & 23.6         & 43.6        & 61.2        \\
                                 & LoFTR~\cite{sun2021loftr}+QuadTree~\cite{tang2022quadtree}           & 23.9         & 43.2        & 60.3        \\
                                 & MatchFormer~\cite{wang2022matchformer}              & 24.3         & 43.9        & 61.4        \\
                                 & QuadTree~\cite{tang2022quadtree}                 & 24.9         & 44.7        & 61.8        \\
                                 & ASpanFormer~\cite{chen2022aspanformer}              & 25.6         & 46.0        & 63.3        \\
                                 & OAMatcher~\cite{dai2023oamatcher}                & 26.1         & 45.3        & 62.1        \\
                                 & PATS~\cite{ni2023pats}                     & 26.0         & 46.9        & 64.3        \\
                                 & CasMTR-4c~\cite{cao2023improving}                & 27.1         & 47.0        & 64.4        \\
                                 & DeepMatcher-L~\cite{xie2023deepmatcher}            & 27.3         & 46.3        & 62.5        \\
                                 & PMatch~\cite{zhu2023pmatch}                   & 29.4         & 50.1        & 67.4       \\
                                 & DKM~\cite{edstedt2023dkm}                      & 29.4         & 50.7        & 68.3        \\
                                 & RoMa~\cite{edstedt2023roma}                      & \textbf{31.8}         & \textbf{53.4}        & \textbf{70.9}        \\

\specialrule{1pt}{0pt}{0pt}
\end{tabular}
\label{tab:ScanNet}
\end{table}

\begin{table*}[]
\centering
\footnotesize
\caption{Evaluation on YFCC100M ~\cite{thomee2016yfcc100m} for outdoor pose estimation. We compare with two groups of the methods, Detector-based and Detector-free methods.}
\begin{tabular*}{\linewidth}{@{\extracolsep{\fill}}cccccccc}
\specialrule{1pt}{0pt}{0pt}

\multirow{2}{*}{Category}        & \multirow{2}{*}{Methods} & \multicolumn{3}{c}{Pose estimation AUC↑} & \multicolumn{3}{c}{Pose estimation mAP↑} \\ \cline{3-8} 
                                 &                          & @5°          & @10°        & @20°        & @5°          & @10°        & @20°        \\ \hline
\multirow{10}{*}{Detector-based} & SuperPoint(SP)~\cite{detone2018superpoint}           & —            & —           & —           & 30.5         & 50.8        & 67.9        \\
                                 & SIFT~\cite{liu2010sift}+RT                  & 24.1         & 40.7        & 58.1        & 45.1         & 55.8        & 67.2        \\
                                 & SP+OANet~\cite{zhang2019learning}                 & 26.8         & 45.0        & 62.2        & 50.9         & 61.4        & 71.8        \\
                                 & SIFT+OANet~\cite{zhang2019learning}               & 29.2         & 48.1        & 65.1        & 55.1         & 65.0        & 74.8        \\
                                 & CoAM~\cite{wiles2021co}                     & —            & —           & —           & 55.6         & 66.8        & —           \\
                                 & SIFT+SuperGlue~\cite{sarlin2020superglue}           & 30.5         & 51.3        & 69.7        & 59.3         & 70.4        & 80.4        \\
                                 & Paraformer~\cite{lu2023paraformer}               & 31.7         & 52.3        & 70.4        & 60.1         & 70.7        & 78.1        \\
                                 & RootSIFT~\cite{arandjelovic2012three}+SGMNet~\cite{chen2021learning}          & 35.5         & 55.2        & 71.9        & —            & —           & —           \\
                                 & SP+SuperGlue~\cite{sarlin2020superglue}             & \textbf{38.7}         & \textbf{59.1}        & \textbf{75.8}        & \textbf{67.8}         & \textbf{77.4}        & \textbf{85.7}        \\ \hline
\multirow{9}{*}{Detector-free}   & DualRC-Net~\cite{li2020dual}               & 29.5         & 50.1        & 66.8        & —            & —           & —           \\
                                 & RANSAC-Flow~\cite{shen2020ransac}              & —            & —           & —           & 64.9         & 73.3        & 81.6        \\
                                 & PDC-Net(MS)~\cite{truong2021learning}              & 35.7         & 55.8        & 72.3        & 63.9         & 73.0        & 73.0        \\
                                 & PDC-Net+(H)~\cite{truong2023pdc}              & 37.5         & 58.1        & 74.5        & 67.4         & 76.6        & 84.6        \\
                                 & LoFTR~\cite{sun2021loftr}                    & 42.4         & 62.5        & 77.3        & —            & —           & —           \\
                                 & ASpanFormer~\cite{chen2022aspanformer}              & 44.5         & 63.8        & 78.4        & —            & —           & —           \\
                                 & PMatch~\cite{zhu2023pmatch}                   & 45.7         & 65.2        & \textbf{79.8}        & \textbf{75.9}         & \textbf{83.1}        & \textbf{89.3}        \\
                                 & PATS~\cite{ni2023pats}                     & \textbf{47.0}         & \textbf{65.3}        & 79.2        & —            & —           & —           \\

\specialrule{1pt}{0pt}{0pt}
\end{tabular*}
\label{tab:YFCC100M}
\end{table*}

\begin{table}[h!]
\centering
\footnotesize
\caption{MegaDepth~\cite{li2018megadepth} Two-View Camera Pose Estimation. We compare with two groups of the methods, Detector-based and Detector-free methods.}
\begin{tabular}{ccccc}
\specialrule{1pt}{0pt}{0pt}

\multirow{2}{*}{Category}       & \multirow{2}{*}{Methods} & \multicolumn{3}{c}{Pose Estimation AUC↑} \\ \cline{3-5} 
                                &                          & @5°          & @10°        & @20°        \\ \hline
\multirow{4}{*}{Detector-based} & SP+SGMNet~\cite{chen2021learning}                & 40.5         & 59.0        & 72.6        \\
                                & SP+DenseGAP~\cite{kuang2022densegap}              & 41.2         & 56.9        & 70.2        \\
                                & SP+SuperGlue~\cite{sarlin2020superglue}             & 42.2         & \textbf{61.2}        & \textbf{75.9}        \\
                                & SP+ClusterGNN~\cite{shi2022clustergnn}            & \textbf{44.2}         & 58.5        & 70.3        \\ \hline
\multirow{16}{*}{Detector-free} & Patch2Pix~\cite{zhou2021patch2pix}                 & 41.4         & 56.3        & 68.3        \\
                                & ECO-TR~\cite{tan2022eco}                   & 48.3         & 65.8        & 78.5        \\
                                & PDC-Net+~\cite{truong2023pdc}                 & 51.5         & 67.2        & 78.5        \\
                                & 3DG-STFM~\cite{mao20223dg}                 & 52.6         & 68.5        & 80.0        \\
                                & SE2-LoFTR~\cite{bokman2022case}                    & 52.6         & 69.2        & 81.4        \\
                                & LoFTR~\cite{sun2021loftr}                    & 52.8         & 69.2        & 81.2        \\
                                & MatchFormer~\cite{wang2022matchformer}              & 52.9         & 69.7        & 82.0        \\
                                & TopicFM~\cite{giang2022topicfm}                  & 54.1         & 70.1        & 81.6        \\
                                & QuadTree~\cite{tang2022quadtree}                 & 54.6         & 70.5        & 82.2        \\
                                & ASpanFormer~\cite{chen2022aspanformer}              & 55.3         & 71.5        & 83.1        \\
                                & OAMatcher~\cite{dai2023oamatcher}                & 56.6         & 72.3        & 83.6        \\
                                & DeepMatcher-L~\cite{xie2023deepmatcher}            & 57.0         & 73.1        & 84.2        \\
                                & SEM~\cite{chang2023structured}                      & 58.0         & 72.9        & 83.7        \\
                                & ASTR~\cite{yu2023adaptive}                     & 58.4         & 73.1        & 83.8        \\
                                & CasMTR-2c~\cite{cao2023improving}                & 59.1         & 74.3        & 84.8        \\
                                & DKM~\cite{edstedt2023dkm}                      & 60.4         & 74.9        & 85.1        \\
                                & PMatch~\cite{zhu2023pmatch}                   & 61.4         & 75.7        & 85.7        \\
                                & RoMa~\cite{edstedt2023roma}                   & \textbf{62.6}         & \textbf{76.7}        & \textbf{86.3}        \\

\specialrule{1pt}{0pt}{0pt}
\end{tabular}
\label{tab:MegaDepth}
\end{table}

\begin{table*}[h!]
\centering
\footnotesize
\caption{Visual localization evaluation on the Aachen Day-Night benchmark v1.0~\cite{sattler2012image} and v1.1~\cite{zhang2021reference}. The evaluation results on the full visual localization track are reported. We compare with two groups of the methods, Detector-based and Detector-free methods.}
\begin{tabular*}{\linewidth}{@{\extracolsep{\fill}}ccccccccc}
\specialrule{1pt}{0pt}{0pt}

\multirow{2}{*}{Dataset}                & \multirow{2}{*}{Category}        & \multirow{2}{*}{Methods} & \multicolumn{3}{c}{Day}             & \multicolumn{3}{c}{Night}           \\ \cline{4-9} 
                                        &                                  &                          & (0.25m,2°) & (0.5m,5°) & (1.0m,10°) & (0.25m,2°) & (0.5m,5°) & (1.0m,10°) \\ \hline
\multirow{11}{*}{v1.0} & \multirow{10}{*}{Detector-based} & SP+NN~\cite{detone2018superpoint}                    & 85.4       & 93.3      & 97.2       & 75.5       & 86.7      & 92.9       \\
                                        &                                  & SP+CAPS~\cite{wang2020learning}+NN               & 86.3       & 93.0      & 95.9       & 83.7       & 90.8      & 96.9       \\
                                        &                                  & SP+SuperGlue~\cite{sarlin2020superglue}             & \textbf{89.6}       & 95.4      & \textbf{98.8}       & 86.7       & 93.9      & \textbf{100.0}      \\
                                        &                                  & SP+SGMNet~\cite{chen2021learning}                & 86.8       & 94.2      & 97.7       & 83.7       & 91.8      & 99.0       \\
                                        &                                  & SP+ClusterGNN~\cite{shi2022clustergnn}            & 89.4       & \textbf{95.5}      & 98.5       & 81.6       & 93.9      & \textbf{100.0}      \\
                                        &                                  & SP+LightGlue~\cite{lindenberger2023lightglue}             & 89.2       & 95.4      & 98.5       & \textbf{87.8}       & 93.9      & \textbf{100.0}      \\
                                        &                                  & ASLFeat~\cite{luo2020aslfeat}+NN               & 82.3       & 89.2      & 92.7       & 67.3       & 79.6      & 85.7       \\
                                        &                                  & ASLFeat~\cite{luo2020aslfeat}+SGMNet~\cite{chen2021learning}           & 86.8       & 93.4      & 97.1       & 86.7       & \textbf{94.9}      & 98.0       \\
                                        &                                  & ASLFeat~\cite{luo2020aslfeat}+SuperGlue~\cite{sarlin2020superglue}        & 87.9       & 95.4      & 98.3       & 81.6       & 91.8      & 99.0       \\
                                        &                                  & ASLFeat~\cite{luo2020aslfeat}+ClusterGNN~\cite{shi2022clustergnn}       & 88.6       & 95.5      & 98.4       & 85.7       & 93.9      & 99.0       \\ \cline{2-9} 
                                        & Detector-free                    & Patch2Pix~\cite{zhou2021patch2pix}                 & \textbf{84.6}       & \textbf{92.1}      & \textbf{96.5}       & \textbf{82.7}       & \textbf{92.9}      & \textbf{99.0}       \\ \hline
\multirow{16}{*}{v1.1} & \multirow{7}{*}{Detector-based}  & ISRF~\cite{melekhov2020image}                     & 87.1       & 94.7      & 98.3       & 74.3       & 86.9      & 97.4       \\
                                        &                                  & Rlocs~\cite{zhou2021retrieval}                    & 88.8       & 95.4      & 99.0       & 74.3       & 90.6      & 98.4       \\
                                        &                                  & KAPTURE+R2D2+APGeM~\cite{humenberger2020robust}       & 90.0       & 96.2      & \textbf{99.5}       & 72.3       & 86.4      & 97.9       \\
                                        &                                  & SP+SuperGlue~\cite{sarlin2020superglue}             & 89.8       & 96.1      & 99.4       & 77.0       & 90.6      & \textbf{100.0}      \\
                                        &                                  & SP+SuperGlue~\cite{sarlin2020superglue}+Patch2Pix~\cite{zhou2021patch2pix}    & 89.3       & 95.8      & 99.2       & \textbf{78.0}       & 90.6      & 99.0       \\
                                        &                                  & SP+GFM~\cite{nan2022learning}                   & \textbf{90.2}       & \textbf{96.4}      & \textbf{99.5}       & 74.0       & \textbf{91.5}      & 99.5       \\
                                        &                                  & SP+LightGlue~\cite{lindenberger2023lightglue}             & \textbf{90.2}       & 96.0      & 99.4       & 77.0       & 91.1      & \textbf{100.0}      \\ \cline{2-9} 
                                        & \multirow{9}{*}{Detector-free}   & Patch2Pix~\cite{zhou2021patch2pix}                 & 86.4       & 93.0      & 97.5       & 72.3       & 88.5      & 97.9       \\
                                        &                                  & LoFTR-DS~\cite{sun2021loftr}                 & —          & —         & —          & 72.8       & 88.5      & 99.0       \\
                                        &                                  & LoFTR-OT~\cite{sun2021loftr}                 & 88.7       & 95.6      & 99.0       & 78.5       & 90.6      & 99.0       \\
                                        &                                  & ASpanFormer~\cite{chen2022aspanformer}              & 89.4       & 95.6      & 99.0       & 77.5       & 91.6      & \textbf{99.5}       \\
                                        &                                  & AdaMatcher-LoFTR~\cite{huang2023adaptive}         & 89.2       & 96.0      & \textbf{99.3}       & \textbf{79.1}       & 90.6      & \textbf{99.5}       \\
                                        &                                  & AdaMatcher-Quad~\cite{huang2023adaptive}          & 89.2       & 95.9      & 99.2       & \textbf{79.1}       & \textbf{92.1}      & \textbf{99.5}       \\
                                        &                                  & ASTR~\cite{yu2023adaptive}                     & 89.9       & 95.6      & 99.2       & 76.4       & \textbf{92.1}      & \textbf{99.5}      \\
                                        &                                  & TopicFM~\cite{giang2022topicfm}                  & 90.2       & 95.9      & 98.9       & 77.5       & 91.1      & \textbf{99.5}       \\
                                        &                                  & CasMTR~\cite{cao2023improving}                   & \textbf{90.4}       & \textbf{96.2}      & \textbf{99.3}       & 78.5       & 91.6      & \textbf{99.5}       \\
                                        
\specialrule{1pt}{0pt}{0pt}
\end{tabular*}
\label{tab:Aachen Day-Night_HLoc}
\end{table*}

\begin{table*}[h!]
\centering
\footnotesize
\caption{Visual localization evaluation on the Aachen Day-Night benchmark v1.0~\cite{sattler2012image} and v1.1~\cite{zhang2021reference}. The evaluation results on  the local feature evaluation track are reported. We compare with two groups of the methods, Detector-based and Detector-free methods.}
\begin{tabular*}{\linewidth}{@{\extracolsep{\fill}}cccccccc}
\specialrule{1pt}{0pt}{0pt}

\multirow{2}{*}{Category}        & \multirow{2}{*}{Method} & \multicolumn{3}{c}{Aachen Day-Night v1.0} & \multicolumn{3}{c}{Aachen Day-Night v1.1} \\ \cline{3-8} 
                                 &                         & (0.5m,2°)     & (1m,5°)    & (5m, 10°)    & (0.5m,2°)     & (1m,5°)    & (5m, 10°)    \\ \hline
\multirow{15}{*}{Detector-based} & SP~\cite{detone2018superpoint}                      & 74.5          & 78.6       & 89.8         & —             & —          & —            \\
                                 & D2Net~\cite{dusmanu2019d2}                  & 74.5          & 86.7       & \textbf{100.0}        & —             & —          & —            \\
                                 & R2D2~\cite{revaud2019r2d2}(K=20k)             & 76.5          & \textbf{90.8}       & \textbf{100.0}        & 71.2          & 86.9       & 97.9         \\
                                 & ASLFeat~\cite{luo2020aslfeat}                 & 81.6          & 87.8       & \textbf{100.0}        & \textbf{75.4}          & 75.4       & 97.6         \\
                                 & ISRF~\cite{melekhov2020image}                    & —             & —          & —            & 69.1          & 87.4       & 98.4         \\
                                 & PoSFeat~\cite{li2022decoupling}                 & 81.6          & \textbf{90.8}       & \textbf{100.0}        & 73.8          & 87.4       & 98.4         \\
                                 & SIFT+CAPS~\cite{wang2020learning}               & 77.6          & 86.7       & 99.0         & —             & —          & —            \\
                                 & SP+S2DNet~\cite{germain2020s2dnet}               & 74.5          & 84.7       & \textbf{100.0}        & —             & —          & —            \\
                                 & SP+CAPS~\cite{wang2020learning}                 & \textbf{82.7}          & 87.8       & \textbf{100.0}        & —             & —          & —            \\
                                 & SP+OANet~\cite{zhang2019learning}                & 77.6          & 86.7       & 98.0         & —             & —          & —            \\
                                 & SP+SGMNet~\cite{chen2021learning}               & 77.6          & 88.8       & 99.0         & 72.3          & 85.3       & 97.9         \\
                                 & SP+SuperGlue~\cite{sarlin2020superglue}            & 79.6          & \textbf{90.8}       & \textbf{100.0}        & 73.3          & \textbf{88.0}       & 98.4         \\
                                 & DSD-MatchingNet~\cite{zhao2022dsd}         & 80.1          & 90.3       & \textbf{100.0}        & 73.0          & \textbf{88.0}       & \textbf{99.3}         \\ \hline
\multirow{6}{*}{Detector-free}   & Patch2Pix~\cite{zhou2021patch2pix}               & 79.6          & 87.8       & \textbf{100.0}        & 72.3          & \textbf{88.5}       & 97.9         \\
                                 & Sparse-NCNet~\cite{rocco2020efficient}            & 76.5          & 84.7       & 98.0         & —             & —          & —            \\
                                 & DualRC-Net~\cite{li2020dual}              & 79.6          & 88.8       & \textbf{100.0}        & 71.2          & 86.9       & 97.9         \\
                                 & PDC-Net~\cite{truong2021learning}                 & 76.5          & 85.7       & \textbf{100.0}        & —             & —          & —            \\
                                 & PDC-Net+~\cite{truong2023pdc}                & \textbf{80.6}          & \textbf{89.8}       & \textbf{100.0}        & —             & —          & —            \\
                                 & LoFTR~\cite{sun2021loftr}                   & —             & —          & —            & \textbf{72.8}          & \textbf{88.5}       & \textbf{99.0}         \\

\specialrule{1pt}{0pt}{0pt}
\end{tabular*}
\label{tab:Aachen Day-Night_Night}
\end{table*}

\begin{table*}[]
\centering
\footnotesize
\caption{Visual Localization on the InLoc benchmark ~\cite{taira2018inloc}. We compare with two groups of the methods, Detector-based and Detector-free methods.}
\begin{tabular*}{\linewidth}{@{\extracolsep{\fill}}cccccccc}
\specialrule{1pt}{0pt}{0pt}

\multirow{2}{*}{Category}        & \multirow{2}{*}{Methods} & \multicolumn{3}{c}{DUC1}               & \multicolumn{3}{c}{DUC2}               \\ \cline{3-8} 
                                 &                          & (0.25m, 10°) & (0.5m, 10°) & (1m, 10°) & (0.25m, 10°) & (0.5m, 10°) & (1m, 10°) \\ \hline
\multirow{15}{*}{Detector-based} & SIFT+CAPS~\cite{wang2020learning}+NN             & 38.4         & 56.6        & 70.7      & 35.1         & 48.9        & 58.8      \\
                                 & ISRF~\cite{melekhov2020image}                     & 39.4         & 58.1        & 70.2      & 41.2         & 61.1        & 69.5      \\
                                 & D2Net~\cite{dusmanu2019d2}+NN                 & 38.4         & 56.1        & 71.2      & 37.4         & 55.0        & 64.9      \\
                                 & R2D2~\cite{revaud2019r2d2}+NN                  & 36.4         & 57.6        & 74.2      & 45.0         & 60.3        & 67.9      \\
                                 & KAPTURE~\cite{humenberger2020robust}+R2D2~\cite{revaud2019r2d2}             & 41.4         & 60.1        & 73.7      & 47.3         & 67.2        & 73.3      \\
                                 & SeLF~\cite{fan2022learning}+NN                  & 41.4         & 61.6        & 73.2      & 44.3         & 61.1        & 68.7      \\
                                 & AWDesc~\cite{wang2023attention}+NN                  & 41.9         & 61.6        & 72.2      & 45.0         & 61.1        & 70.2      \\
                                 & ASLFeat~\cite{luo2020aslfeat}+NN               & 39.9         & 59.1        & 71.7      & 43.5         & 58.8        & 64.9      \\
                                 & ASLFeat~\cite{luo2020aslfeat}+SGMNet~\cite{chen2021learning}           & 43.9         & 62.1        & 68.2      & 45.0         & 63.4        & 73.3      \\
                                 & ASLFeat~\cite{luo2020aslfeat}+SuperGlue~\cite{sarlin2020superglue}        & 51.5         & 66.7        & 75.8      & 53.4         & 76.3        & 84.0      \\
                                 & ASLFeat~\cite{luo2020aslfeat}+ClusterGNN~\cite{shi2022clustergnn}       & \textbf{52.5}         & 68.7        & 76.8      & \textbf{55.0}         & 76.0        & 82.4      \\
                                 & SP+NN~\cite{detone2018superpoint}                    & 40.4         & 58.1        & 69.7      & 42.0         & 58.8        & 69.5      \\
                                 & SP+ClusterGNN~\cite{shi2022clustergnn}            & 47.5         & \textbf{69.7}        & 79.8      & 53.4         & \textbf{77.1}        & \textbf{84.7}      \\
                                 & SP+SuperGlue~\cite{sarlin2020superglue}             & 49.0         & 68.7        & \textbf{80.8}      & 53.4         & \textbf{77.1}        & 82.4      \\
                                 & SP+CAPS~\cite{wang2020learning}+NN               & 40.9         & 60.6        & 72.7      & 43.5         & 58.8        & 68.7      \\
                                 & SP+LightGlue~\cite{lindenberger2023lightglue}             & 49.0         & 68.2        & 79.3      & \textbf{55.0}         & 74.8        & 79.4      \\ \hline
\multirow{14}{*}{Detector-free}  & Sparse-NCNet~\cite{rocco2020efficient}             & 41.9         & 62.1        & 72.7      & 35.1         & 48.1        & 55.0      \\
                                 & MTLDesc~\cite{wang2022mtldesc}                  & 41.9         & 61.6        & 72.2      & 45.0         & 61.1        & 70.2      \\
                                 & COTR~\cite{jiang2021cotr}                     & 41.9         & 61.1        & 73.2      & 42.7         & 67.9        & 75.6      \\
                                 & Patch2Pix~\cite{zhou2021patch2pix}                 & 44.4         & 66.7        & 78.3      & 49.6         & 64.9        & 72.5      \\
                                 & LoFTR-OT~\cite{sun2021loftr}                 & 47.5         & 72.2        & 84.8      & 54.2         & 74.8        & 85.5      \\
                                 & MatchFormer~\cite{wang2022matchformer}              & 46.5         & 73.2        & 85.9      & 55.7         & 71.8        & 81.7      \\
                                 & ASpanFormer~\cite{chen2022aspanformer}              & 51.5         & 73.7        & 86.4      & 55.0         & 74.0        & 81.7      \\
                                 & TopicFM~\cite{giang2022topicfm}                  & 52.0         & 74.7        & \textbf{87.4}      & 53.4         & 74.8        & 83.2      \\
                                 & GlueStick~\cite{pautrat2023gluestick}                & 49.0         & 70.2        & 84.3      & 55.0         & \textbf{83.2}        & \textbf{87.0}      \\
                                 & SEM~\cite{chang2023structured}                      & 52.0         & 74.2        & \textbf{87.4}      & 50.4         & 76.3        & 83.2      \\
                                 & ASTR~\cite{yu2023adaptive}                     & 53.0         & 73.7        & \textbf{87.4}      & 52.7         & 76.3        & 84.0      \\
                                 & CasMTR~\cite{cao2023improving}                   & 53.5         & \textbf{76.8}        & 85.4      & 51.9         & 70.2        & 83.2      \\
                                 & ASTR~\cite{yu2023adaptive}                     & 53.0         & 73.7        & \textbf{87.4}      & 52.7         & 76.3        & 84.0      \\
                                 & PATS~\cite{ni2023pats}                     & \textbf{55.6}         & 71.2        & 81.0      & \textbf{58.8}         & 80.9        & 85.5      \\

\specialrule{1pt}{0pt}{0pt}
\end{tabular*}
\label{tab:InLoc}
\end{table*}

\begin{table}[h!]
\centering
\footnotesize
\caption{Optical flow results on the training splits of KITTI ~\cite{geiger2013vision}. 
Following ~\cite{jiang2021cotr,tan2022eco}, ("+intp") represents interpolating the output of the model to obtain the corresponding relationship per pixel.
† means evaluated it with DenseMatching tools provided by the authors of GLU-Net. 
This part contains generic matching networks.}
\begin{tabular}{ccccc}
\specialrule{1pt}{0pt}{0pt}

\multirow{2}{*}{Methods} & \multicolumn{2}{c}{KITTI-2012}                                  & \multicolumn{2}{c}{KITTI-2015}                                  \\ \cline{2-5} 
                         & \multicolumn{1}{l}{APAE↓} & \multicolumn{1}{l}{Fl-all{[}\%{]}↓} & \multicolumn{1}{l}{APAE↓} & \multicolumn{1}{l}{Fl-all{[}\%{]}↓} \\ \hline
DGC-Net~\cite{melekhov2019dgc}                  & 8.50                      & 32.28                               & 14.97                     & 50.98                               \\
DGC-Net†~\cite{melekhov2019dgc}                 & 7.96                      & 34.35                               & 14.33                     & 50.35                               \\
GLU-Net~\cite{truong2020glu}                  & 3.14                      & 19.76                               & 7.49                      & 33.83                               \\
GLU-Net+GOCor~\cite{truong2020gocor}            & 2.68                      & 15.43                               & 6.68                      & 27.57                               \\
RANSAC-Flow~\cite{shen2020ransac}              & —                         & —                                   & 12.48                     & —                                   \\
COTR~\cite{jiang2021cotr}                     & 1.28                      & 7.36                                & 2.62                      & 9.92                                \\
COTR+Intp.~\cite{jiang2021cotr}               & 2.26                      & 10.50                               & 6.12                      & 16.90                               \\
PDC-Net(D)~\cite{truong2021learning}              & 2.08                      & 7.98                                & 5.22                      & 15.13                               \\
PDC-Net+(D)~\cite{truong2023pdc}              & 1.76                      & 6.60                                & 4.53                      & 12.62                               \\
COTR†~\cite{jiang2021cotr}                    & 1.15                      & 6.98                                & 2.06                      & 9.14                                \\
COTR†+Intp.~\cite{jiang2021cotr}              & 1.47                      & 8.79                                & 3.65                      & 13.65                               \\
ECO-TR~\cite{tan2022eco}                   & \textbf{0.96}                      & \textbf{3.77}                                & \textbf{1.40}                      & \textbf{6.39}                                \\
ECO-TR+Intp.~\cite{tan2022eco}             & 1.46                      & 6.64                                & 3.16                      & 12.10                               \\
PATS+Intp.~\cite{ni2023pats}               & 1.17                      & 4.04                                & 3.39                      & 9.68                                \\

\specialrule{1pt}{0pt}{0pt}
\end{tabular}
\label{tab:KITTI}
\end{table}

\begin{table*}[]
\centering
\footnotesize
\caption{Evaluation of SfM Methods on the ETH3D~\cite{schops2017multi} for Multi-View Camera Pose Estimation and 3D Triangulation. The table segregates methods into Detector-based and Detector-free categories. The results are derived from the DetectorFreeSfM~\cite{he2023detector}.}
\begin{tabular*}{\linewidth}{@{\extracolsep{\fill}}ccccccccccc}
\specialrule{1pt}{0pt}{0pt}

\multirow{3}{*}{Category}       & \multirow{3}{*}{Methods}         & \multicolumn{3}{c}{Multi-View Camera Pose   Estimation} & \multicolumn{6}{c}{3D Triangulation}                                      \\
                                &                                  & \multicolumn{3}{c}{AUC}                                 & \multicolumn{3}{c}{Accuracy (\%)} & \multicolumn{3}{c}{Completeness (\%)} \\ \cline{3-11} 
                                &                                  & @1°               & @3°              & @5°              & 1cm       & 2cm       & 5cm       & 1cm        & 2cm         & 5cm        \\ \hline
\multirow{4}{*}{Detector-based} & SIFT~\cite{liu2010sift}+NN+PixSfM~\cite{lindenberger2021pixel}                   & 26.94             & 39.01            & 42.19            & 76.18     & 85.60     & 93.16     & 0.17       & 0.71        & 3.29       \\
                                & D2Net~\cite{dusmanu2019d2}+NN+PixSfM~\cite{lindenberger2021pixel}                  & 34.50             & 49.77            & 53.58            & 74.75     & 83.81     & 91.98     & \textbf{0.83}       & 2.69        & 8.95       \\
                                & R2D2~\cite{revaud2019r2d2}+NN+PixSfM~\cite{lindenberger2021pixel}                   & 43.58             & 62.09            & 66.89            & 74.12     & 84.49     & 91.98     & 0.43       & 1.58        & 6.71       \\
                                & SP+SG~\cite{sarlin2020superglue}+PixSfM~\cite{lindenberger2021pixel}                     & \textbf{50.82}             & \textbf{68.52}            & \textbf{72.86}            & \textbf{79.01}     & \textbf{87.04}     & \textbf{93.80}     & 0.75       & \textbf{2.77}        & \textbf{11.28}      \\ \hline
\multirow{3}{*}{Detector-free}  & LoFTR~\cite{sun2021loftr}+DetectorFreeSfM~\cite{he2023detector}            & \textbf{59.12}             & \textbf{75.59}            & \textbf{79.53}            & \textbf{80.38}     & \textbf{89.01}     & \textbf{95.83}     & 3.73       & 11.07       & 29.54      \\
                                & ASpanFormer~\cite{chen2022aspanformer}+DetectorFreeSfM~\cite{he2023detector} & 57.23             & 73.71            & 77.70            & 77.63     & 87.40     & 95.02     & \textbf{3.97}       & \textbf{12.18}       & \textbf{32.42}      \\
                                & MatchFormer~\cite{wang2022matchformer}+DetectorFreeSfM~\cite{he2023detector}      & 56.70             & 73.00            & 76.84            & 79.86     & 88.51     & 95.48     & 3.76       & 11.06       & 29.05     \\

\specialrule{1pt}{0pt}{0pt}
\end{tabular*}
\label{tab:SFM}
\end{table*}

\section{Challenges and Opportunities}

Deep learning has brought significant advantages to image-based local feature matching. However, there are still several challenges that remain to be addressed. In the following sections, we will explore potential research directions that we believe will provide valuable momentum for further advancements in image-based local feature matching algorithms.

\subsection{Efficient Attention and Transformer}

For local feature matching, integrating transformers into GNN models can be considered, framing the task as a graph matching problem involving two sets of features. To enhance the construction of better matchers, the self-attention and cross-attention layers within transformers are commonly employed to exchange global visual and geometric information between nodes. In addition to sparse descriptors generated by matching detectors, direct application of self-attention and cross-attention to feature maps extracted by CNNs, and generating matches in a coarse-to-fine manner, has also been explored~\cite{sarlin2020superglue, sun2021loftr}.
However, the computational cost of basic matrix multiplication in transformers remains high when dealing with a large number of keypoints. In recent years, efforts have been made to enhance the efficiency of transformers and attempts to compute the two types of attention in parallel have been made, continuously reducing complexity while maintaining performance~\cite{chen2021learning, shi2022clustergnn, deng2023resmatch, lu2023paraformer}. Future research will further optimize the structures of attention mechanisms and transformers, aiming to maintain high matching performance while reducing computational complexity. This will make local feature matching methods more efficient and applicable in real-time and resource-constrained environments.

\subsection{Adaptation strategy}

In recent years, significant progress has been made in the research on adaptability in local feature matching~\cite{lindenberger2023lightglue,yu2023adaptive,huang2023adaptive,ni2023pats,chen2022aspanformer}. For latency-sensitive applications, adaptive mechanisms can be incorporated into the matching process. This allows the modulation of network depth and width based on factors such as visual overlap and appearance variation, enabling fine-grained control over the difficulty of the matching task.
Furthermore, researchers have proposed various innovative approaches to address issues such as scale variations. One key challenge is how to adaptively adjust the size of the cropping grid based on image scale variations to avoid matching failures. Geometric inconsistencies in patch-level feature matching can also be alleviated through adaptive allocation strategies, combined with adaptive patch subdivision strategies that enhance correspondence quality progressively from coarse to fine during the matching process.
On the other hand, attention spans can be adaptively adjusted based on the difficulty of matching, achieving variable-sized adaptive attention regions at different positions. This allows the network to better adapt to features at different locations while capturing contextual information, thereby improving matching performance.

In summary, the research on adaptability in local feature matching offers vast prospects and opportunities for future developments, while enhancing efficiency in terms of memory and computation. With the emergence of more demands and challenges across various fields, it is anticipated that adaptive mechanisms will play an increasingly important role in local feature matching. Future research could further explore finer-grained adaptive strategies to achieve more efficient and accurate matching results.

\subsection{Weakly Supervised Learning}

The field of local feature matching has not only achieved significant progress under fully supervised settings but has also shown potential in the realm of weakly supervised learning. Traditional fully supervised methods rely on dense ground-truth correspondence labels. In recent years, researchers have turned their attention to self-supervised and weakly supervised learning to reduce the dependency on precise annotations. Self-supervised learning methods like SuperPoint~\cite{detone2018superpoint} train on pairs of images generated through virtual homography transformations, yielding promising results. However, these simple geometric transformations might not work effectively under complex scenarios.
Weakly supervised learning has become a research focus in the domain of local feature learning~\cite{wang2020learning,tyszkiewicz2020disk,zhou2021patch2pix,li2022decoupling,sun2022shared}. These methods often combine weakly supervised learning with the describe-detect pipeline, but direct use of weakly supervised loss leads to noticeable performance drops. Some approaches rely solely on solutions involving relative camera poses, learning descriptors via epipolar loss.
The limitations of weakly supervised methods lie in their difficulty to differentiate errors introduced by descriptors and keypoints, as well as accurately distinguishing different descriptors. To overcome these challenges, carefully designed decoupled training pipelines have emerged, where the description network and detection network are trained separately until high-quality descriptors are obtained. 
Chen et al.~\cite{chen2022deep} propose innovative methods using convolutional neural networks for feature shape estimation, orientation assignment, and descriptor learning. Their approach establishes a standard shape and orientation for each feature, enabling a transition from supervised to self-supervised learning by eliminating the need for known feature matching relationships. They also introduce a 'weak match finder' in descriptor learning, enhancing feature appearance variability and improving descriptor invariance. These advancements signify a significant progress in weakly supervised learning for feature matching, especially in cases involving substantial viewpoint and viewing direction changes.

These weakly supervised methods open up new prospects and opportunities for local feature learning, allowing models to be trained on larger and more diverse datasets, thereby obtaining more generalized descriptors. However, these methods still face challenges, such as effectively leveraging weak supervision signals and addressing the uncertainty of descriptors and keypoints.
In the future, developments in weakly supervised learning in the domain of local feature matching might focus on finer loss function designs, better utilization of weak supervision signals, and broader application domains. Exploring mechanisms for weakly supervised learning in conjunction with traditional fully supervised methods holds promise for enhancing the performance and generalization capabilities of local feature matching in complex scenes.

\subsection{Foundation Segmentation Models}

Typically, semantic segmentation models, trained on datasets such as Cityscapes~\cite{cordts2016cityscapes} and MIT ADE20k~\cite{zhou2017scene}, offer fundamental semantic information and play a crucial role in enhancing the detection and description processes of specific environments ~\cite{fan2022learning, xue2023sfd2}.

However, the advent of large foundation models such as SAM~\cite{kirillov2023segment}, DINO~\cite{caron2021emerging}, and DINOv2~\cite{oquab2023dinov2} marks a new era in artificial intelligence. While traditional segmentation models excel in their specific domains, these foundation models introduce a broader, more versatile approach. Their extensive pre-training on massive, diverse datasets equips them with a remarkable zero-shot generalization capability, enabling them to adapt to a wide range of scenarios.
For instance, SAMFeat~\cite{wu2023segment} demonstrates how SAM, a model adept at segmenting "anything" in "any scene," can guide local feature learning with its rich, category-agnostic semantic knowledge. By distilling fine-grained semantic relations and focusing on edge detection, SAMFeat showcases a significant enhancement in local feature description and accuracy. Similarly, SelaVPR~\cite{anonymous2023towards} demonstrates how to effectively adjust the DINOv2 model using lightweight adapters to address the challenges in Visual Place Recognition (VPR) by proficiently matching local features without the need for extensive spatial verification, thus streamlining the retrieval process.

Looking towards an open-world scenario, the versatility and robust generalization offered by these large foundation models present exciting prospects. Their ability to understand and interpret a vast array of scenes and objects far exceeds the scope of traditional segmentation networks, paving the way for advancements in feature matching across diverse and dynamic environments.
In summary, while the contributions of traditional semantic segmentation networks remain invaluable, the integration of large foundation models offers a complementary and expansive approach, essential for pushing the boundaries of what is achievable in feature matching, particularly in open-world applications.

\subsection{Mismatch Removal}

Image matching, involving the establishment of reliable connections between two images portraying a shared object or scene, poses intricate challenges due to the combinatorial nature of the process and the presence of outliers. Direct matching methodologies, such as point set registration and graph matching, often grapple with formidable computational demands and erratic performance. Consequently, a bifurcated approach, commencing with preliminary match construction utilizing feature descriptors like SIFT, ORB, and SURF~\cite{lowe2004distinctive,rublee2011orb,bay2006surf}, succeeded by the application of local and global geometrical constraints, has become a prevalent strategy. Nevertheless, these methodologies encounter constraints, notably when confronted with multi-modal images or substantial variations in viewpoint and lighting~\cite{jiang2022robust}.

The evolution of methodologies for outlier rejection has been crucial in overcoming the challenges of mismatch elimination, as highlighted by Ma et al.~\cite{ma2019locality}. Traditional methods, epitomized by RANSAC~\cite{fischler1981random} and its variants such as USAC~\cite{raguram2012usac} and MAGSAC++~\cite{barath2020magsac++}, have significantly improved efficiency and accuracy in outlier rejection. Nonetheless, these approaches are limited by computational time constraints and their suitability for non-rigid contexts. Techniques specific to non-rigid scenarios, like ICF~\cite{li2010rejecting}, have shown efficacy in addressing geometric distortions. The Advent of Learning-Driven Strategies in Mismatch Elimination The integration of deep learning into mismatch elimination has opened new pathways for enhancing feature matching. Yi et al.~\cite{yi2018learning} introduced Context Normalization (CNe), a groundbreaking concept that has transformed wide-baseline stereo correspondences by effectively distinguishing inliers from outliers. Expanding upon this, Sun et al.~\cite{sun2020acne} developed Attentive Context Networks (ACNe), which improved the management of permutation-equivariant data through Attentive Context Normalization, yielding significant advancements in camera pose estimation and point cloud classification. Zhang et al.~\cite{zhang2019learning} proposed the OANet, a novel methodology that precisely determines two-view correspondences and bolsters geometric estimations using a hierarchical clustering approach. Zhao et al.~\cite{zhao2019nm} introduced NM-Net, a layered network focusing on the selection of feature correspondences with compatibility-specific mining, demonstrating outstanding performance in various settings. 
Shape-Former~\cite{chen2023shape} innovatively addresses multimodal and multiview image matching challenges, focusing on robust mismatch removal through a hybrid neural network. Leveraging CNNs and Transformers, Shape-Former introduces ShapeConv for sparse matches learning, excelling in outlier estimation and consensus representation while showcasing superior performance.
Given that RANSAC is an integral part of the matching pipeline, recent innovations have significantly enhanced its integration with deep learning approaches for improved performance.
DSAC~\cite{brachmann2017dsac} introduces a paradigm shift by making RANSAC differentiable, employing a probabilistic selection mechanism that facilitates its integration into end-to-end trainable deep learning pipelines. This innovative approach not only maintains the robustness of traditional RANSAC but also leverages deep learning to directly minimize expected losses.
On the other hand, CA-RANSAC~\cite{cavalli2023consensus} evolves the RANSAC framework by incorporating an adaptive consensus mechanism through a novel attention layer. This mechanism dynamically refines per-point estimation states based on accumulated residuals, enhancing model refinement and sample selection.
Recent developments like LSVANet~\cite{chen2021lsv}, LGSC~\cite{jiang2022robust}, and HCA-Net~\cite{chen2022hierarchical}, have shown promise in more effectively discerning inliers from outliers. These approaches leverage deep learning modules for geometric estimation and feature correspondence categorization, signifying an advancement over conventional methods.

Primarily, the development of more generalized and robust learning-based methodologies adept at handling a diverse array of scenarios, encompassing non-rigid transformations and multi-modal images, is imperative. Secondly, there exists a necessity for methodologies that amalgamate the virtues of both traditional geometric approaches and contemporary learning-based techniques. Such hybrid approaches hold the potential to deliver superior performance by capitalizing on the strengths of both paradigms.   Lastly, the exploration of innovative learning architectures and loss functions tailored for mismatch elimination can unveil novel prospects in feature matching, elevating the overall resilience and precision of computer vision systems. In conclusion, the elimination of mismatches persists as a pivotal yet formidable facet of local feature matching. The ongoing evolution of both traditional and learning-based methodologies unfolds promising trajectories to address extant limitations and unlock newfound potentials in computer vision applications.

\subsection{Deep Learning and Handcrafted Analogies}

The field of image matching is witnessing a unique blend of deep learning and traditional handcrafted techniques. This convergence is evident in the adoption of foundational elements from classic methods, such as the "SIFT" pipeline, in recent semi-dense, detector-free approaches. Notable examples of this trend include the Hybrid Pipeline (HP) by Bellavia et al.\cite{bellavia2022image}, HarrisZ+\cite{bellavia2022harrisz+}, and Slime~\cite{bellavia2024image}, all demonstrating competitive capabilities alongside state-of-the-art deep methods.
The HP method integrates handcrafted and learning-based approaches, maintaining crucial rotational invariance for photogrammetric surveys. It features the novel Keypoint Filtering by Coverage (KFC) module, enhancing the accuracy of the overall pipeline. HarrisZ+ represents an evolution of the classic Harris corner detector, optimized to synergize with modern image matching components. It yields more discriminative and accurately placed keypoints, aligning closely with results from contemporary deep learning models. Slime, employs a novel strategy of modeling scenes with local overlapping planes, merging local affine approximation principles with global matching constraints. This hybrid approach echoes traditional image matching processes, challenging the performance of deep learning methods.
Meanwhile, it is important to highlight a key distinction between deep learning methods and traditional handcrafted detectors in recent advancements in image matching: rotation equivariance. Despite the remarkable matching performances of modern methods, they often fall short in handling in-plane rotations—a fundamental feature inherently integrated into handcrafted detectors. This oversight reveals a performance gap under rotational transformations, underscoring the importance of designing or training deep learning models to explicitly address this challenge. By focusing on the development of rotation-equivariant approaches, as exemplified by SE2-LoFTR~\cite{bokman2022case} and S-TREK~\cite{santellani2023s}, the field moves closer to bridging this gap, combining the precision of deep learning with the robustness of handcrafted detectors against orientation changes.

These advancements signify that despite the significant strides made by deep learning methods like LoFTR and SuperGlue, the foundational principles of handcrafted techniques remain vital. The integration of classical concepts with modern computational power, as seen in HP, HarrisZ+, Slime, SE2-LoFTR, and S-TREK, leads to robust image matching solutions. These methods offer potential avenues for future research that blends diverse methodologies, bridging the gap between traditional and modern approaches in image matching.

\subsection{Utilizing geometric information}

When facing challenges such as texturelessness, occlusion, and repetitive patterns, traditional local feature matching methods may perform poorly. In recent years, researchers have started to focus on better utilizing geometric information to enhance the effectiveness of local feature matching in the presence of these challenges.
Several studies~\cite{chang2023structured,edstedt2023dkm,truong2023pdc,zhu2023pmatch} have indicated that leveraging geometric information holds significant potential for local feature matching. By capturing geometric relationships between pixels more accurately and combining geometric priors with image appearance information, these methods can enhance the robustness and accuracy of matching in complex scenes. However, this direction presents numerous opportunities and challenges for future development. Firstly, how to model geometric information more profoundly to better address scenarios involving large displacements, occlusions, and textureless regions remains a critical question. Secondly, improving the performance of confidence estimation to yield more reliable matching results is also a direction worthy of investigation.

The introduction of geometric priors expands feature matching beyond mere appearance similarity to consider an object's behavior from different viewpoints. This trend suggests that dense matching methods hold promise for tackling challenges posed by large displacements and appearance variations. It also implies that the future development in the field of geometric matching may increasingly focus on dense feature matching, leveraging geometric information and prior knowledge to enhance matching performance.

\subsection{Advancing Cultural Heritage Preservation}

The integration of deep learning into the realm of historical image matching heralds a new era for cultural heritage preservation, offering unparalleled opportunities while presenting unique challenges.  Insights from recent research underscore the potential of advanced deep learning methods to overcome the limitations of traditional techniques~\cite{maiwald2021fully}.  These approaches have shown remarkable resilience to the inherent radiometric and geometric discrepancies that plague the matching of historical and contemporary imagery, thereby facilitating the accurate co-registration of multi-temporal scenes.  Techniques such as SuperGlue~\cite{sarlin2020superglue}, LoFTR~\cite{sun2021loftr}, and DISK~\cite{tyszkiewicz2020disk} have been identified as particularly effective, surpassing classical methods by achieving higher robustness against severe illumination and viewpoint shifts.  This advancement enables more accurate reconstructions of historical sites and artifacts~\cite{morelli2022photogrammetry}, thus enhancing knowledge transfer and cultural heritage promotion through immersive technologies like virtual and augmented reality.

However, the journey towards fully harnessing the capabilities of deep learning in this domain is not without its obstacles.  Challenges persist in managing extensive image rotations and variations in scale, which are common in historical datasets~\cite{bellavia2022challenges}.  Additionally, the computational intensity required to process high-resolution images presents a significant hurdle, particularly when aiming for real-time application in web and VR/AR platforms~\cite{maiwald2021automatic}.  These technical challenges underscore the necessity for ongoing research and development to refine deep learning algorithms and optimize them for the specific demands of cultural heritage applications.

Moreover, the dynamic nature of deep learning presents an opportunity to continually improve the accuracy and efficiency of historical image matching processes.  As algorithms evolve and new approaches emerge, there is the potential for even more sophisticated methods of feature detection, extraction, and matching that could further revolutionize the field.  The exploration of these novel strategies, alongside the adaptation of current methodologies to the unique characteristics of historical imagery, is essential for advancing our ability to digitally preserve and explore our cultural heritage~\cite{bellavia2022challenges}.  In summary, the confluence of deep learning with historical image matching offers a promising pathway towards bridging the gap between the past and present.  Future research can unlock new possibilities for the preservation, understanding, and dissemination of cultural heritage, making it more accessible and engaging for generations to come.

\section{Conclusions}

We have investigated various algorithms related to local feature matching based on deep learning models over the past five years. These algorithms have demonstrated impressive performance in various local feature matching tasks and benchmark tests. They can be broadly categorized into Detector-based Models and Detector-free Models. The application of feature detectors reduces the scope of matching and relies on the processes of keypoint detection and feature description. Detector-free methods, on the other hand, directly capture richer context from the raw images to generate dense matches. Subsequently, we discuss the strengths and weaknesses of existing local feature matching algorithms, introduce popular datasets and evaluation standards, and summarize the quantitative performance analysis of these models on some common benchmarks such as HPatches, ScanNet, YFCC100M, MegaDepth, and Aachen Day-Night datasets. Lastly, we explore the open challenges and potential research avenues that the field of local feature matching may encounter in the forthcoming years. Our aim is not only to enhance researchers' understanding of local feature matching but also to inspire and guide future research endeavors in this domain.


\bibliography{IF}

\begin{thebibliography}{100}
\expandafter\ifx\csname url\endcsname\relax
  \def\url#1{\texttt{#1}}\fi
\expandafter\ifx\csname urlprefix\endcsname\relax\def\urlprefix{URL }\fi
\expandafter\ifx\csname href\endcsname\relax
  \def\href#1#2{#2} \def\path#1{#1}\fi

\bibitem{tang2022image}
L.~Tang, J.~Yuan, J.~Ma, Image fusion in the loop of high-level vision tasks: A semantic-aware real-time infrared and visible image fusion network, Information Fusion 82 (2022) 28--42.

\bibitem{cao2023pcnet}
S.-Y. Cao, B.~Yu, L.~Luo, R.~Zhang, S.-J. Chen, C.~Li, H.-L. Shen, Pcnet: A structure similarity enhancement method for multispectral and multimodal image registration, Information Fusion 94 (2023) 200--214.

\bibitem{hu2023multiscale}
M.~Hu, B.~Sun, X.~Kang, S.~Li, Multiscale structural feature transform for multi-modal image matching, Information Fusion 95 (2023) 341--354.

\bibitem{sun2023unified}
K.~Sun, J.~Yu, W.~Tao, X.~Li, C.~Tang, Y.~Qian, A unified feature-spatial cycle consistency fusion framework for robust image matching, Information Fusion 97 (2023) 101810.

\bibitem{hou2024pos}
Z.~Hou, Y.~Liu, L.~Zhang, Pos-gift: A geometric and intensity-invariant feature transformation for multimodal images, Information Fusion 102 (2024) 102027.

\bibitem{sattler2012improving}
T.~Sattler, B.~Leibe, L.~Kobbelt, Improving image-based localization by active correspondence search, in: Computer Vision--ECCV 2012: 12th European Conference on Computer Vision, Florence, Italy, October 7-13, 2012, Proceedings, Part I 12, Springer, 2012, pp. 752--765.

\bibitem{sattler2017large}
T.~Sattler, A.~Torii, J.~Sivic, M.~Pollefeys, H.~Taira, M.~Okutomi, T.~Pajdla, Are large-scale 3d models really necessary for accurate visual localization?, in: Proceedings of the IEEE Conference on Computer Vision and Pattern Recognition, 2017, pp. 1637--1646.

\bibitem{cai2019ground}
S.~Cai, Y.~Guo, S.~Khan, J.~Hu, G.~Wen, Ground-to-aerial image geo-localization with a hard exemplar reweighting triplet loss, in: Proceedings of the IEEE/CVF International Conference on Computer Vision, 2019, pp. 8391--8400.

\bibitem{zhang2021reference}
Z.~Zhang, T.~Sattler, D.~Scaramuzza, Reference pose generation for long-term visual localization via learned features and view synthesis, International Journal of Computer Vision 129 (2021) 821--844.

\bibitem{agarwal2011building}
S.~Agarwal, Y.~Furukawa, N.~Snavely, I.~Simon, B.~Curless, S.~M. Seitz, R.~Szeliski, Building rome in a day, Communications of the ACM 54~(10) (2011) 105--112.

\bibitem{heinly2015reconstructing}
J.~Heinly, J.~L. Schonberger, E.~Dunn, J.-M. Frahm, Reconstructing the world* in six days*(as captured by the yahoo 100 million image dataset), in: Proceedings of the IEEE conference on computer vision and pattern recognition, 2015, pp. 3287--3295.

\bibitem{schonberger2016structure}
J.~L. Schonberger, J.-M. Frahm, Structure-from-motion revisited, in: Proceedings of the IEEE conference on computer vision and pattern recognition, 2016, pp. 4104--4113.

\bibitem{wang2021deep}
J.~Wang, Y.~Zhong, Y.~Dai, S.~Birchfield, K.~Zhang, N.~Smolyanskiy, H.~Li, Deep two-view structure-from-motion revisited, in: Proceedings of the IEEE/CVF conference on Computer Vision and Pattern Recognition, 2021, pp. 8953--8962.

\bibitem{cadena2016past}
C.~Cadena, L.~Carlone, H.~Carrillo, Y.~Latif, D.~Scaramuzza, J.~Neira, I.~Reid, J.~J. Leonard, Past, present, and future of simultaneous localization and mapping: Toward the robust-perception age, IEEE Transactions on robotics 32~(6) (2016) 1309--1332.

\bibitem{mur2017orb}
R.~Mur-Artal, J.~D. Tard{\'o}s, Orb-slam2: An open-source slam system for monocular, stereo, and rgb-d cameras, IEEE transactions on robotics 33~(5) (2017) 1255--1262.

\bibitem{zhao2019gslam}
Y.~Zhao, S.~Xu, S.~Bu, H.~Jiang, P.~Han, Gslam: A general slam framework and benchmark, in: Proceedings of the IEEE/CVF International Conference on Computer Vision, 2019, pp. 1110--1120.

\bibitem{liu2010sift}
C.~Liu, J.~Yuen, A.~Torralba, Sift flow: Dense correspondence across scenes and its applications, IEEE transactions on pattern analysis and machine intelligence 33~(5) (2010) 978--994.

\bibitem{weinzaepfel2013deepflow}
P.~Weinzaepfel, J.~Revaud, Z.~Harchaoui, C.~Schmid, Deepflow: Large displacement optical flow with deep matching, in: Proceedings of the IEEE international conference on computer vision, 2013, pp. 1385--1392.

\bibitem{dosovitskiy2015flownet}
A.~Dosovitskiy, P.~Fischer, E.~Ilg, P.~Hausser, C.~Hazirbas, V.~Golkov, P.~Van Der~Smagt, D.~Cremers, T.~Brox, Flownet: Learning optical flow with convolutional networks, in: Proceedings of the IEEE international conference on computer vision, 2015, pp. 2758--2766.

\bibitem{radenovic2018fine}
F.~Radenovi{\'c}, G.~Tolias, O.~Chum, Fine-tuning cnn image retrieval with no human annotation, IEEE transactions on pattern analysis and machine intelligence 41~(7) (2018) 1655--1668.

\bibitem{cao2020unifying}
B.~Cao, A.~Araujo, J.~Sim, Unifying deep local and global features for image search, in: Computer Vision--ECCV 2020: 16th European Conference, Glasgow, UK, August 23--28, 2020, Proceedings, Part XX 16, Springer, 2020, pp. 726--743.

\bibitem{chhabra2020content}
P.~Chhabra, N.~K. Garg, M.~Kumar, Content-based image retrieval system using orb and sift features, Neural Computing and Applications 32 (2020) 2725--2733.

\bibitem{zhang2023task}
D.~Zhang, H.~Li, W.~Cong, R.~Xu, J.~Dong, X.~Chen, Task relation distillation and prototypical pseudo label for incremental named entity recognition, in: Proceedings of the 32nd ACM International Conference on Information and Knowledge Management, 2023, pp. 3319--3329.

\bibitem{wang2023generalized}
K.~Wang, X.~Fu, Y.~Huang, C.~Cao, G.~Shi, Z.-J. Zha, Generalized uav object detection via frequency domain disentanglement, in: Proceedings of the IEEE/CVF Conference on Computer Vision and Pattern Recognition, 2023, pp. 1064--1073.

\bibitem{cao2023event}
C.~Cao, X.~Fu, H.~Liu, Y.~Huang, K.~Wang, J.~Luo, Z.-J. Zha, Event-guided person re-identification via sparse-dense complementary learning, in: Proceedings of the IEEE/CVF Conference on Computer Vision and Pattern Recognition, 2023, pp. 17990--17999.

\bibitem{harris1988combined}
C.~Harris, M.~Stephens, et~al., A combined corner and edge detector, in: Alvey vision conference, Vol.~15, Citeseer, 1988, pp. 10--5244.

\bibitem{smith1997susan}
S.~M. Smith, J.~M. Brady, Susan—a new approach to low level image processing, International journal of computer vision 23~(1) (1997) 45--78.

\bibitem{rosten2005fusing}
E.~Rosten, T.~Drummond, Fusing points and lines for high performance tracking, in: Tenth IEEE International Conference on Computer Vision (ICCV'05) Volume 1, Vol.~2, Ieee, 2005, pp. 1508--1515.

\bibitem{matas2004robust}
J.~Matas, O.~Chum, M.~Urban, T.~Pajdla, Robust wide-baseline stereo from maximally stable extremal regions, Image and vision computing 22~(10) (2004) 761--767.

\bibitem{dalal2005histograms}
N.~Dalal, B.~Triggs, Histograms of oriented gradients for human detection, in: 2005 IEEE computer society conference on computer vision and pattern recognition (CVPR'05), Vol.~1, Ieee, 2005, pp. 886--893.

\bibitem{mikolajczyk2005performance}
K.~Mikolajczyk, C.~Schmid, A performance evaluation of local descriptors, IEEE transactions on pattern analysis and machine intelligence 27~(10) (2005) 1615--1630.

\bibitem{calonder2010brief}
M.~Calonder, V.~Lepetit, C.~Strecha, P.~Fua, Brief: Binary robust independent elementary features, in: Computer Vision--ECCV 2010: 11th European Conference on Computer Vision, Heraklion, Crete, Greece, September 5-11, 2010, Proceedings, Part IV 11, Springer, 2010, pp. 778--792.

\bibitem{lowe2004distinctive}
D.~G. Lowe, Distinctive image features from scale-invariant keypoints, International journal of computer vision 60 (2004) 91--110.

\bibitem{bay2006surf}
H.~Bay, T.~Tuytelaars, L.~Van~Gool, Surf: Speeded up robust features, Lecture notes in computer science 3951 (2006) 404--417.

\bibitem{rublee2011orb}
E.~Rublee, V.~Rabaud, K.~Konolige, G.~Bradski, Orb: An efficient alternative to sift or surf, in: 2011 International conference on computer vision, Ieee, 2011, pp. 2564--2571.

\bibitem{leutenegger2011brisk}
S.~Leutenegger, M.~Chli, R.~Y. Siegwart, Brisk: Binary robust invariant scalable keypoints, in: 2011 International conference on computer vision, Ieee, 2011, pp. 2548--2555.

\bibitem{alcantarilla2012kaze}
P.~F. Alcantarilla, A.~Bartoli, A.~J. Davison, Kaze features, in: Computer Vision--ECCV 2012: 12th European Conference on Computer Vision, Florence, Italy, October 7-13, 2012, Proceedings, Part VI 12, Springer, 2012, pp. 214--227.

\bibitem{alcantarilla2011fast}
P.~F. Alcantarilla, T.~Solutions, Fast explicit diffusion for accelerated features in nonlinear scale spaces, IEEE Trans. Patt. Anal. Mach. Intell 34~(7) (2011) 1281--1298.

\bibitem{fischler1981random}
M.~A. Fischler, R.~C. Bolles, Random sample consensus: a paradigm for model fitting with applications to image analysis and automated cartography, Communications of the ACM 24~(6) (1981) 381--395.

\bibitem{xu2022domaindesc}
R.~Xu, C.~Wang, B.~Fan, Y.~Zhang, S.~Xu, W.~Meng, X.~Zhang, Domaindesc: Learning local descriptors with domain adaptation, in: ICASSP 2022-2022 IEEE International Conference on Acoustics, Speech and Signal Processing (ICASSP), IEEE, 2022, pp. 2505--2509.

\bibitem{xu2023domainfeat}
R.~Xu, C.~Wang, S.~Xu, W.~Meng, Y.~Zhang, B.~Fan, X.~Zhang, Domainfeat: Learning local features with domain adaptation, IEEE Transactions on Circuits and Systems for Video Technology (2023).

\bibitem{wu2023segment}
J.~Wu, R.~Xu, Z.~Wood-Doughty, C.~Wang, Segment anything model is a good teacher for local feature learning, arXiv preprint arXiv:2309.16992 (2023).

\bibitem{jiang2021review}
X.~Jiang, J.~Ma, G.~Xiao, Z.~Shao, X.~Guo, A review of multimodal image matching: Methods and applications, Information Fusion 73 (2021) 22--71.

\bibitem{xu2022instance}
R.~Xu, Y.~Li, C.~Wang, S.~Xu, W.~Meng, X.~Zhang, Instance segmentation of biological images using graph convolutional network, Engineering Applications of Artificial Intelligence 110 (2022) 104739.

\bibitem{xu2021dc}
R.~Xu, C.~Wang, S.~Xu, W.~Meng, X.~Zhang, Dc-net: Dual context network for 2d medical image segmentation, in: Medical Image Computing and Computer Assisted Intervention--MICCAI 2021: 24th International Conference, Strasbourg, France, September 27--October 1, 2021, Proceedings, Part I 24, Springer, 2021, pp. 503--513.

\bibitem{xu2023wave}
R.~Xu, C.~Wang, S.~Xu, W.~Meng, X.~Zhang, Wave-like class activation map with representation fusion for weakly-supervised semantic segmentation, IEEE Transactions on Multimedia (2023).

\bibitem{xu2023spectral}
W.~Xu, R.~Xu, C.~Wang, S.~Xu, L.~Guo, M.~Zhang, X.~Zhang, Spectral prompt tuning: Unveiling unseen classes for zero-shot semantic segmentation, arXiv preprint arXiv:2312.12754 (2023).

\bibitem{wang2023treating}
C.~Wang, R.~Xu, S.~Xu, W.~Meng, X.~Zhang, Treating pseudo-labels generation as image matting for weakly supervised semantic segmentation, in: Proceedings of the IEEE/CVF International Conference on Computer Vision, 2023, pp. 755--765.

\bibitem{awrangjeb2012performance}
M.~Awrangjeb, G.~Lu, C.~S. Fraser, Performance comparisons of contour-based corner detectors, IEEE Transactions on Image Processing 21~(9) (2012) 4167--4179.

\bibitem{li2015survey}
Y.~Li, S.~Wang, Q.~Tian, X.~Ding, A survey of recent advances in visual feature detection, Neurocomputing 149 (2015) 736--751.

\bibitem{krig2016interest}
S.~Krig, S.~Krig, Interest point detector and feature descriptor survey, Computer Vision Metrics: Textbook Edition (2016) 187--246.

\bibitem{joshi2020recent}
K.~Joshi, M.~I. Patel, Recent advances in local feature detector and descriptor: a literature survey, International Journal of Multimedia Information Retrieval 9~(4) (2020) 231--247.

\bibitem{ma2021image}
J.~Ma, X.~Jiang, A.~Fan, J.~Jiang, J.~Yan, Image matching from handcrafted to deep features: A survey, International Journal of Computer Vision 129 (2021) 23--79.

\bibitem{jing2022image}
J.~Jing, T.~Gao, W.~Zhang, Y.~Gao, C.~Sun, Image feature information extraction for interest point detection: A comprehensive review, IEEE Transactions on Pattern Analysis and Machine Intelligence (2022).

\bibitem{bellavia2022challenges}
F.~Bellavia, C.~Colombo, L.~Morelli, F.~Remondino, Challenges in image matching for cultural heritage: an overview and perspective, in: International Conference on Image Analysis and Processing, Springer, 2022, pp. 210--222.

\bibitem{haskins2020deep}
G.~Haskins, U.~Kruger, P.~Yan, Deep learning in medical image registration: a survey, Machine Vision and Applications 31 (2020) 1--18.

\bibitem{bharati2022deep}
S.~Bharati, M.~Mondal, P.~Podder, V.~Prasath, Deep learning for medical image registration: A comprehensive review, arXiv preprint arXiv:2204.11341 (2022).

\bibitem{chen2023survey}
J.~Chen, Y.~Liu, S.~Wei, Z.~Bian, S.~Subramanian, A.~Carass, J.~L. Prince, Y.~Du, A survey on deep learning in medical image registration: New technologies, uncertainty, evaluation metrics, and beyond, arXiv preprint arXiv:2307.15615 (2023).

\bibitem{paul2021comprehensive}
S.~Paul, U.~C. Pati, A comprehensive review on remote sensing image registration, International Journal of Remote Sensing 42~(14) (2021) 5396--5432.

\bibitem{zhu2023advances}
B.~Zhu, L.~Zhou, S.~Pu, J.~Fan, Y.~Ye, Advances and challenges in multimodal remote sensing image registration, IEEE Journal on Miniaturization for Air and Space Systems (2023).

\bibitem{detone2018superpoint}
D.~DeTone, T.~Malisiewicz, A.~Rabinovich, Superpoint: Self-supervised interest point detection and description, in: Proceedings of the IEEE conference on computer vision and pattern recognition workshops, 2018, pp. 224--236.

\bibitem{dusmanu2019d2}
M.~Dusmanu, I.~Rocco, T.~Pajdla, M.~Pollefeys, J.~Sivic, A.~Torii, T.~Sattler, D2-net: A trainable cnn for joint description and detection of local features, in: Proceedings of the ieee/cvf conference on computer vision and pattern recognition, 2019, pp. 8092--8101.

\bibitem{revaud2019r2d2}
J.~Revaud, P.~Weinzaepfel, C.~De~Souza, N.~Pion, G.~Csurka, Y.~Cabon, M.~Humenberger, R2d2: repeatable and reliable detector and descriptor, arXiv preprint arXiv:1906.06195 (2019).

\bibitem{rocco2018neighbourhood}
I.~Rocco, M.~Cimpoi, R.~Arandjelovi{\'c}, A.~Torii, T.~Pajdla, J.~Sivic, Neighbourhood consensus networks, Advances in neural information processing systems 31 (2018).

\bibitem{rocco2020efficient}
I.~Rocco, R.~Arandjelovi{\'c}, J.~Sivic, Efficient neighbourhood consensus networks via submanifold sparse convolutions, in: Computer Vision--ECCV 2020: 16th European Conference, Glasgow, UK, August 23--28, 2020, Proceedings, Part IX 16, Springer, 2020, pp. 605--621.

\bibitem{li2020dual}
X.~Li, K.~Han, S.~Li, V.~Prisacariu, Dual-resolution correspondence networks, Advances in Neural Information Processing Systems 33 (2020) 17346--17357.

\bibitem{truong2020glu}
P.~Truong, M.~Danelljan, R.~Timofte, Glu-net: Global-local universal network for dense flow and correspondences, in: Proceedings of the IEEE/CVF conference on computer vision and pattern recognition, 2020, pp. 6258--6268.

\bibitem{truong2021learning}
P.~Truong, M.~Danelljan, L.~Van~Gool, R.~Timofte, Learning accurate dense correspondences and when to trust them, in: Proceedings of the IEEE/CVF Conference on Computer Vision and Pattern Recognition, 2021, pp. 5714--5724.

\bibitem{sarlin2020superglue}
P.-E. Sarlin, D.~DeTone, T.~Malisiewicz, A.~Rabinovich, Superglue: Learning feature matching with graph neural networks, in: Proceedings of the IEEE/CVF conference on computer vision and pattern recognition, 2020, pp. 4938--4947.

\bibitem{chen2021learning}
H.~Chen, Z.~Luo, J.~Zhang, L.~Zhou, X.~Bai, Z.~Hu, C.-L. Tai, L.~Quan, Learning to match features with seeded graph matching network, in: Proceedings of the IEEE/CVF International Conference on Computer Vision, 2021, pp. 6301--6310.

\bibitem{shi2022clustergnn}
Y.~Shi, J.-X. Cai, Y.~Shavit, T.-J. Mu, W.~Feng, K.~Zhang, Clustergnn: Cluster-based coarse-to-fine graph neural network for efficient feature matching, in: Proceedings of the IEEE/CVF Conference on Computer Vision and Pattern Recognition, 2022, pp. 12517--12526.

\bibitem{sun2021loftr}
J.~Sun, Z.~Shen, Y.~Wang, H.~Bao, X.~Zhou, Loftr: Detector-free local feature matching with transformers, in: Proceedings of the IEEE/CVF conference on computer vision and pattern recognition, 2021, pp. 8922--8931.

\bibitem{chen2022aspanformer}
H.~Chen, Z.~Luo, L.~Zhou, Y.~Tian, M.~Zhen, T.~Fang, D.~McKinnon, Y.~Tsin, L.~Quan, Aspanformer: Detector-free image matching with adaptive span transformer, in: Computer Vision--ECCV 2022: 17th European Conference, Tel Aviv, Israel, October 23--27, 2022, Proceedings, Part XXXII, Springer, 2022, pp. 20--36.

\bibitem{zhang20233d}
J.~Zhang, L.~Dai, F.~Meng, Q.~Fan, X.~Chen, K.~Xu, H.~Wang, 3d-aware object goal navigation via simultaneous exploration and identification, in: Proceedings of the IEEE/CVF Conference on Computer Vision and Pattern Recognition, 2023, pp. 6672--6682.

\bibitem{zhang2022asro}
J.~Zhang, Y.~Tang, H.~Wang, K.~Xu, Asro-dio: Active subspace random optimization based depth inertial odometry, IEEE Transactions on Robotics 39~(2) (2022) 1496--1508.

\bibitem{bian2017gms}
J.~Bian, W.-Y. Lin, Y.~Matsushita, S.-K. Yeung, T.-D. Nguyen, M.-M. Cheng, Gms: Grid-based motion statistics for fast, ultra-robust feature correspondence, in: Proceedings of the IEEE conference on computer vision and pattern recognition, 2017, pp. 4181--4190.

\bibitem{zhang2019learning}
J.~Zhang, D.~Sun, Z.~Luo, A.~Yao, L.~Zhou, T.~Shen, Y.~Chen, L.~Quan, H.~Liao, Learning two-view correspondences and geometry using order-aware network, in: Proceedings of the IEEE/CVF international conference on computer vision, 2019, pp. 5845--5854.

\bibitem{yi2016lift}
K.~M. Yi, E.~Trulls, V.~Lepetit, P.~Fua, Lift: Learned invariant feature transform, in: Computer Vision--ECCV 2016: 14th European Conference, Amsterdam, The Netherlands, October 11-14, 2016, Proceedings, Part VI 14, Springer, 2016, pp. 467--483.

\bibitem{brown2010discriminative}
M.~Brown, G.~Hua, S.~Winder, Discriminative learning of local image descriptors, IEEE transactions on pattern analysis and machine intelligence 33~(1) (2010) 43--57.

\bibitem{tian2017l2}
Y.~Tian, B.~Fan, F.~Wu, L2-net: Deep learning of discriminative patch descriptor in euclidean space, in: Proceedings of the IEEE conference on computer vision and pattern recognition, 2017, pp. 661--669.

\bibitem{yi2016learning}
K.~M. Yi, Y.~Verdie, P.~Fua, V.~Lepetit, Learning to assign orientations to feature points, in: Proceedings of the IEEE conference on computer vision and pattern recognition, 2016, pp. 107--116.

\bibitem{bromley1993signature}
J.~Bromley, I.~Guyon, Y.~LeCun, E.~S{\"a}ckinger, R.~Shah, Signature verification using a" siamese" time delay neural network, Advances in neural information processing systems 6 (1993).

\bibitem{mishchuk2017working}
A.~Mishchuk, D.~Mishkin, F.~Radenovic, J.~Matas, Working hard to know your neighbor's margins: Local descriptor learning loss, Advances in neural information processing systems 30 (2017).

\bibitem{he2018local}
K.~He, Y.~Lu, S.~Sclaroff, Local descriptors optimized for average precision, in: Proceedings of the IEEE conference on computer vision and pattern recognition, 2018, pp. 596--605.

\bibitem{wei2018kernelized}
X.~Wei, Y.~Zhang, Y.~Gong, N.~Zheng, Kernelized subspace pooling for deep local descriptors, in: Proceedings of the IEEE conference on computer vision and pattern recognition, 2018, pp. 1867--1875.

\bibitem{lin2018unsupervised}
K.~Lin, J.~Lu, C.-S. Chen, J.~Zhou, M.-T. Sun, Unsupervised deep learning of compact binary descriptors, IEEE transactions on pattern analysis and machine intelligence 41~(6) (2018) 1501--1514.

\bibitem{zieba2018bingan}
M.~Zieba, P.~Semberecki, T.~El-Gaaly, T.~Trzcinski, Bingan: Learning compact binary descriptors with a regularized gan, Advances in neural information processing systems 31 (2018).

\bibitem{wei2018glad}
L.~Wei, S.~Zhang, H.~Yao, W.~Gao, Q.~Tian, Glad: Global--local-alignment descriptor for scalable person re-identification, IEEE Transactions on Multimedia 21~(4) (2018) 986--999.

\bibitem{luo2018geodesc}
Z.~Luo, T.~Shen, L.~Zhou, S.~Zhu, R.~Zhang, Y.~Yao, T.~Fang, L.~Quan, Geodesc: Learning local descriptors by integrating geometry constraints, in: Proceedings of the European conference on computer vision (ECCV), 2018, pp. 168--183.

\bibitem{liu2019gift}
Y.~Liu, Z.~Shen, Z.~Lin, S.~Peng, H.~Bao, X.~Zhou, Gift: Learning transformation-invariant dense visual descriptors via group cnns, Advances in Neural Information Processing Systems 32 (2019).

\bibitem{lee2021learning}
J.~Lee, Y.~Jeong, S.~Kim, J.~Min, M.~Cho, Learning to distill convolutional features into compact local descriptors, in: Proceedings of the IEEE/CVF Winter Conference on Applications of Computer Vision, 2021, pp. 898--908.

\bibitem{tian2019sosnet}
Y.~Tian, X.~Yu, B.~Fan, F.~Wu, H.~Heijnen, V.~Balntas, Sosnet: Second order similarity regularization for local descriptor learning, in: Proceedings of the IEEE/CVF Conference on Computer Vision and Pattern Recognition, 2019, pp. 11016--11025.

\bibitem{ebel2019beyond}
P.~Ebel, A.~Mishchuk, K.~M. Yi, P.~Fua, E.~Trulls, Beyond cartesian representations for local descriptors, in: Proceedings of the IEEE/CVF international conference on computer vision, 2019, pp. 253--262.

\bibitem{tian2020hynet}
Y.~Tian, A.~Barroso~Laguna, T.~Ng, V.~Balntas, K.~Mikolajczyk, Hynet: Learning local descriptor with hybrid similarity measure and triplet loss, Advances in neural information processing systems 33 (2020) 7401--7412.

\bibitem{wang2022cndesc}
C.~Wang, R.~Xu, S.~Xu, W.~Meng, X.~Zhang, Cndesc: Cross normalization for local descriptors learning, IEEE Transactions on Multimedia (2022).

\bibitem{barroso2019key}
A.~Barroso-Laguna, E.~Riba, D.~Ponsa, K.~Mikolajczyk, Key. net: Keypoint detection by handcrafted and learned cnn filters, in: Proceedings of the IEEE/CVF international conference on computer vision, 2019, pp. 5836--5844.

\bibitem{zhao2022alike}
X.~Zhao, X.~Wu, J.~Miao, W.~Chen, P.~C. Chen, Z.~Li, Alike: Accurate and lightweight keypoint detection and descriptor extraction, IEEE Transactions on Multimedia (2022).

\bibitem{santellani2023s}
E.~Santellani, C.~Sormann, M.~Rossi, A.~Kuhn, F.~Fraundorfer, S-trek: Sequential translation and rotation equivariant keypoints for local feature extraction, in: Proceedings of the IEEE/CVF International Conference on Computer Vision, 2023, pp. 9728--9737.

\bibitem{kanakis2023zippypoint}
M.~Kanakis, S.~Maurer, M.~Spallanzani, A.~Chhatkuli, L.~Van~Gool, Zippypoint: Fast interest point detection, description, and matching through mixed precision discretization, in: Proceedings of the IEEE/CVF Conference on Computer Vision and Pattern Recognition, 2023, pp. 6113--6122.

\bibitem{tang2020neural}
J.~Tang, H.~Kim, V.~Guizilini, S.~Pillai, A.~Rares, Neural outlier rejection for self-supervised keypoint learning, in: 8th International Conference on Learning Representations, ICLR 2020, International Conference on Learning Representations, ICLR, 2020.

\bibitem{luo2019contextdesc}
Z.~Luo, T.~Shen, L.~Zhou, J.~Zhang, Y.~Yao, S.~Li, T.~Fang, L.~Quan, Contextdesc: Local descriptor augmentation with cross-modality context, in: Proceedings of the IEEE/CVF conference on computer vision and pattern recognition, 2019, pp. 2527--2536.

\bibitem{wang2022mtldesc}
C.~Wang, R.~Xu, Y.~Zhang, S.~Xu, W.~Meng, B.~Fan, X.~Zhang, Mtldesc: Looking wider to describe better, in: Proceedings of the AAAI Conference on Artificial Intelligence, Vol.~36, 2022, pp. 2388--2396.

\bibitem{wang2023attention}
C.~Wang, R.~Xu, K.~Lv, S.~Xu, W.~Meng, Y.~Zhang, B.~Fan, X.~Zhang, Attention weighted local descriptors, IEEE Transactions on Pattern Analysis and Machine Intelligence (2023).

\bibitem{chen2021igs}
J.~Chen, S.~Chen, Y.~Liu, X.~Chen, X.~Fan, Y.~Rao, C.~Zhou, Y.~Yang, Igs-net: Seeking good correspondences via interactive generative structure learning, IEEE Transactions on Geoscience and Remote Sensing 60 (2021) 1--13.

\bibitem{li2019rift}
J.~Li, Q.~Hu, M.~Ai, Rift: Multi-modal image matching based on radiation-variation insensitive feature transform, IEEE Transactions on Image Processing 29 (2019) 3296--3310.

\bibitem{rosten2006machine}
E.~Rosten, T.~Drummond, Machine learning for high-speed corner detection, in: Computer Vision--ECCV 2006: 9th European Conference on Computer Vision, Graz, Austria, May 7-13, 2006. Proceedings, Part I 9, Springer, 2006, pp. 430--443.

\bibitem{cui2020modality}
S.~Cui, M.~Xu, A.~Ma, Y.~Zhong, Modality-free feature detector and descriptor for multimodal remote sensing image registration, Remote Sensing 12~(18) (2020) 2937.

\bibitem{xie2023semantics}
H.~Xie, Y.~Zhang, J.~Qiu, X.~Zhai, X.~Liu, Y.~Yang, S.~Zhao, Y.~Luo, J.~Zhong, Semantics lead all: Towards unified image registration and fusion from a semantic perspective, Information Fusion 98 (2023) 101835.

\bibitem{mishkin2018repeatability}
D.~Mishkin, F.~Radenovic, J.~Matas, Repeatability is not enough: Learning affine regions via discriminability, in: Proceedings of the European conference on computer vision (ECCV), 2018, pp. 284--300.

\bibitem{truong2019glampoints}
P.~Truong, S.~Apostolopoulos, A.~Mosinska, S.~Stucky, C.~Ciller, S.~D. Zanet, Glampoints: Greedily learned accurate match points, in: Proceedings of the IEEE/CVF International Conference on Computer Vision, 2019, pp. 10732--10741.

\bibitem{wang2020learning}
Q.~Wang, X.~Zhou, B.~Hariharan, N.~Snavely, Learning feature descriptors using camera pose supervision, in: Computer Vision--ECCV 2020: 16th European Conference, Glasgow, UK, August 23--28, 2020, Proceedings, Part I 16, Springer, 2020, pp. 757--774.

\bibitem{tyszkiewicz2020disk}
M.~Tyszkiewicz, P.~Fua, E.~Trulls, Disk: Learning local features with policy gradient, Advances in Neural Information Processing Systems 33 (2020) 14254--14265.

\bibitem{lee2023learning}
J.~Lee, B.~Kim, S.~Kim, M.~Cho, Learning rotation-equivariant features for visual correspondence, in: Proceedings of the IEEE/CVF Conference on Computer Vision and Pattern Recognition, 2023, pp. 21887--21897.

\bibitem{zhou2016evaluating}
H.~Zhou, T.~Sattler, D.~W. Jacobs, Evaluating local features for day-night matching, in: Computer Vision--ECCV 2016 Workshops: Amsterdam, The Netherlands, October 8-10 and 15-16, 2016, Proceedings, Part III 14, Springer, 2016, pp. 724--736.

\bibitem{sattler2018benchmarking}
T.~Sattler, W.~Maddern, C.~Toft, A.~Torii, L.~Hammarstrand, E.~Stenborg, D.~Safari, M.~Okutomi, M.~Pollefeys, J.~Sivic, et~al., Benchmarking 6dof outdoor visual localization in changing conditions, in: Proceedings of the IEEE conference on computer vision and pattern recognition, 2018, pp. 8601--8610.

\bibitem{taira2018inloc}
H.~Taira, M.~Okutomi, T.~Sattler, M.~Cimpoi, M.~Pollefeys, J.~Sivic, T.~Pajdla, A.~Torii, Inloc: Indoor visual localization with dense matching and view synthesis, in: Proceedings of the IEEE Conference on Computer Vision and Pattern Recognition, 2018, pp. 7199--7209.

\bibitem{long2015fully}
J.~Long, E.~Shelhamer, T.~Darrell, Fully convolutional networks for semantic segmentation, in: Proceedings of the IEEE conference on computer vision and pattern recognition, 2015, pp. 3431--3440.

\bibitem{ono2018lf}
Y.~Ono, E.~Trulls, P.~Fua, K.~M. Yi, Lf-net: Learning local features from images, Advances in neural information processing systems 31 (2018).

\bibitem{shen2019rf}
X.~Shen, C.~Wang, X.~Li, Z.~Yu, J.~Li, C.~Wen, M.~Cheng, Z.~He, Rf-net: An end-to-end image matching network based on receptive field, in: Proceedings of the IEEE/CVF conference on computer vision and pattern recognition, 2019, pp. 8132--8140.

\bibitem{bhowmik2020reinforced}
A.~Bhowmik, S.~Gumhold, C.~Rother, E.~Brachmann, Reinforced feature points: Optimizing feature detection and description for a high-level task, in: Proceedings of the IEEE/CVF conference on computer vision and pattern recognition, 2020, pp. 4948--4957.

\bibitem{parihar2021rord}
U.~S. Parihar, A.~Gujarathi, K.~Mehta, S.~Tourani, S.~Garg, M.~Milford, K.~M. Krishna, Rord: Rotation-robust descriptors and orthographic views for local feature matching, in: 2021 IEEE/RSJ International Conference on Intelligent Robots and Systems (IROS), IEEE, 2021, pp. 1593--1600.

\bibitem{barroso2020hdd}
A.~Barroso-Laguna, Y.~Verdie, B.~Busam, K.~Mikolajczyk, Hdd-net: Hybrid detector descriptor with mutual interactive learning, in: Proceedings of the Asian Conference on Computer Vision, 2020.

\bibitem{zhang2020mlifeat}
Y.~Zhang, J.~Wang, S.~Xu, X.~Liu, X.~Zhang, Mlifeat: Multi-level information fusion based deep local features, in: Proceedings of the Asian Conference on Computer Vision, 2020.

\bibitem{suwanwimolkul2021learning}
S.~Suwanwimolkul, S.~Komorita, K.~Tasaka, Learning of low-level feature keypoints for accurate and robust detection, in: Proceedings of the IEEE/CVF Winter Conference on Applications of Computer Vision, 2021, pp. 2262--2271.

\bibitem{wang2023featurebooster}
X.~Wang, Z.~Liu, Y.~Hu, W.~Xi, W.~Yu, D.~Zou, Featurebooster: Boosting feature descriptors with a lightweight neural network, in: Proceedings of the IEEE/CVF Conference on Computer Vision and Pattern Recognition, 2023, pp. 7630--7639.

\bibitem{vaswani2017attention}
A.~Vaswani, N.~Shazeer, N.~Parmar, J.~Uszkoreit, L.~Jones, A.~N. Gomez, {\L}.~Kaiser, I.~Polosukhin, Attention is all you need, Advances in neural information processing systems 30 (2017).

\bibitem{luo2020aslfeat}
Z.~Luo, L.~Zhou, X.~Bai, H.~Chen, J.~Zhang, Y.~Yao, S.~Li, T.~Fang, L.~Quan, Aslfeat: Learning local features of accurate shape and localization, in: Proceedings of the IEEE/CVF conference on computer vision and pattern recognition, 2020, pp. 6589--6598.

\bibitem{fan2022learning}
B.~Fan, J.~Zhou, W.~Feng, H.~Pu, Y.~Yang, Q.~Kong, F.~Wu, H.~Liu, Learning semantic-aware local features for long term visual localization, IEEE Transactions on Image Processing 31 (2022) 4842--4855.

\bibitem{xue2023sfd2}
F.~Xue, I.~Budvytis, R.~Cipolla, Sfd2: Semantic-guided feature detection and description, in: Proceedings of the IEEE/CVF Conference on Computer Vision and Pattern Recognition, 2023, pp. 5206--5216.

\bibitem{tian2020d2d}
Y.~Tian, V.~Balntas, T.~Ng, A.~Barroso-Laguna, Y.~Demiris, K.~Mikolajczyk, D2d: Keypoint extraction with describe to detect approach, in: Proceedings of the Asian conference on computer vision, 2020.

\bibitem{li2022decoupling}
K.~Li, L.~Wang, L.~Liu, Q.~Ran, K.~Xu, Y.~Guo, Decoupling makes weakly supervised local feature better, in: Proceedings of the IEEE/CVF Conference on Computer Vision and Pattern Recognition, 2022, pp. 15838--15848.

\bibitem{deng2022redfeat}
Y.~Deng, J.~Ma, Redfeat: Recoupling detection and description for multimodal feature learning, IEEE Transactions on Image Processing 32 (2022) 591--602.

\bibitem{sun2022shared}
J.~Sun, J.~Zhu, L.~Ji, Shared coupling-bridge for weakly supervised local feature learning, arXiv preprint arXiv:2212.07047 (2022).

\bibitem{zhang2023dualgats}
D.~Zhang, F.~Chen, X.~Chen, Dualgats: Dual graph attention networks for emotion recognition in conversations, in: Proceedings of the 61st Annual Meeting of the Association for Computational Linguistics (Volume 1: Long Papers), 2023, pp. 7395--7408.

\bibitem{li2023learning}
Z.~Li, J.~Ma, Learning feature matching via matchable keypoint-assisted graph neural network, arXiv preprint arXiv:2307.01447 (2023).

\bibitem{pautrat2023gluestick}
R.~Pautrat, I.~Su{\'a}rez, Y.~Yu, M.~Pollefeys, V.~Larsson, Gluestick: Robust image matching by sticking points and lines together, arXiv preprint arXiv:2304.02008 (2023).

\bibitem{lindenberger2023lightglue}
P.~Lindenberger, P.-E. Sarlin, M.~Pollefeys, Lightglue: Local feature matching at light speed, arXiv preprint arXiv:2306.13643 (2023).

\bibitem{kuang2022densegap}
Z.~Kuang, J.~Li, M.~He, T.~Wang, Y.~Zhao, Densegap: graph-structured dense correspondence learning with anchor points, in: 2022 26th International Conference on Pattern Recognition (ICPR), IEEE, 2022, pp. 542--549.

\bibitem{cai2023htmatch}
Y.~Cai, L.~Li, D.~Wang, X.~Li, X.~Liu, Htmatch: An efficient hybrid transformer based graph neural network for local feature matching, Signal Processing 204 (2023) 108859.

\bibitem{lu2023paraformer}
X.~Lu, Y.~Yan, B.~Kang, S.~Du, Paraformer: Parallel attention transformer for efficient feature matching, arXiv preprint arXiv:2303.00941 (2023).

\bibitem{deng2023resmatch}
Y.~Deng, J.~Ma, Resmatch: Residual attention learning for local feature matching, arXiv preprint arXiv:2307.05180 (2023).

\bibitem{xie2023deepmatcher}
T.~Xie, K.~Dai, K.~Wang, R.~Li, L.~Zhao, Deepmatcher: a deep transformer-based network for robust and accurate local feature matching, arXiv preprint arXiv:2301.02993 (2023).

\bibitem{truong2020gocor}
P.~Truong, M.~Danelljan, L.~V. Gool, R.~Timofte, Gocor: Bringing globally optimized correspondence volumes into your neural network, Advances in Neural Information Processing Systems 33 (2020) 14278--14290.

\bibitem{xu2017accurate}
J.~Xu, R.~Ranftl, V.~Koltun, Accurate optical flow via direct cost volume processing, in: Proceedings of the IEEE Conference on Computer Vision and Pattern Recognition, 2017, pp. 1289--1297.

\bibitem{teed2020raft}
Z.~Teed, J.~Deng, Raft: Recurrent all-pairs field transforms for optical flow, in: Computer Vision--ECCV 2020: 16th European Conference, Glasgow, UK, August 23--28, 2020, Proceedings, Part II 16, Springer, 2020, pp. 402--419.

\bibitem{truong2023pdc}
P.~Truong, M.~Danelljan, R.~Timofte, L.~Van~Gool, Pdc-net+: Enhanced probabilistic dense correspondence network, IEEE Transactions on Pattern Analysis and Machine Intelligence (2023).

\bibitem{revaud2022pump}
J.~Revaud, V.~Leroy, P.~Weinzaepfel, B.~Chidlovskii, Pump: Pyramidal and uniqueness matching priors for unsupervised learning of local descriptors, in: Proceedings of the IEEE/CVF Conference on Computer Vision and Pattern Recognition, 2022, pp. 3926--3936.

\bibitem{revaud2016deepmatching}
J.~Revaud, P.~Weinzaepfel, Z.~Harchaoui, C.~Schmid, Deepmatching: Hierarchical deformable dense matching, International Journal of Computer Vision 120 (2016) 300--323.

\bibitem{efe2021dfm}
U.~Efe, K.~G. Ince, A.~Alatan, Dfm: A performance baseline for deep feature matching, in: Proceedings of the IEEE/CVF conference on computer vision and pattern recognition, 2021, pp. 4284--4293.

\bibitem{dosovitskiy2020image}
A.~Dosovitskiy, L.~Beyer, A.~Kolesnikov, D.~Weissenborn, X.~Zhai, T.~Unterthiner, M.~Dehghani, M.~Minderer, G.~Heigold, S.~Gelly, et~al., An image is worth 16x16 words: Transformers for image recognition at scale, arXiv preprint arXiv:2010.11929 (2020).

\bibitem{carion2020end}
N.~Carion, F.~Massa, G.~Synnaeve, N.~Usunier, A.~Kirillov, S.~Zagoruyko, End-to-end object detection with transformers, in: European conference on computer vision, Springer, 2020, pp. 213--229.

\bibitem{xu2023rssformer}
R.~Xu, C.~Wang, J.~Zhang, S.~Xu, W.~Meng, X.~Zhang, Rssformer: Foreground saliency enhancement for remote sensing land-cover segmentation, IEEE Transactions on Image Processing 32 (2023) 1052--1064.

\bibitem{xu2023scd}
R.~Xu, C.~Wang, J.~Sun, S.~Xu, W.~Meng, X.~Zhang, Self correspondence distillation for end-to-end weakly-supervised semantic segmentation, in: Proceedings of the AAAI Conference on Artificial Intelligence, 2023.

\bibitem{xu2023dual}
R.~Xu, C.~Wang, S.~Xu, W.~Meng, X.~Zhang, Dual-stream representation fusion learning for accurate medical image segmentation, Engineering Applications of Artificial Intelligence 123 (2023) 106402.

\bibitem{10328690}
W.~Cong, Y.~Cong, J.~Dong, G.~Sun, H.~Ding, Gradient-semantic compensation for incremental semantic segmentation, IEEE Transactions on Multimedia (2023) 1--14\href {https://doi.org/10.1109/TMM.2023.3336243} {\path{doi:10.1109/TMM.2023.3336243}}.

\bibitem{10214591}
W.~Cong, Y.~Cong, G.~Sun, Y.~Liu, J.~Dong, Self-paced weight consolidation for continual learning, IEEE Transactions on Circuits and Systems for Video Technology (2023) 1--1\href {https://doi.org/10.1109/TCSVT.2023.3304567} {\path{doi:10.1109/TCSVT.2023.3304567}}.

\bibitem{jiang2021cotr}
W.~Jiang, E.~Trulls, J.~Hosang, A.~Tagliasacchi, K.~M. Yi, Cotr: Correspondence transformer for matching across images, in: Proceedings of the IEEE/CVF International Conference on Computer Vision, 2021, pp. 6207--6217.

\bibitem{tan2022eco}
D.~Tan, J.-J. Liu, X.~Chen, C.~Chen, R.~Zhang, Y.~Shen, S.~Ding, R.~Ji, Eco-tr: Efficient correspondences finding via coarse-to-fine refinement, in: European Conference on Computer Vision, Springer, 2022, pp. 317--334.

\bibitem{bokman2022case}
G.~B{\"o}kman, F.~Kahl, A case for using rotation invariant features in state of the art feature matchers, in: Proceedings of the IEEE/CVF Conference on Computer Vision and Pattern Recognition, 2022, pp. 5110--5119.

\bibitem{tang2022quadtree}
S.~Tang, J.~Zhang, S.~Zhu, P.~Tan, Quadtree attention for vision transformers, arXiv preprint arXiv:2201.02767 (2022).

\bibitem{chen2022guide}
Y.~Chen, D.~Huang, S.~Xu, J.~Liu, Y.~Liu, Guide local feature matching by overlap estimation, in: Proceedings of the AAAI Conference on Artificial Intelligence, Vol.~36, 2022, pp. 365--373.

\bibitem{wang2022matchformer}
Q.~Wang, J.~Zhang, K.~Yang, K.~Peng, R.~Stiefelhagen, Matchformer: Interleaving attention in transformers for feature matching, in: Proceedings of the Asian Conference on Computer Vision, 2022, pp. 2746--2762.

\bibitem{ma2022correspondence}
J.~Ma, Y.~Wang, A.~Fan, G.~Xiao, R.~Chen, Correspondence attention transformer: A context-sensitive network for two-view correspondence learning, IEEE Transactions on Multimedia (2022).

\bibitem{giang2022topicfm}
K.~T. Giang, S.~Song, S.~Jo, Topicfm: robust and interpretable feature matching with topic-assisted, arXiv preprint arXiv:2207.00328 (2022).

\bibitem{yu2023adaptive}
J.~Yu, J.~Chang, J.~He, T.~Zhang, J.~Yu, F.~Wu, Adaptive spot-guided transformer for consistent local feature matching, in: Proceedings of the IEEE/CVF Conference on Computer Vision and Pattern Recognition, 2023, pp. 21898--21908.

\bibitem{dai2023oamatcher}
K.~Dai, T.~Xie, K.~Wang, Z.~Jiang, R.~Li, L.~Zhao, Oamatcher: An overlapping areas-based network for accurate local feature matching, arXiv preprint arXiv:2302.05846 (2023).

\bibitem{cao2023improving}
C.~Cao, Y.~Fu, Improving transformer-based image matching by cascaded capturing spatially informative keypoints, arXiv preprint arXiv:2303.02885 (2023).

\bibitem{zhu2023pmatch}
S.~Zhu, X.~Liu, Pmatch: Paired masked image modeling for dense geometric matching, in: Proceedings of the IEEE/CVF Conference on Computer Vision and Pattern Recognition, 2023, pp. 21909--21918.

\bibitem{chang2023structured}
J.~Chang, J.~Yu, T.~Zhang, Structured epipolar matcher for local feature matching, in: Proceedings of the IEEE/CVF Conference on Computer Vision and Pattern Recognition, 2023, pp. 6176--6185.

\bibitem{edstedt2023dkm}
J.~Edstedt, I.~Athanasiadis, M.~Wadenb{\"a}ck, M.~Felsberg, Dkm: Dense kernelized feature matching for geometry estimation, in: Proceedings of the IEEE/CVF Conference on Computer Vision and Pattern Recognition, 2023, pp. 17765--17775.

\bibitem{edstedt2023roma}
J.~Edstedt, Q.~Sun, G.~B{\"o}kman, M.~Wadenb{\"a}ck, M.~Felsberg, Roma: Revisiting robust losses for dense feature matching, arXiv preprint arXiv:2305.15404 (2023).

\bibitem{zhou2021patch2pix}
Q.~Zhou, T.~Sattler, L.~Leal-Taixe, Patch2pix: Epipolar-guided pixel-level correspondences, in: Proceedings of the IEEE/CVF conference on computer vision and pattern recognition, 2021, pp. 4669--4678.

\bibitem{huang2023adaptive}
D.~Huang, Y.~Chen, Y.~Liu, J.~Liu, S.~Xu, W.~Wu, Y.~Ding, F.~Tang, C.~Wang, Adaptive assignment for geometry aware local feature matching, in: Proceedings of the IEEE/CVF Conference on Computer Vision and Pattern Recognition, 2023, pp. 5425--5434.

\bibitem{ni2023pats}
J.~Ni, Y.~Li, Z.~Huang, H.~Li, H.~Bao, Z.~Cui, G.~Zhang, Pats: Patch area transportation with subdivision for local feature matching, in: Proceedings of the IEEE/CVF Conference on Computer Vision and Pattern Recognition, 2023, pp. 17776--17786.

\bibitem{zhang2023searching}
Y.~Zhang, X.~Zhao, D.~Qian, Searching from area to point: A hierarchical framework for semantic-geometric combined feature matching, arXiv preprint arXiv:2305.00194 (2023).

\bibitem{snavely2006photo}
N.~Snavely, S.~M. Seitz, R.~Szeliski, Photo tourism: exploring photo collections in 3d, in: ACM siggraph 2006 papers, 2006, pp. 835--846.

\bibitem{lindenberger2021pixel}
P.~Lindenberger, P.-E. Sarlin, V.~Larsson, M.~Pollefeys, Pixel-perfect structure-from-motion with featuremetric refinement, in: Proceedings of the IEEE/CVF international conference on computer vision, 2021, pp. 5987--5997.

\bibitem{parameshwara2022diffposenet}
C.~M. Parameshwara, G.~Hari, C.~Ferm{\"u}ller, N.~J. Sanket, Y.~Aloimonos, Diffposenet: direct differentiable camera pose estimation, in: Proceedings of the IEEE/CVF Conference on Computer Vision and Pattern Recognition, 2022, pp. 6845--6854.

\bibitem{zhang2022relpose}
J.~Y. Zhang, D.~Ramanan, S.~Tulsiani, Relpose: Predicting probabilistic relative rotation for single objects in the wild, in: European Conference on Computer Vision, Springer, 2022, pp. 592--611.

\bibitem{tang2018ba}
C.~Tang, P.~Tan, Ba-net: Dense bundle adjustment network, arXiv preprint arXiv:1806.04807 (2018).

\bibitem{gu2021dro}
X.~Gu, W.~Yuan, Z.~Dai, C.~Tang, S.~Zhu, P.~Tan, Dro: Deep recurrent optimizer for structure-from-motion, arXiv preprint arXiv:2103.13201 (2021).

\bibitem{he2023detector}
X.~He, J.~Sun, Y.~Wang, S.~Peng, Q.~Huang, H.~Bao, X.~Zhou, Detector-free structure from motion, arXiv preprint arXiv:2306.15669 (2023).

\bibitem{hughes2020deep}
L.~H. Hughes, D.~Marcos, S.~Lobry, D.~Tuia, M.~Schmitt, A deep learning framework for matching of sar and optical imagery, ISPRS Journal of Photogrammetry and Remote Sensing 169 (2020) 166--179.

\bibitem{ye2022multiscale}
Y.~Ye, T.~Tang, B.~Zhu, C.~Yang, B.~Li, S.~Hao, A multiscale framework with unsupervised learning for remote sensing image registration, IEEE Transactions on Geoscience and Remote Sensing 60 (2022) 1--15.

\bibitem{chen2022hierarchical}
S.~Chen, J.~Chen, Y.~Rao, X.~Chen, X.~Fan, H.~Bai, L.~Xing, C.~Zhou, Y.~Yang, A hierarchical consensus attention network for feature matching of remote sensing images, IEEE Transactions on Geoscience and Remote Sensing 60 (2022) 1--11.

\bibitem{liu2022progressive}
Y.~Liu, B.~N. Zhao, S.~Zhao, L.~Zhang, Progressive motion coherence for remote sensing image matching, IEEE Transactions on Geoscience and Remote Sensing 60 (2022) 1--13.

\bibitem{ye2022robust}
Y.~Ye, B.~Zhu, T.~Tang, C.~Yang, Q.~Xu, G.~Zhang, A robust multimodal remote sensing image registration method and system using steerable filters with first-and second-order gradients, ISPRS Journal of Photogrammetry and Remote Sensing 188 (2022) 331--350.

\bibitem{hughes2018identifying}
L.~H. Hughes, M.~Schmitt, L.~Mou, Y.~Wang, X.~X. Zhu, Identifying corresponding patches in sar and optical images with a pseudo-siamese cnn, IEEE Geoscience and Remote Sensing Letters 15~(5) (2018) 784--788.

\bibitem{quan2018deep}
D.~Quan, S.~Wang, X.~Liang, R.~Wang, S.~Fang, B.~Hou, L.~Jiao, Deep generative matching network for optical and sar image registration, in: IGARSS 2018-2018 IEEE International Geoscience and Remote Sensing Symposium, IEEE, 2018, pp. 6215--6218.

\bibitem{merkle2018exploring}
N.~Merkle, S.~Auer, R.~Mueller, P.~Reinartz, Exploring the potential of conditional adversarial networks for optical and sar image matching, IEEE Journal of Selected Topics in Applied Earth Observations and Remote Sensing 11~(6) (2018) 1811--1820.

\bibitem{shi2012visual}
W.~Shi, F.~Su, R.~Wang, J.~Fan, A visual circle based image registration algorithm for optical and sar imagery, in: 2012 IEEE International Geoscience and Remote Sensing Symposium, IEEE, 2012, pp. 2109--2112.

\bibitem{zampieri2018multimodal}
A.~Zampieri, G.~Charpiat, N.~Girard, Y.~Tarabalka, Multimodal image alignment through a multiscale chain of neural networks with application to remote sensing, in: Proceedings of the European Conference on Computer Vision (ECCV), 2018, pp. 657--673.

\bibitem{ye2018remote}
F.~Ye, Y.~Su, H.~Xiao, X.~Zhao, W.~Min, Remote sensing image registration using convolutional neural network features, IEEE Geoscience and Remote Sensing Letters 15~(2) (2018) 232--236.

\bibitem{wang2017multi}
T.~Wang, G.~Zhang, L.~Yu, R.~Zhao, M.~Deng, K.~Xu, Multi-mode gf-3 satellite image geometric accuracy verification using the rpc model, Sensors 17~(9) (2017) 2005.

\bibitem{ma2019novel}
W.~Ma, J.~Zhang, Y.~Wu, L.~Jiao, H.~Zhu, W.~Zhao, A novel two-step registration method for remote sensing images based on deep and local features, IEEE Transactions on Geoscience and Remote Sensing 57~(7) (2019) 4834--4843.

\bibitem{zhou2021robust}
L.~Zhou, Y.~Ye, T.~Tang, K.~Nan, Y.~Qin, Robust matching for sar and optical images using multiscale convolutional gradient features, IEEE Geoscience and Remote Sensing Letters 19 (2021) 1--5.

\bibitem{cui2021map}
S.~Cui, A.~Ma, L.~Zhang, M.~Xu, Y.~Zhong, Map-net: Sar and optical image matching via image-based convolutional network with attention mechanism and spatial pyramid aggregated pooling, IEEE Transactions on Geoscience and Remote Sensing 60 (2021) 1--13.

\bibitem{wang2023review}
Z.~Wang, Y.~Ma, Y.~Zhang, Review of pixel-level remote sensing image fusion based on deep learning, Information Fusion 90 (2023) 36--58.

\bibitem{bian2022learning}
Z.~Bian, A.~Jabri, A.~A. Efros, A.~Owens, Learning pixel trajectories with multiscale contrastive random walks, in: Proceedings of the IEEE/CVF Conference on Computer Vision and Pattern Recognition, 2022, pp. 6508--6519.

\bibitem{ranjan2019competitive}
A.~Ranjan, V.~Jampani, L.~Balles, K.~Kim, D.~Sun, J.~Wulff, M.~J. Black, Competitive collaboration: Joint unsupervised learning of depth, camera motion, optical flow and motion segmentation, in: Proceedings of the IEEE/CVF conference on computer vision and pattern recognition, 2019, pp. 12240--12249.

\bibitem{harley2022particle}
A.~W. Harley, Z.~Fang, K.~Fragkiadaki, Particle video revisited: Tracking through occlusions using point trajectories, in: European Conference on Computer Vision, Springer, 2022, pp. 59--75.

\bibitem{qin2023generative}
C.~Qin, S.~Wang, C.~Chen, W.~Bai, D.~Rueckert, Generative myocardial motion tracking via latent space exploration with biomechanics-informed prior, Medical Image Analysis 83 (2023) 102682.

\bibitem{ye2021deeptag}
M.~Ye, M.~Kanski, D.~Yang, Q.~Chang, Z.~Yan, Q.~Huang, L.~Axel, D.~Metaxas, Deeptag: An unsupervised deep learning method for motion tracking on cardiac tagging magnetic resonance images, in: Proceedings of the IEEE/CVF conference on computer vision and pattern recognition, 2021, pp. 7261--7271.

\bibitem{bian2023drimet}
Z.~Bian, F.~Xing, J.~Yu, M.~Shao, Y.~Liu, A.~Carass, J.~Woo, J.~L. Prince, Drimet: Deep registration-based 3d incompressible motion estimation in tagged-mri with application to the tongue, in: Medical Imaging with Deep Learning, 2023.

\bibitem{fechter2020one}
T.~Fechter, D.~Baltas, One-shot learning for deformable medical image registration and periodic motion tracking, IEEE transactions on medical imaging 39~(7) (2020) 2506--2517.

\bibitem{zhang2021groupregnet}
Y.~Zhang, X.~Wu, H.~M. Gach, H.~Li, D.~Yang, Groupregnet: a groupwise one-shot deep learning-based 4d image registration method, Physics in Medicine \& Biology 66~(4) (2021) 045030.

\bibitem{ji2022one}
Y.~Ji, Z.~Zhu, Y.~Wei, A one-shot lung 4d-ct image registration method with temporal-spatial features, in: 2022 IEEE Biomedical Circuits and Systems Conference (BioCAS), IEEE, 2022, pp. 203--207.

\bibitem{iqbal2024hybrid}
M.~Z. Iqbal, I.~Razzak, A.~Qayyum, T.~T. Nguyen, M.~Tanveer, A.~Sowmya, Hybrid unsupervised paradigm based deformable image fusion for 4d ct lung image modality, Information Fusion 102 (2024) 102061.

\bibitem{pfandler2019technical}
M.~Pfandler, P.~Stefan, C.~Mehren, M.~Lazarovici, M.~Weigl, Technical and nontechnical skills in surgery: a simulated operating room environment study, Spine 44~(23) (2019) E1396--E1400.

\bibitem{maes1997multimodality}
F.~Maes, A.~Collignon, D.~Vandermeulen, G.~Marchal, P.~Suetens, Multimodality image registration by maximization of mutual information, IEEE transactions on Medical Imaging 16~(2) (1997) 187--198.

\bibitem{unberath2021impact}
M.~Unberath, C.~Gao, Y.~Hu, M.~Judish, R.~H. Taylor, M.~Armand, R.~Grupp, The impact of machine learning on 2d/3d registration for image-guided interventions: A systematic review and perspective, Frontiers in Robotics and AI 8 (2021) 716007.

\bibitem{jaganathan2023self}
S.~Jaganathan, M.~Kukla, J.~Wang, K.~Shetty, A.~Maier, Self-supervised 2d/3d registration for x-ray to ct image fusion, in: Proceedings of the IEEE/CVF Winter Conference on Applications of Computer Vision, 2023, pp. 2788--2798.

\bibitem{huang2022novel}
D.-X. Huang, X.-H. Zhou, X.-L. Xie, S.-Q. Liu, Z.-Q. Feng, J.-L. Hao, Z.-G. Hou, N.~Ma, L.~Yan, A novel two-stage framework for 2d/3d registration in neurological interventions, in: 2022 IEEE International Conference on Robotics and Biomimetics (ROBIO), IEEE, 2022, pp. 266--271.

\bibitem{pei2017non}
Y.~Pei, Y.~Zhang, H.~Qin, G.~Ma, Y.~Guo, T.~Xu, H.~Zha, Non-rigid craniofacial 2d-3d registration using cnn-based regression, in: Deep Learning in Medical Image Analysis and Multimodal Learning for Clinical Decision Support: Third International Workshop, DLMIA 2017, and 7th International Workshop, ML-CDS 2017, Held in Conjunction with MICCAI 2017, Qu{\'e}bec City, QC, Canada, September 14, Proceedings 3, Springer, 2017, pp. 117--125.

\bibitem{foote2019real}
M.~D. Foote, B.~E. Zimmerman, A.~Sawant, S.~C. Joshi, Real-time 2d-3d deformable registration with deep learning and application to lung radiotherapy targeting, in: Information Processing in Medical Imaging: 26th International Conference, IPMI 2019, Hong Kong, China, June 2--7, 2019, Proceedings 26, Springer, 2019, pp. 265--276.

\bibitem{yu2017non}
W.~Yu, M.~Tannast, G.~Zheng, Non-rigid free-form 2d--3d registration using a b-spline-based statistical deformation model, Pattern recognition 63 (2017) 689--699.

\bibitem{li2020non}
P.~Li, Y.~Pei, Y.~Guo, G.~Ma, T.~Xu, H.~Zha, Non-rigid 2d-3d registration using convolutional autoencoders, in: 2020 IEEE 17th International Symposium on Biomedical Imaging (ISBI), IEEE, 2020, pp. 700--704.

\bibitem{dong20232d}
G.~Dong, J.~Dai, N.~Li, C.~Zhang, W.~He, L.~Liu, Y.~Chan, Y.~Li, Y.~Xie, X.~Liang, 2d/3d non-rigid image registration via two orthogonal x-ray projection images for lung tumor tracking, Bioengineering 10~(2) (2023) 144.

\bibitem{balntas2017hpatches}
V.~Balntas, K.~Lenc, A.~Vedaldi, K.~Mikolajczyk, Hpatches: A benchmark and evaluation of handcrafted and learned local descriptors, in: Proceedings of the IEEE conference on computer vision and pattern recognition, 2017, pp. 5173--5182.

\bibitem{shen2024gim}
X.~Shen, Z.~Cai, W.~Yin, M.~M{\"u}ller, Z.~Li, K.~Wang, X.~Chen, C.~Wang, Gim: Learning generalizable image matcher from internet videos, arXiv preprint arXiv:2402.11095 (2024).

\bibitem{cvpr2020_image_matching_challenge}
\href{https://www.cs.ubc.ca/research/image-matching-challenge/2020/}{Cvpr 2020 image matching challenge}, accessed March 1, 2024.
\newline\urlprefix\url{https://www.cs.ubc.ca/research/image-matching-challenge/2020/}

\bibitem{cvpr2021_image_matching_challenge}
\href{https://www.cs.ubc.ca/research/image-matching-challenge/current/}{Cvpr 2021 image matching challenge}, accessed March 1, 2024.
\newline\urlprefix\url{https://www.cs.ubc.ca/research/image-matching-challenge/current/}

\bibitem{cvpr2022_image_matching_challenge}
\href{https://www.kaggle.com/competitions/image-matching-challenge-2022/overview}{Cvpr 2022 image matching challenge}, accessed March 1, 2024.
\newline\urlprefix\url{https://www.kaggle.com/competitions/image-matching-challenge-2022/overview}

\bibitem{cvpr2023_image_matching_challenge}
\href{https://www.kaggle.com/competitions/image-matching-challenge-2023/overview}{Cvpr 2023 image matching challenge}, accessed March 1, 2024.
\newline\urlprefix\url{https://www.kaggle.com/competitions/image-matching-challenge-2023/overview}

\bibitem{dai2017scannet}
A.~Dai, A.~X. Chang, M.~Savva, M.~Halber, T.~Funkhouser, M.~Nie{\ss}ner, Scannet: Richly-annotated 3d reconstructions of indoor scenes, in: Proceedings of the IEEE conference on computer vision and pattern recognition, 2017, pp. 5828--5839.

\bibitem{thomee2016yfcc100m}
B.~Thomee, D.~A. Shamma, G.~Friedland, B.~Elizalde, K.~Ni, D.~Poland, D.~Borth, L.-J. Li, Yfcc100m: The new data in multimedia research, Communications of the ACM 59~(2) (2016) 64--73.

\bibitem{li2018megadepth}
Z.~Li, N.~Snavely, Megadepth: Learning single-view depth prediction from internet photos, in: Proceedings of the IEEE conference on computer vision and pattern recognition, 2018, pp. 2041--2050.

\bibitem{mishkin2015mods}
D.~Mishkin, J.~Matas, M.~Perdoch, Mods: Fast and robust method for two-view matching, Computer vision and image understanding 141 (2015) 81--93.

\bibitem{mishkin2015wxbs}
D.~Mishkin, J.~Matas, M.~Perdoch, K.~Lenc, Wxbs: Wide baseline stereo generalizations, arXiv preprint arXiv:1504.06603 (2015).

\bibitem{sattler2012image}
T.~Sattler, T.~Weyand, B.~Leibe, L.~Kobbelt, Image retrieval for image-based localization revisited., in: BMVC, Vol.~1, 2012, p.~4.

\bibitem{maddern20171}
W.~Maddern, G.~Pascoe, C.~Linegar, P.~Newman, 1 year, 1000 km: The oxford robotcar dataset, The International Journal of Robotics Research 36~(1) (2017) 3--15.

\bibitem{sarlin2022lamar}
P.-E. Sarlin, M.~Dusmanu, J.~L. Sch{\"o}nberger, P.~Speciale, L.~Gruber, V.~Larsson, O.~Miksik, M.~Pollefeys, Lamar: Benchmarking localization and mapping for augmented reality, in: European Conference on Computer Vision, Springer, 2022, pp. 686--704.

\bibitem{geiger2013vision}
A.~Geiger, P.~Lenz, C.~Stiller, R.~Urtasun, Vision meets robotics: The kitti dataset, The International Journal of Robotics Research 32~(11) (2013) 1231--1237.

\bibitem{xiao2013sun3d}
J.~Xiao, A.~Owens, A.~Torralba, Sun3d: A database of big spaces reconstructed using sfm and object labels, in: Proceedings of the IEEE international conference on computer vision, 2013, pp. 1625--1632.

\bibitem{aanaes2016large}
H.~Aan{\ae}s, R.~R. Jensen, G.~Vogiatzis, E.~Tola, A.~B. Dahl, Large-scale data for multiple-view stereopsis, International Journal of Computer Vision 120 (2016) 153--168.

\bibitem{knapitsch2017tanks}
A.~Knapitsch, J.~Park, Q.-Y. Zhou, V.~Koltun, Tanks and temples: Benchmarking large-scale scene reconstruction, ACM Transactions on Graphics (ToG) 36~(4) (2017) 1--13.

\bibitem{schonberger2017comparative}
J.~L. Schonberger, H.~Hardmeier, T.~Sattler, M.~Pollefeys, Comparative evaluation of hand-crafted and learned local features, in: Proceedings of the IEEE conference on computer vision and pattern recognition, 2017, pp. 1482--1491.

\bibitem{wilson2014robust}
K.~Wilson, N.~Snavely, Robust global translations with 1dsfm, in: Computer Vision--ECCV 2014: 13th European Conference, Zurich, Switzerland, September 6-12, 2014, Proceedings, Part III 13, Springer, 2014, pp. 61--75.

\bibitem{schops2017multi}
T.~Schops, J.~L. Schonberger, S.~Galliani, T.~Sattler, K.~Schindler, M.~Pollefeys, A.~Geiger, A multi-view stereo benchmark with high-resolution images and multi-camera videos, in: Proceedings of the IEEE conference on computer vision and pattern recognition, 2017, pp. 3260--3269.

\bibitem{marelli2023enrich}
D.~Marelli, L.~Morelli, E.~M. Farella, S.~Bianco, G.~Ciocca, F.~Remondino, Enrich: Multi-purpose dataset for benchmarking in computer vision and photogrammetry, ISPRS Journal of Photogrammetry and Remote Sensing 198 (2023) 84--98.

\bibitem{schmid2000evaluation}
C.~Schmid, R.~Mohr, C.~Bauckhage, Evaluation of interest point detectors, International Journal of computer vision 37~(2) (2000) 151--172.

\bibitem{choy2016universal}
C.~B. Choy, J.~Gwak, S.~Savarese, M.~Chandraker, Universal correspondence network, Advances in neural information processing systems 29 (2016).

\bibitem{melekhov2019dgc}
I.~Melekhov, A.~Tiulpin, T.~Sattler, M.~Pollefeys, E.~Rahtu, J.~Kannala, Dgc-net: Dense geometric correspondence network, in: 2019 IEEE Winter Conference on Applications of Computer Vision (WACV), IEEE, 2019, pp. 1034--1042.

\bibitem{shen2020ransac}
X.~Shen, F.~Darmon, A.~A. Efros, M.~Aubry, Ransac-flow: generic two-stage image alignment, in: Computer Vision--ECCV 2020: 16th European Conference, Glasgow, UK, August 23--28, 2020, Proceedings, Part IV 16, Springer, 2020, pp. 618--637.

\bibitem{sarlin2019coarse}
P.-E. Sarlin, C.~Cadena, R.~Siegwart, M.~Dymczyk, From coarse to fine: Robust hierarchical localization at large scale, in: Proceedings of the IEEE/CVF Conference on Computer Vision and Pattern Recognition, 2019, pp. 12716--12725.

\bibitem{nan2022learning}
X.~Nan, L.~Ding, Learning geometric feature embedding with transformers for image matching, Sensors 22~(24) (2022) 9882.

\bibitem{mao20223dg}
R.~Mao, C.~Bai, Y.~An, F.~Zhu, C.~Lu, 3dg-stfm: 3d geometric guided student-teacher feature matching, in: European Conference on Computer Vision, Springer, 2022, pp. 125--142.

\bibitem{yi2018learning}
K.~M. Yi, E.~Trulls, Y.~Ono, V.~Lepetit, M.~Salzmann, P.~Fua, Learning to find good correspondences, in: Proceedings of the IEEE conference on computer vision and pattern recognition, 2018, pp. 2666--2674.

\bibitem{wiles2021co}
O.~Wiles, S.~Ehrhardt, A.~Zisserman, Co-attention for conditioned image matching, in: Proceedings of the IEEE/CVF conference on computer vision and pattern recognition, 2021, pp. 15920--15929.

\bibitem{arandjelovic2012three}
R.~Arandjelovi{\'c}, A.~Zisserman, Three things everyone should know to improve object retrieval, in: 2012 IEEE conference on computer vision and pattern recognition, IEEE, 2012, pp. 2911--2918.

\bibitem{melekhov2020image}
I.~Melekhov, G.~J. Brostow, J.~Kannala, D.~Turmukhambetov, Image stylization for robust features, arXiv preprint arXiv:2008.06959 (2020).

\bibitem{zhou2021retrieval}
Y.~Zhou, H.~Fan, S.~Gao, Y.~Yang, X.~Zhang, J.~Li, Y.~Guo, Retrieval and localization with observation constraints, in: 2021 IEEE International Conference on Robotics and Automation (ICRA), IEEE, 2021, pp. 5237--5244.

\bibitem{humenberger2020robust}
M.~Humenberger, Y.~Cabon, N.~Guerin, J.~Morat, V.~Leroy, J.~Revaud, P.~Rerole, N.~Pion, C.~de~Souza, G.~Csurka, Robust image retrieval-based visual localization using kapture, arXiv preprint arXiv:2007.13867 (2020).

\bibitem{germain2020s2dnet}
H.~Germain, G.~Bourmaud, V.~Lepetit, S2dnet: Learning image features for accurate sparse-to-dense matching, in: Computer Vision--ECCV 2020: 16th European Conference, Glasgow, UK, August 23--28, 2020, Proceedings, Part III 16, Springer, 2020, pp. 626--643.

\bibitem{zhao2022dsd}
Y.~Zhao, H.~Zhang, P.~Lu, P.~Li, E.~Wu, B.~Sheng, Dsd-matchingnet: Deformable sparse-to-dense feature matching for learning accurate correspondences, Virtual Reality \& Intelligent Hardware 4~(5) (2022) 432--443.

\bibitem{chen2022deep}
L.~Chen, C.~Heipke, Deep learning feature representation for image matching under large viewpoint and viewing direction change, ISPRS Journal of Photogrammetry and Remote Sensing 190 (2022) 94--112.

\bibitem{cordts2016cityscapes}
M.~Cordts, M.~Omran, S.~Ramos, T.~Rehfeld, M.~Enzweiler, R.~Benenson, U.~Franke, S.~Roth, B.~Schiele, The cityscapes dataset for semantic urban scene understanding, in: Proceedings of the IEEE conference on computer vision and pattern recognition, 2016, pp. 3213--3223.

\bibitem{zhou2017scene}
B.~Zhou, H.~Zhao, X.~Puig, S.~Fidler, A.~Barriuso, A.~Torralba, Scene parsing through ade20k dataset, in: Proceedings of the IEEE conference on computer vision and pattern recognition, 2017, pp. 633--641.

\bibitem{kirillov2023segment}
A.~Kirillov, E.~Mintun, N.~Ravi, H.~Mao, C.~Rolland, L.~Gustafson, T.~Xiao, S.~Whitehead, A.~C. Berg, W.-Y. Lo, et~al., Segment anything, arXiv preprint arXiv:2304.02643 (2023).

\bibitem{caron2021emerging}
M.~Caron, H.~Touvron, I.~Misra, H.~J{\'e}gou, J.~Mairal, P.~Bojanowski, A.~Joulin, Emerging properties in self-supervised vision transformers, in: Proceedings of the IEEE/CVF international conference on computer vision, 2021, pp. 9650--9660.

\bibitem{oquab2023dinov2}
M.~Oquab, T.~Darcet, T.~Moutakanni, H.~Vo, M.~Szafraniec, V.~Khalidov, P.~Fernandez, D.~Haziza, F.~Massa, A.~El-Nouby, et~al., Dinov2: Learning robust visual features without supervision, arXiv preprint arXiv:2304.07193 (2023).

\bibitem{anonymous2023towards}
Anonymous, \href{https://openreview.net/forum?id=TVg6hlfsKa}{Towards seamless adaptation of pre-trained models for visual place recognition}, in: Submitted to The Twelfth International Conference on Learning Representations, 2023, under review.
\newline\urlprefix\url{https://openreview.net/forum?id=TVg6hlfsKa}

\bibitem{jiang2022robust}
X.~Jiang, Y.~Xia, X.-P. Zhang, J.~Ma, Robust image matching via local graph structure consensus, Pattern Recognition 126 (2022) 108588.

\bibitem{ma2019locality}
J.~Ma, J.~Zhao, J.~Jiang, H.~Zhou, X.~Guo, Locality preserving matching, International Journal of Computer Vision 127 (2019) 512--531.

\bibitem{raguram2012usac}
R.~Raguram, O.~Chum, M.~Pollefeys, J.~Matas, J.-M. Frahm, Usac: A universal framework for random sample consensus, IEEE transactions on pattern analysis and machine intelligence 35~(8) (2012) 2022--2038.

\bibitem{barath2020magsac++}
D.~Barath, J.~Noskova, M.~Ivashechkin, J.~Matas, Magsac++, a fast, reliable and accurate robust estimator, in: Proceedings of the IEEE/CVF conference on computer vision and pattern recognition, 2020, pp. 1304--1312.

\bibitem{li2010rejecting}
X.~Li, Z.~Hu, Rejecting mismatches by correspondence function, International Journal of Computer Vision 89 (2010) 1--17.

\bibitem{sun2020acne}
W.~Sun, W.~Jiang, E.~Trulls, A.~Tagliasacchi, K.~M. Yi, Acne: Attentive context normalization for robust permutation-equivariant learning, in: Proceedings of the IEEE/CVF conference on computer vision and pattern recognition, 2020, pp. 11286--11295.

\bibitem{zhao2019nm}
C.~Zhao, Z.~Cao, C.~Li, X.~Li, J.~Yang, Nm-net: Mining reliable neighbors for robust feature correspondences, in: Proceedings of the IEEE/CVF conference on computer vision and pattern recognition, 2019, pp. 215--224.

\bibitem{chen2023shape}
J.~Chen, X.~Chen, S.~Chen, Y.~Liu, Y.~Rao, Y.~Yang, H.~Wang, D.~Wu, Shape-former: Bridging cnn and transformer via shapeconv for multimodal image matching, Information Fusion 91 (2023) 445--457.

\bibitem{brachmann2017dsac}
E.~Brachmann, A.~Krull, S.~Nowozin, J.~Shotton, F.~Michel, S.~Gumhold, C.~Rother, Dsac-differentiable ransac for camera localization, in: Proceedings of the IEEE conference on computer vision and pattern recognition, 2017, pp. 6684--6692.

\bibitem{cavalli2023consensus}
L.~Cavalli, D.~Barath, M.~Pollefeys, V.~Larsson, Consensus-adaptive ransac, arXiv preprint arXiv:2307.14030 (2023).

\bibitem{chen2021lsv}
J.~Chen, S.~Chen, X.~Chen, Y.~Yang, L.~Xing, X.~Fan, Y.~Rao, Lsv-anet: Deep learning on local structure visualization for feature matching, IEEE Transactions on Geoscience and Remote Sensing 60 (2021) 1--18.

\bibitem{bellavia2022image}
F.~Bellavia, L.~Morelli, F.~Menna, F.~Remondino, Image orientation with a hybrid pipeline robust to rotations and wide-baselines, The International Archives of the Photogrammetry, Remote Sensing and Spatial Information Sciences 46 (2022) 73--80.

\bibitem{bellavia2022harrisz+}
F.~Bellavia, D.~Mishkin, Harrisz+: Harris corner selection for next-gen image matching pipelines, Pattern Recognition Letters 158 (2022) 141--147.

\bibitem{bellavia2024image}
F.~Bellavia, Image matching by bare homography, IEEE Transactions on Image Processing (2024).

\bibitem{maiwald2021fully}
F.~Maiwald, C.~Lehmann, T.~Lazariv, Fully automated pose estimation of historical images in the context of 4d geographic information systems utilizing machine learning methods, ISPRS International Journal of Geo-Information 10~(11) (2021) 748.

\bibitem{morelli2022photogrammetry}
L.~Morelli, F.~Bellavia, F.~Menna, F.~Remondino, Photogrammetry now and then--from hand-crafted to deep-learning tie points--, The International Archives of the Photogrammetry, Remote Sensing and Spatial Information Sciences 48 (2022) 163--170.

\bibitem{maiwald2021automatic}
F.~Maiwald, H.-G. Maas, An automatic workflow for orientation of historical images with large radiometric and geometric differences, The Photogrammetric Record 36~(174) (2021) 77--103.

\end{thebibliography}

\end{sloppypar}
\end{document}